%% file: main-ACCV.tex
\documentclass[runningheads]{llncs}

\usepackage{accv}

\usepackage{accvabbrv}

\usepackage{graphicx}
\usepackage{booktabs}
\usepackage{wrapfig}

\usepackage{amssymb}
\usepackage{amsmath}
\usepackage{multirow}
\usepackage{csquotes}
\usepackage[table]{xcolor}

\usepackage[accsupp]{axessibility}  

\usepackage[pagebackref,breaklinks,colorlinks,citecolor=accvblue]{hyperref}

\usepackage{orcidlink}

\begin{document}

\title{MSRNet: A Multi-Scale Recursive Network for Camouflaged Object Detection} 


\author{Leena Alghamdi\inst{1}\and
Muhammad Usman\inst{2}\and
Hafeez Anwar\inst{3}\and
Abdul Bais\inst{4}\and
Saeed Anwar\inst{5}
}

\authorrunning{L.~Alghamdi et al.}

\institute{King Fahd University of Petroleum and Minerals, KSA\\\and
Ontario Tech University, Canada\\\and
National University of Computer and Emerging Sciences, Pakistan\\\and
The University of Regina, Canada\\\and
The University of Western Australia, Australia
}

\maketitle
\begin{abstract}
Camouflaged object detection is an emerging and challenging computer vision task that requires identifying and segmenting objects that blend seamlessly into their environments due to high similarity in color, texture, and size. This task is further complicated by low-light conditions, partial occlusion, small object size, intricate background patterns, and multiple objects. While many sophisticated methods have been proposed for this task, current methods still struggle to precisely detect camouflaged objects in complex scenarios, especially with small and multiple objects, indicating room for improvement. We propose a Multi-Scale Recursive Network that extracts multi-scale features using a Pyramid Vision Transformer backbone and combines them with specialized Attention-Based Scale Integration Units, thereby enabling selective feature merging. For more precise object detection, our decoder recursively refines features by incorporating Multi-Granularity Fusion Units. A novel recursive-feedback decoding strategy is developed to enhance the model's understanding of global context, thereby helping it overcome the challenges of this task. By jointly leveraging multi-scale learning and recursive feature optimization, our proposed method achieves performance gains, successfully detecting small and multiple camouflaged objects. Our model achieves state-of-the-art results on two benchmark datasets for camouflaged object detection and ranks second on the remaining two. 
Our code, model weights, and results are available at \href{https://github.com/linaagh98/MSRNet}{https://github.com/linaagh98/MSRNet}.
  
\end{abstract}

\section{Introduction}
\label{sec:intro}
Camouflaged object detection (COD)~\cite{Han_2026_CVPR,Du_2025_CVPR} is an emerging and challenging domain in computer vision that focuses on identifying and segmenting objects that blend seamlessly with their surroundings~\cite{bhajantri2006camouflage}. The complexity of this task arises from the high similarity between camouflaged objects and their backgrounds in color, texture, and size. Additional factors, including low-light conditions, occlusion, small size, and complex patterns, further complicate the task in certain scenarios~\cite{zhuge2022cubenet}. While COD primarily focuses on the recognition of camouflaged objects, such as animals concealed from predators or soldiers in camouflage uniforms~\cite{singh2013survey}, it also has considerable relevance across multiple domains and applications. For instance, it proves beneficial in medical imaging for activities such as polyp segmentation~\cite{fan2020pranet, ji2022video, zhao2021automatic} and lung infection detection~\cite{fan2020inf}, as well as in the management of agricultural operations~\cite{liu2019pestnet, rizzo2023fruit} and in search-and-rescue missions~\cite{fan2021concealed}. Furthermore, it contributes to the development of additional vision-related tasks, including transparent object detection~\cite{khaing2019transparent} and defect identification~\cite{zeng2022small}.

\begin{figure}[tb]
  \centering
\begin{tabular}{cccccccc}
\includegraphics[width=0.118\textwidth]{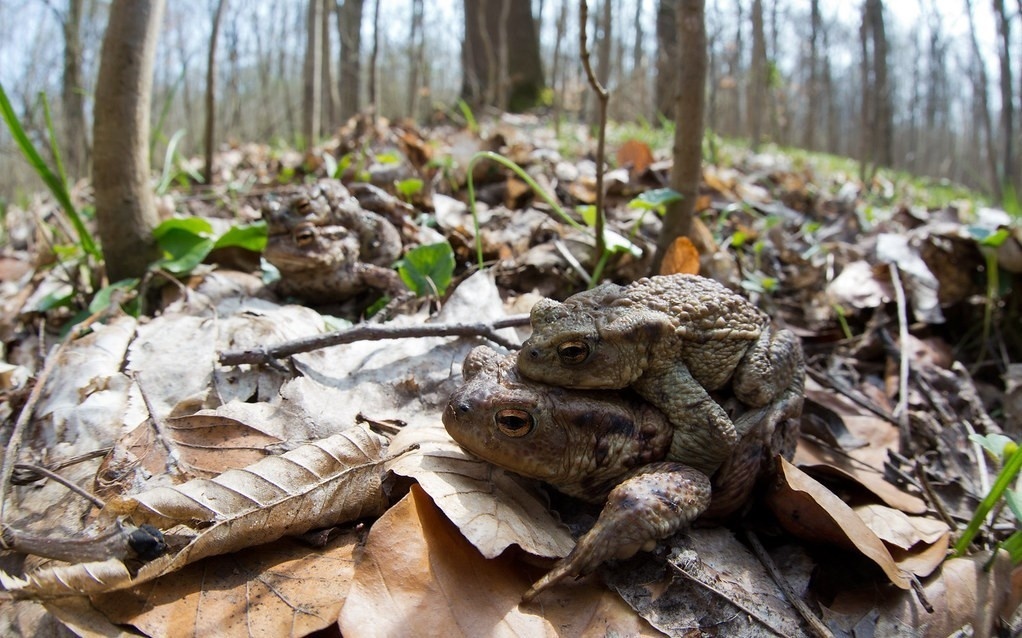} & 
\includegraphics[width=0.118\textwidth]{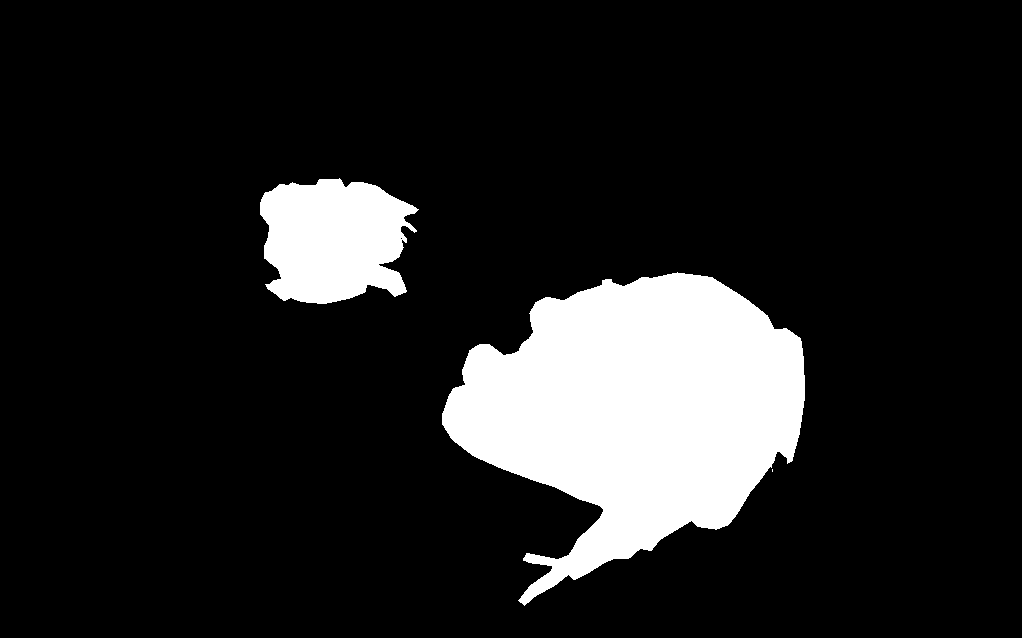} & 
\includegraphics[width=0.118\textwidth]{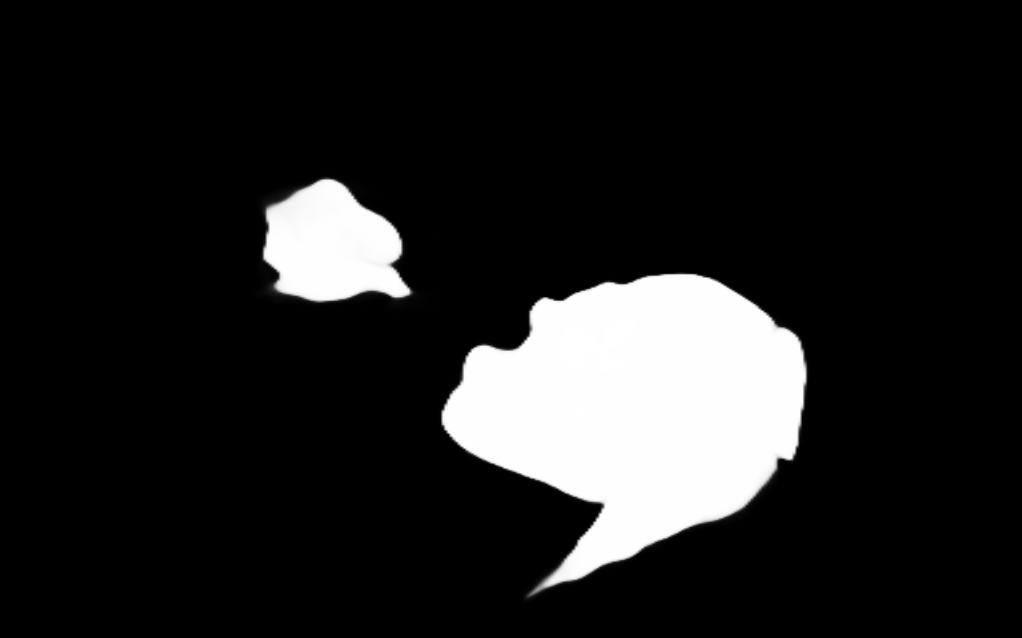} & 
\includegraphics[width=0.118\textwidth]{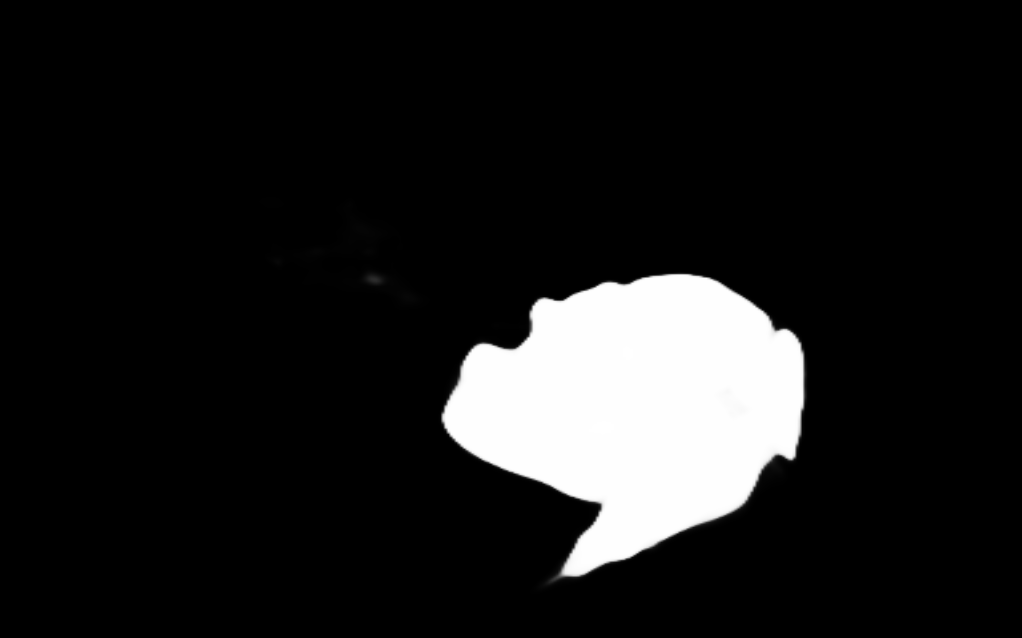} & 
\includegraphics[width=0.118\textwidth]{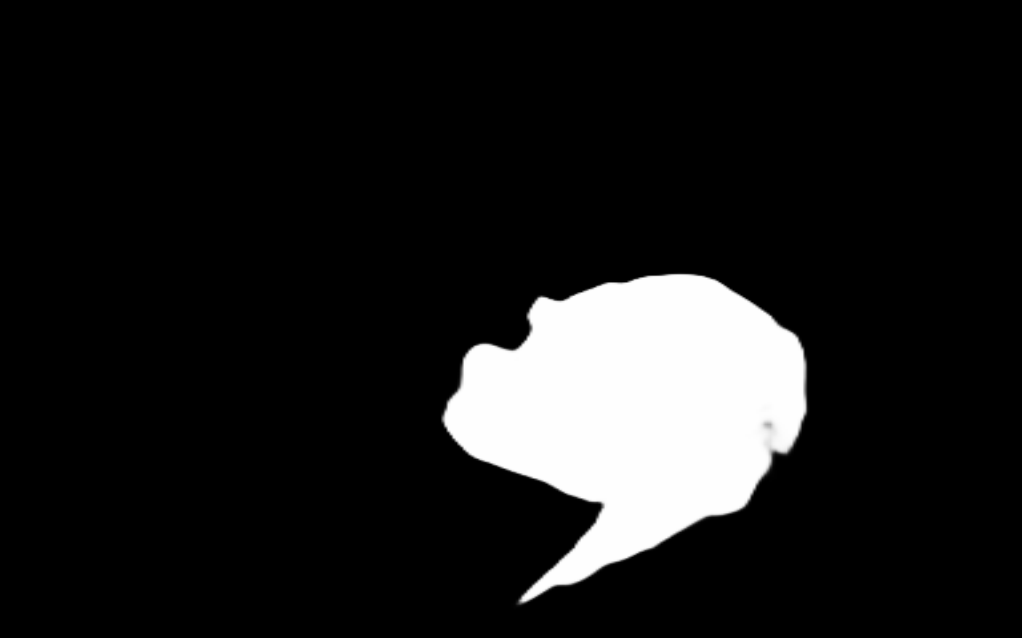} & 
\includegraphics[width=0.118\textwidth]{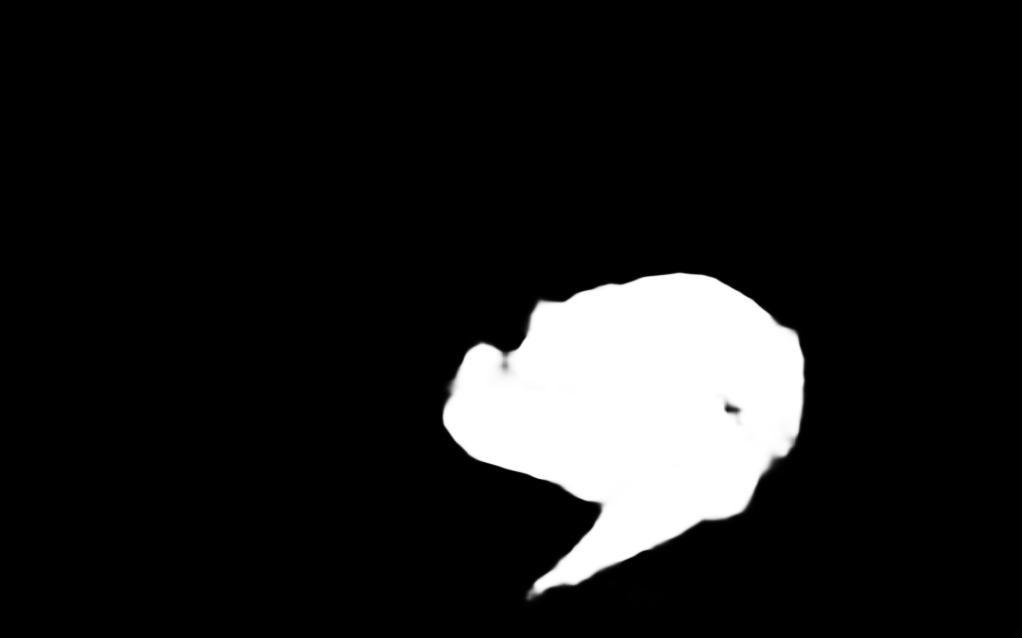} & 
\includegraphics[width=0.118\textwidth]{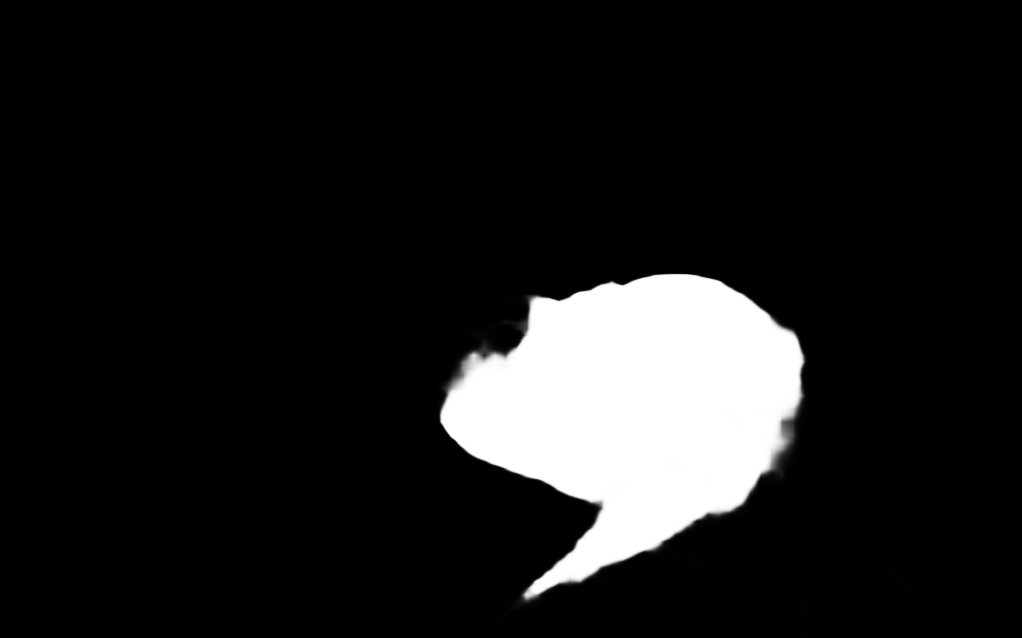} & 
\includegraphics[width=0.118\textwidth]{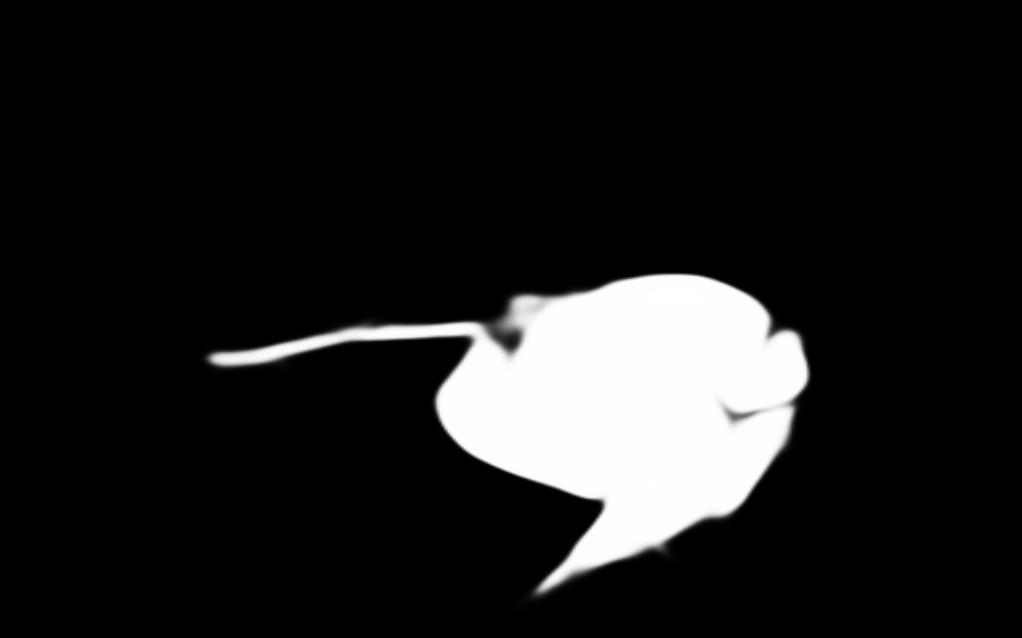} \\

\includegraphics[width=0.118\textwidth]{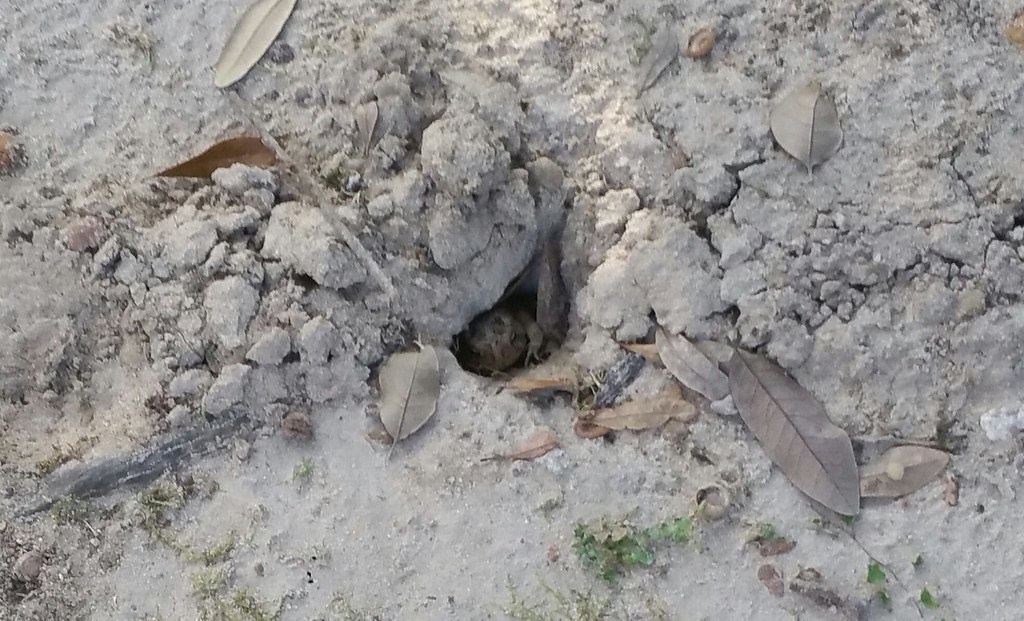} & 
\includegraphics[width=0.118\textwidth]{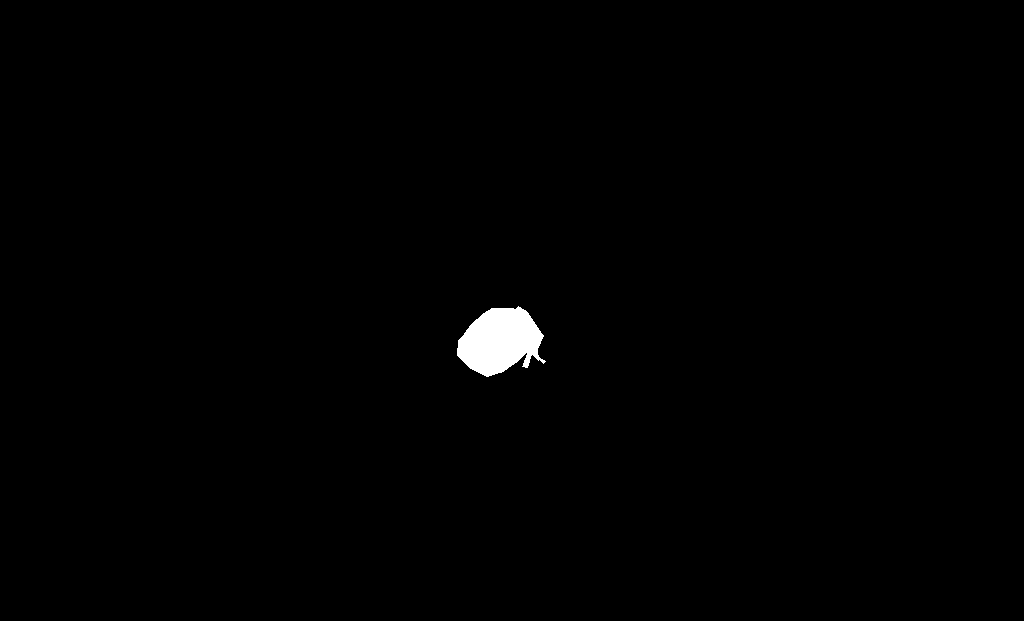} & 
\includegraphics[width=0.118\textwidth]{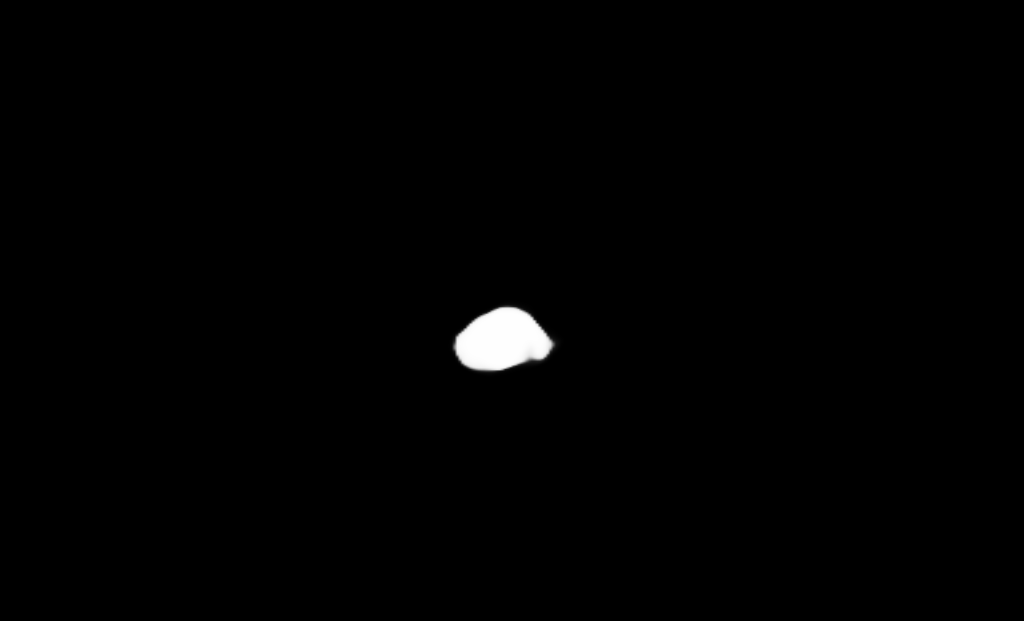} & 
\includegraphics[width=0.118\textwidth]{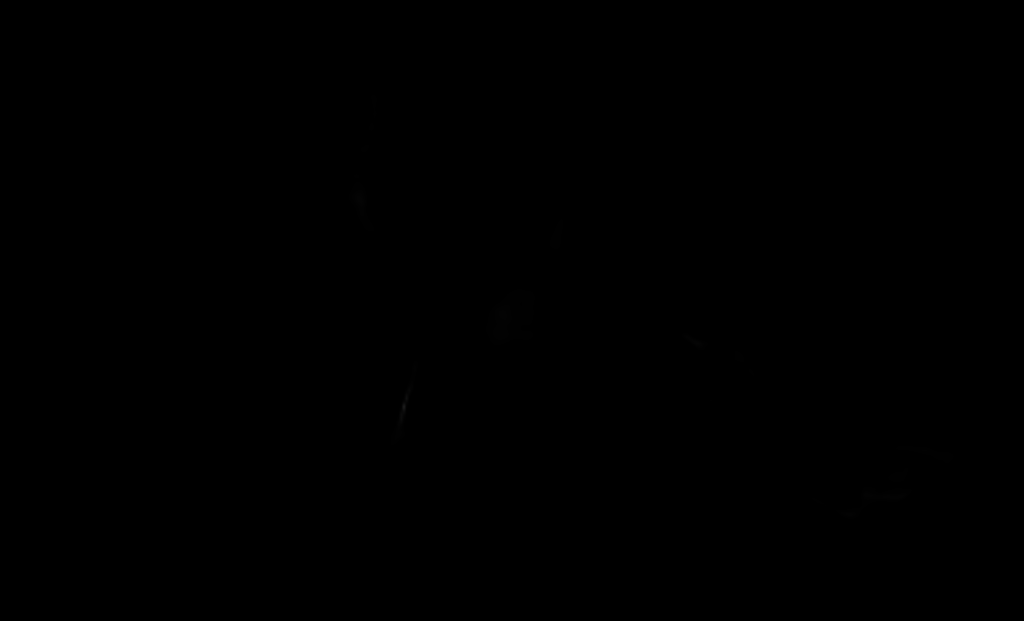} & 
\includegraphics[width=0.118\textwidth]{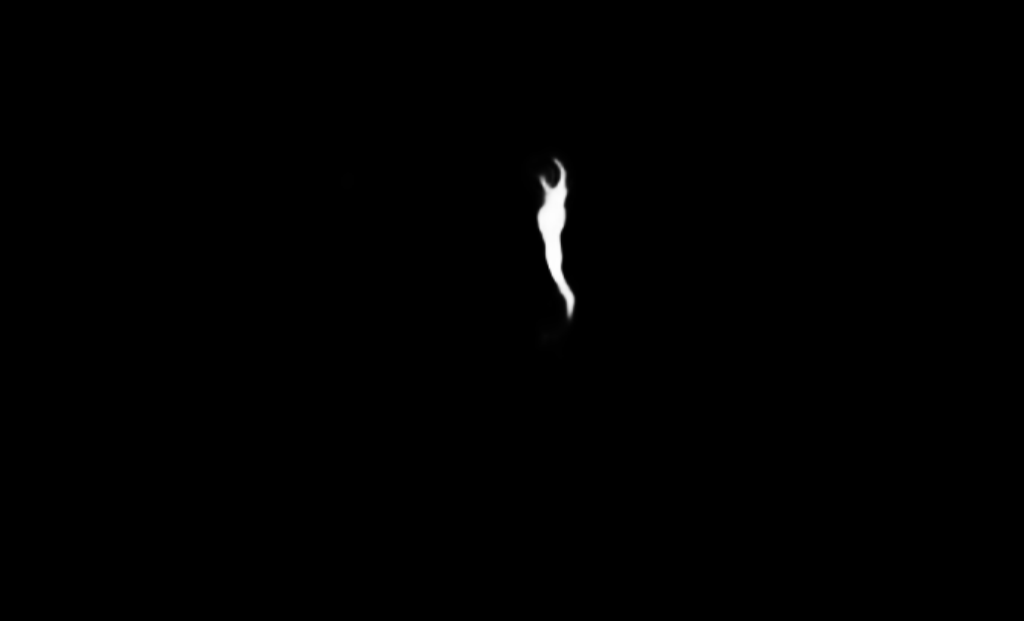} & 
\includegraphics[width=0.118\textwidth]{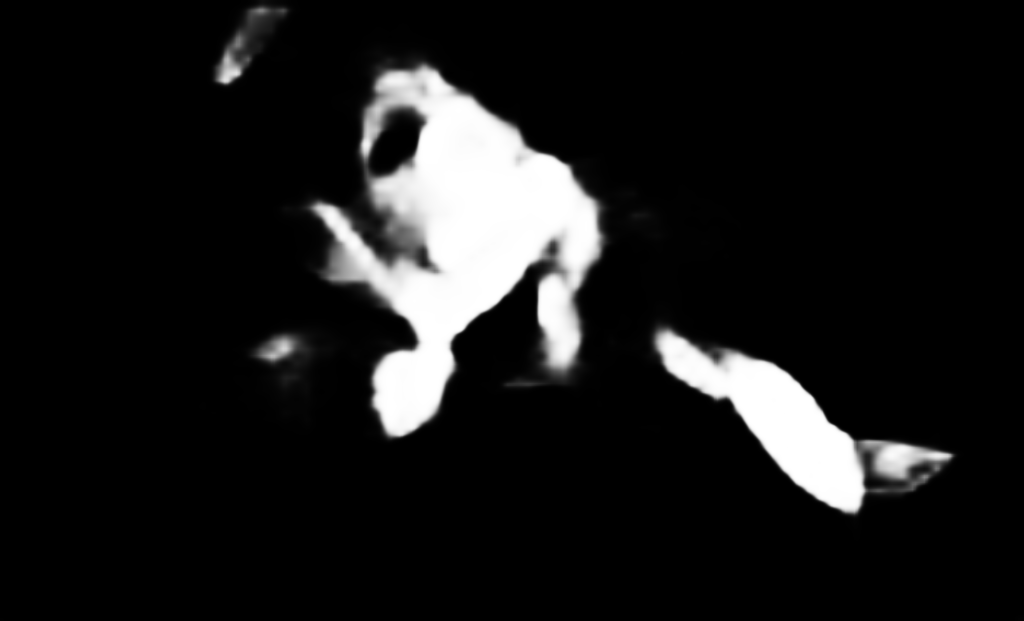} & 
\includegraphics[width=0.118\textwidth]{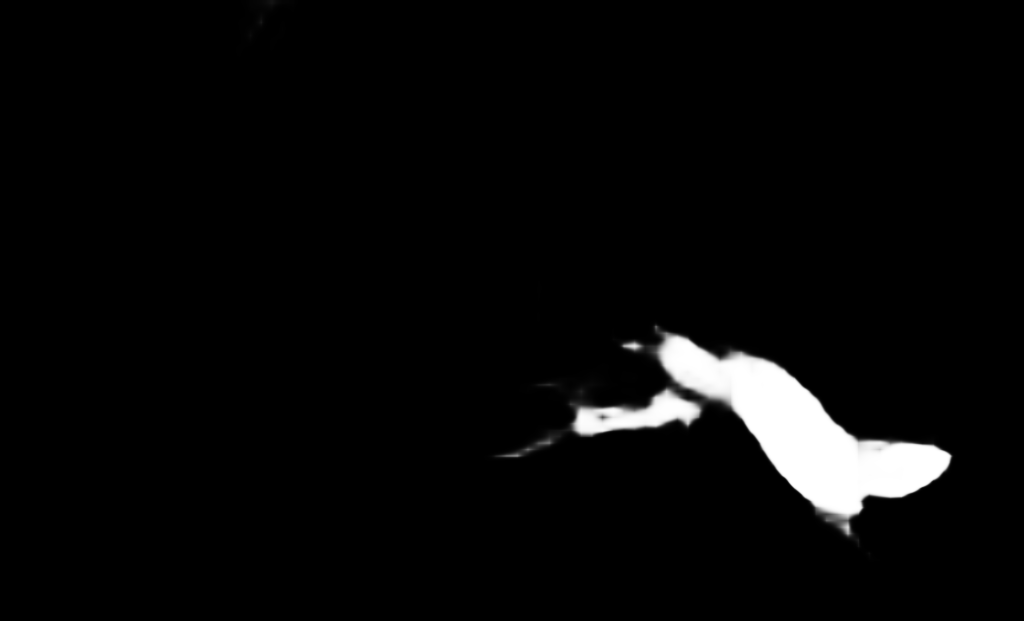} & 
\includegraphics[width=0.118\textwidth]{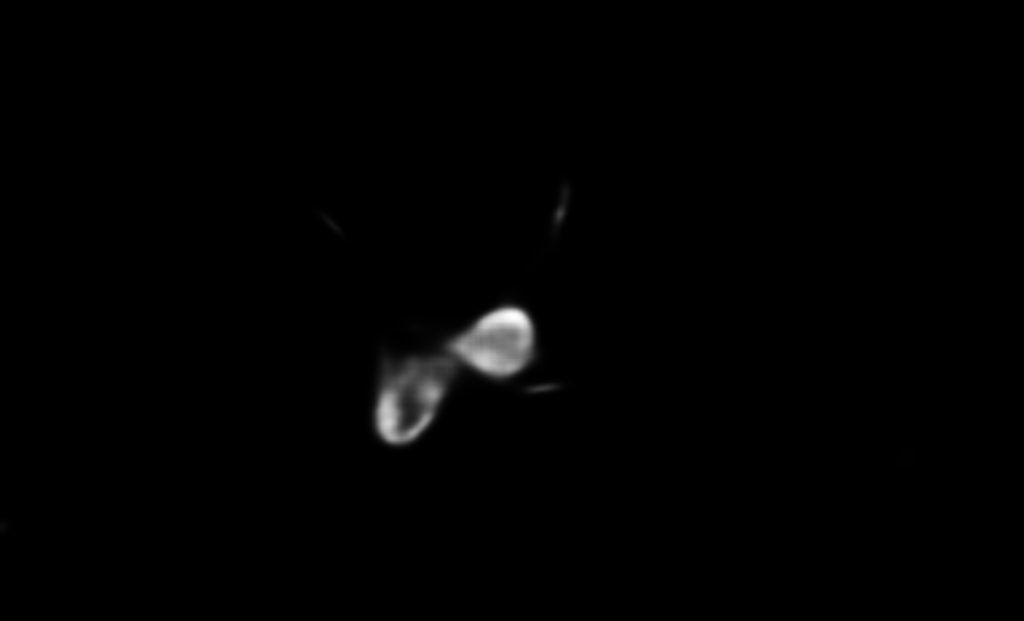} \\

\includegraphics[width=0.118\textwidth]{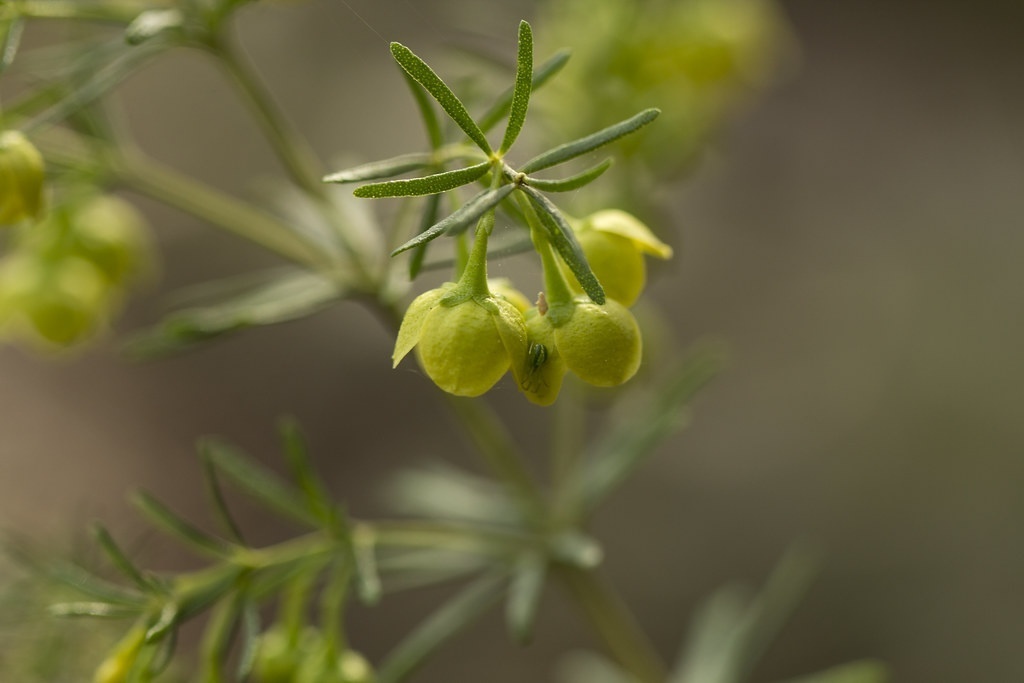} & 
\includegraphics[width=0.118\textwidth]{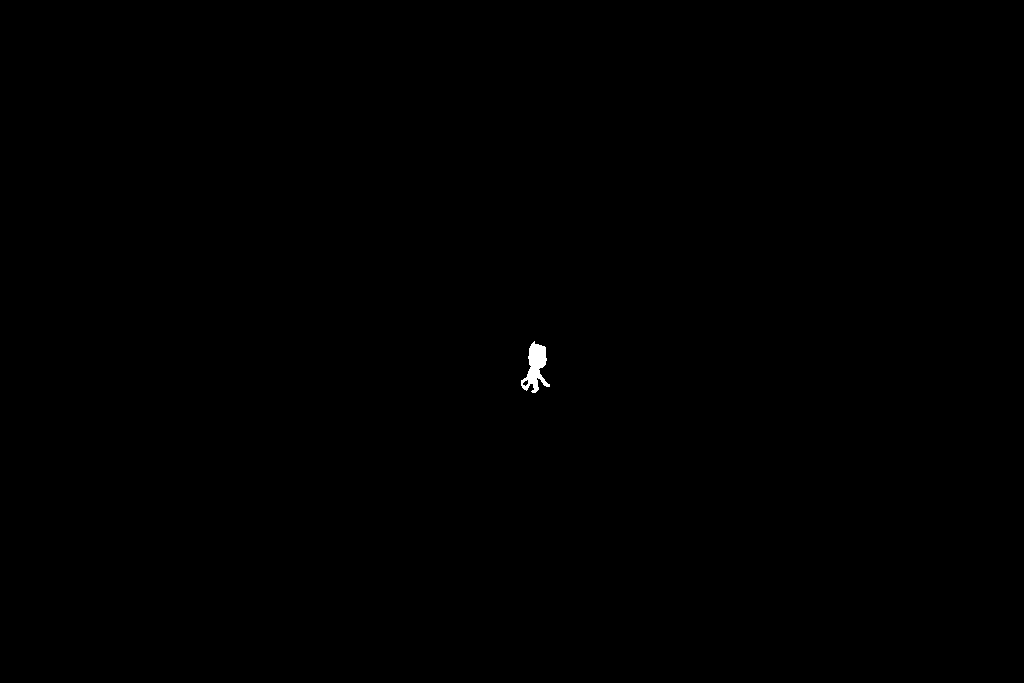} & 
\includegraphics[width=0.118\textwidth]{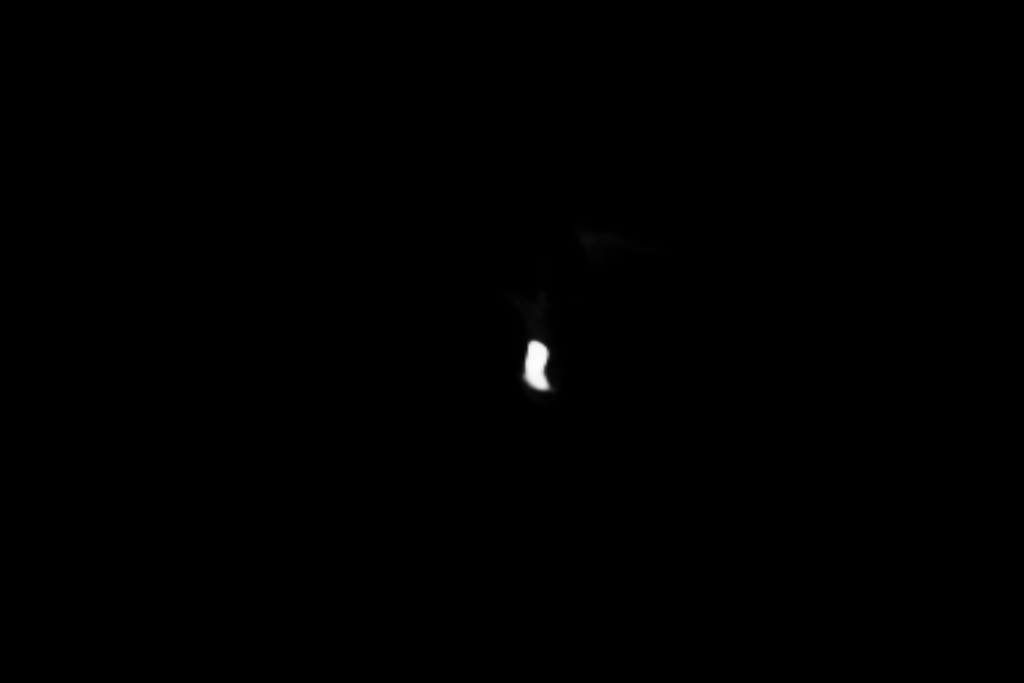} & 
\includegraphics[width=0.118\textwidth]{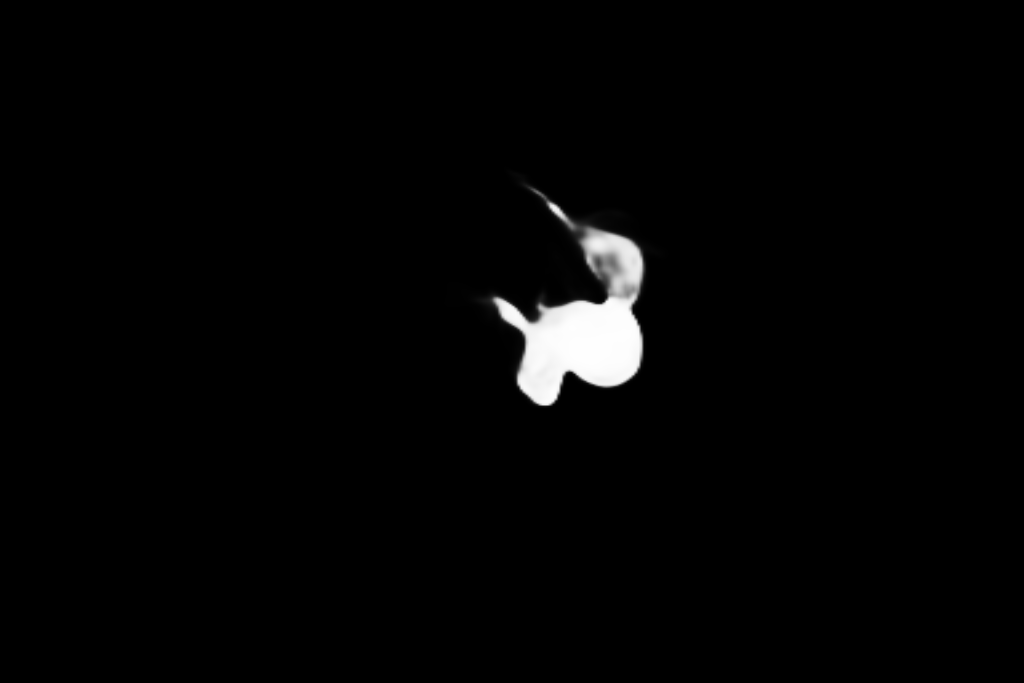} & 
\includegraphics[width=0.118\textwidth]{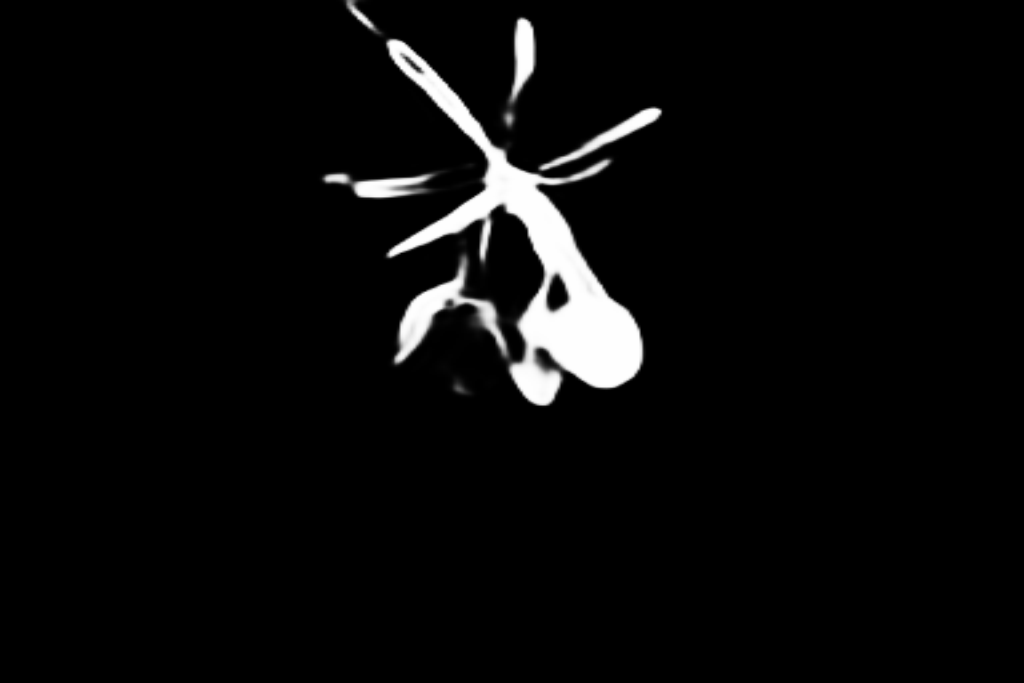} & 
\includegraphics[width=0.118\textwidth]{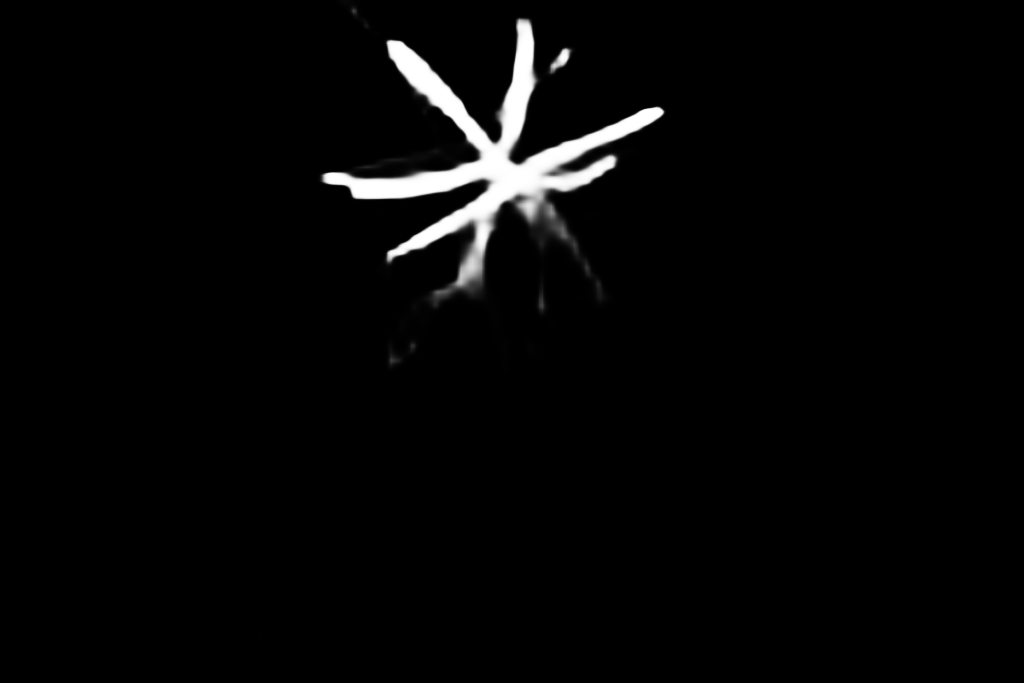} & 
\includegraphics[width=0.118\textwidth]{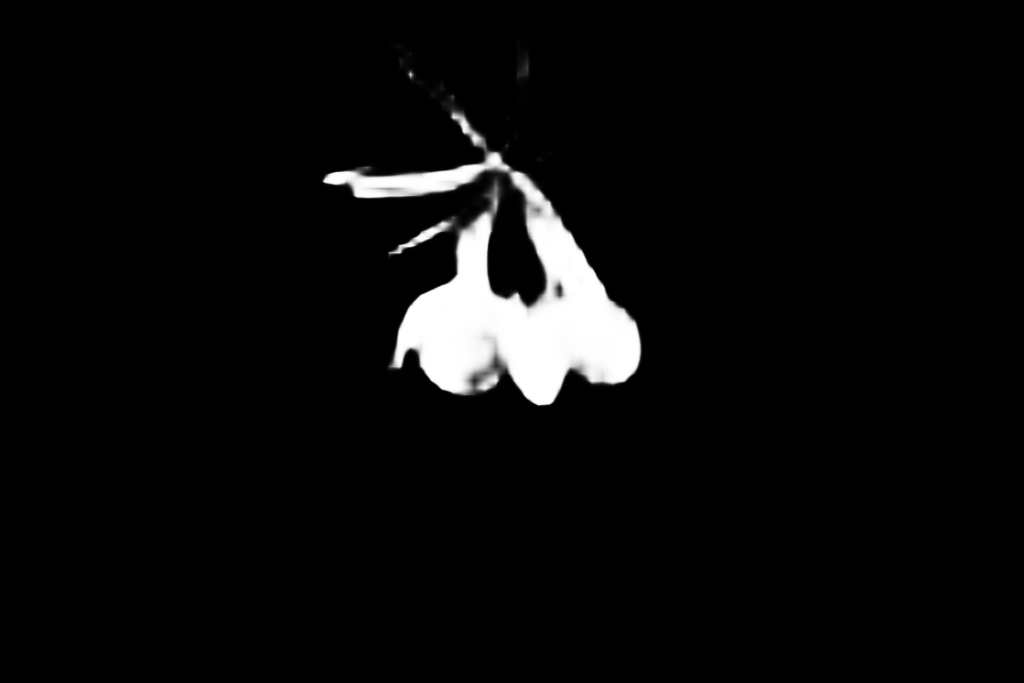} & 
\includegraphics[width=0.118\textwidth]{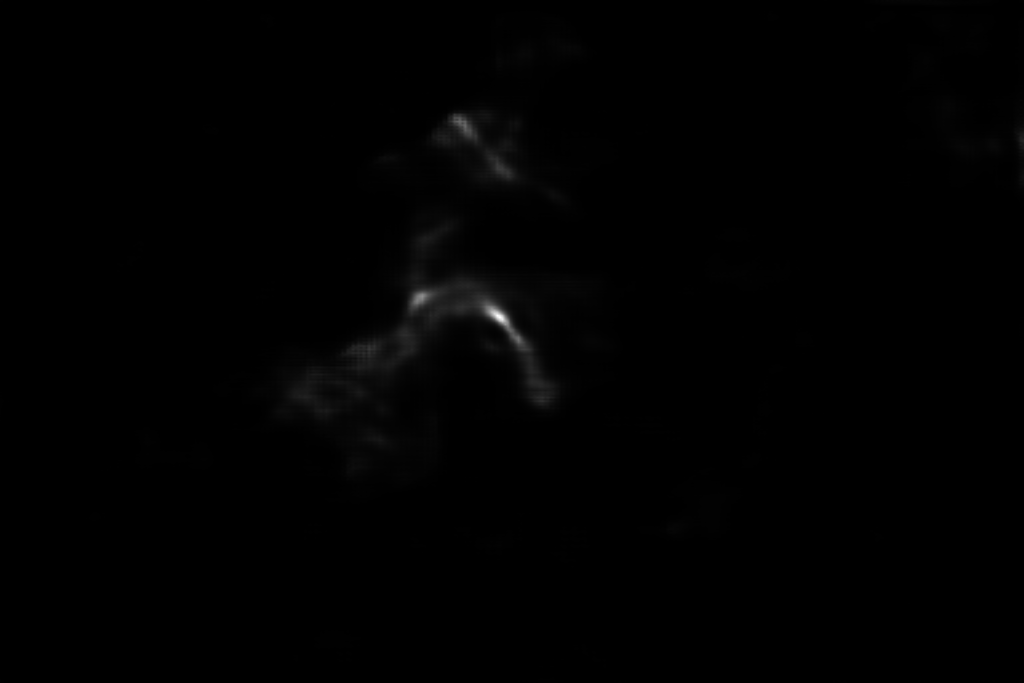} \\

Input & GT & Ours & \cite{pang2024zoomnext} & \cite{xing2023go} & \cite{liu2023mscaf} & \cite{yin2024camoformer} & \cite{huang2023feature}\\ 
\end{tabular}
\caption{Some challenging camouflage scenarios, including: multiple objects (row 1), small objects (row 2), and tiny objects (row 3).}
\label{fig:IntroChallenges}
\vspace{-4mm} \end{figure}

The COD10K~\cite{fan2020camouflaged}, introduced by Fan \etal, is a pioneering COD dataset comprising 5,066 camouflaged images sourced from real-world contexts. Furthermore, the authors established one of the initial COD networks, known as SINet, which employs a dual-module architecture that implements localization followed by object segmentation. Subsequently, an enhanced iteration, SINet-v2~\cite{fan2020camouflaged}, was developed utilizing an optimized decoder and an attention mechanism. In the wake of this development, various advanced deep learning-based networks~\cite{mei2021camouflaged, pang2022zoom,Zhou_2025_ICCV} have emerged to address this challenge. Nonetheless, numerous models continue to struggle to effectively detect camouflaged objects in complex scenarios, particularly in scenes with small or multiple objects, underscoring the need for further improvements. \Cref{fig:IntroChallenges} delineates these challenging camouflage scenarios, which entail detecting various objects within a scene (row 1), small objects (row 2), and tiny objects (row 3).

To address these challenges, we propose MSRNet, an innovative transformer-based network for detecting camouflaged objects that leverages multi-scale feature extraction and recursive feedback to refine features. Our architecture can process images at multiple scales and efficiently extract features using a Pyramid Vision Transformer encoder. This methodology facilitates understanding of the global context from low-resolution feature maps and enables detection of local details from high-resolution feature maps. It effectively addresses the challenges of detecting multiple objects within a scene, including those of various sizes, even tiny ones. Furthermore, our model integrates multi-scale features using Attention-Based Scale Integration Units to selectively incorporate the most relevant ones. A novel recursive feedback decoding strategy is implemented to obtain feedback from lower-resolution feature maps while preserving the global contextual information they contain. Our decoder is equipped with Multi-Granularity Fusion Units that enhance feature representations, enabling more precise detection. By jointly leveraging multi-scale learning with large input scales and applying recursive-feedback feature optimization, MSRNet effectively captures local and global features. This enables the detection of small and multiple camouflaged objects, thereby addressing the challenges illustrated in \cref{fig:IntroChallenges}.

\noindent\textbf{Contributions:} We claim the following contributions.
\begin{itemize}
    \item We propose MSRNet, a multi-scale recursive network that leverages multi-scale feature extraction and recursive feature refinement to improve the detection of small, tiny, and multiple camouflaged objects.
     \item We introduce a novel recursive-feedback decoding strategy that propagates global contextual information from lower-resolution feature maps to all subsequent higher-resolution feature maps, thereby strengthening the model's ability to detect multiple objects within a scene.
     \item We conduct extensive experiments on four benchmark COD datasets, comparing our model with 20 SOTA methods. The results show that MSRNet achieves SOTA performance on two datasets and ranks second on the other two.
\end{itemize}

\section{Related Works}
Early works on COD~\cite{beiderman2010optical, galun2003texture, guo2008robust, hall2013camouflage, zhang2016bayesian} relied on manually crafted features to distinguish camouflaged objects from their backgrounds. While these methods performed well in simple scenes where objects were somewhat visible, they struggled in complex scenes where objects were nearly invisible or occluded. This limitation arose from their restricted feature representation. Therefore, studies~\cite{cheng2023large, luo2023camdiff, luo2024vscode, zhang2023referring,Ye_2025_ICCV} began to focus on incorporating deep learning-based methods that automatically learn features during training. This advancement enabled the learning of more robust features, significantly improving segmentation accuracy in such challenging scenes.

\noindent\textbf{CNN-based methods.} Recently, researchers have addressed the COD task by building CNN-based models. Their methods can be categorized into three main approaches: i) The Multi-scale feature aggregation approach~\cite{sun2021context, zhuge2022cubenet, zhu2022can, pang2022zoom, ji2023deep}, which focuses on merging features from different scales or resolutions to capture more details. Following this approach, C2FNet~\cite{sun2021context} employed an attention-induced cross-level module for feature fusion and a dual-branch module to generate multi-scale representations that leverage global context. CubeNet~\cite{zhuge2022cubenet} employed square fusion decoders to enhance feature representations and a sub-edge decoder to improve object boundary modeling. BSA-Net~\cite{zhu2022can} enhanced boundary understanding by utilizing a separate attention mechanism. ZoomNet~\cite{pang2022zoom} comprises two modules: one for extracting and merging scale-specific features, and the other for identifying mixed-scale features. DGNet~\cite{ji2023deep} concentrated on separately extracting context and texture features before aggregating them to enhance the detection process. ii) The Multi-stage approach~\cite{fan2020camouflaged, fan2021concealed, mei2021camouflaged, yang2021uncertainty, jia2022segment, zhang2022preynet}, which breaks the COD task into multiple focused stages, improves the model's ability to manage the task's complexity. Following this approach, SINet~\cite{fan2020camouflaged} and SINetV2~\cite{fan2021concealed} focused on searching for and identifying camouflaged objects. PFNet~\cite{mei2021camouflaged} applied a positioning process to detect objects and a focusing process to refine predictions. UGTR~\cite{yang2021uncertainty} produced initial predictions and refined them leveraging attention mechanisms. SegMaR~\cite{jia2022segment} employed an iterative refinement strategy incorporating segmentation, magnification, and reiteration processes. The PreyNet~\cite{zhang2022preynet} framework consisted of two stages: initial detection and predator learning. iii) The Joint training approach~\cite{lv2021simultaneously, zhai2021mutual, li2021uncertainty, sun2022boundary, he2023camouflaged}, which involves training the model on several related tasks to enhance its robustness by allowing it to learn from diverse information sources. Adopting this approach, SLSR~\cite{lv2021simultaneously} executed localization, segmentation, and ranking of camouflaged objects. MGL-R~\cite{zhai2021mutual} performed object and boundary localization, leveraging mutual learning through graph-based reasoning. UJSC~\cite{li2021uncertainty} conducted salient and camouflaged object detection simultaneously, utilizing the contradictory information of both tasks. BGNet~\cite{sun2022boundary} integrated edge semantics to enhance object detection and boundary localization tasks. FEDER~\cite{he2023camouflaged} simultaneously tackled COD and edge reconstruction.

\noindent\textbf{Transformers-based methods.} Transformers have demonstrated their capability to encode global contextual information more effectively than CNNs. Consequently, they have been extensively utilized in various computer vision tasks, including image classification~\cite{dosovitskiy2020image, touvron2021training, wang2021pyramid}, image segmentation~\cite{jiang2021all, xie2021segformer}, object detection~\cite{carion2020end}, and salient object detection~\cite{gu2020pyramid, liu2021visual, zhuge2022salient}. Therefore, Transformer-based models have become the standard for building COD models, with the aim of improving this task. 

\begin{figure}[tb]
  \centering
  \includegraphics[height=4.5cm]{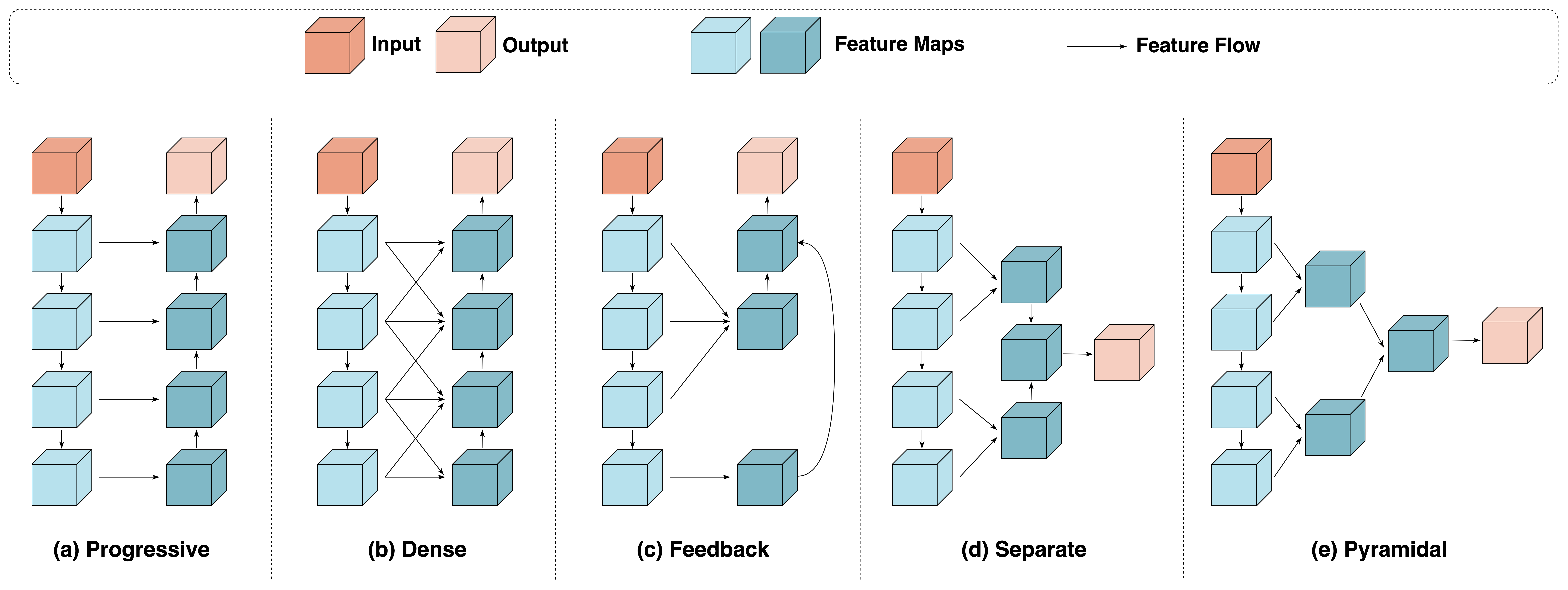}
  \caption{The five decoding strategies in the literature: \textbf{(a)} The progressive decoding strategy, \textbf{(b)} The dense decoding strategy, \textbf{(c)} The feedback decoding strategy, \textbf{(d)} The separate decoding strategy, and \textbf{(e)} The pyramidal decoding strategy.}
  \label{fig:LR_Decoding}
\end{figure}
In addition to the previously introduced main approaches, transformer-based COD methods can also be categorized by their decoding strategy. \Cref{fig:LR_Decoding} demonstrates the five decoding strategies utilized in the literature and includes (a) the progressive decoding strategy, where features are progressively refined and decoded from the lowest-resolution features to the highest-resolution features. (b) The dense decoding strategy, where all features from adjacent resolutions are aggregated. (c) The feedback decoding strategy treats the lowest-resolution features separately and aggregates them with the output to enhance learning of global information. Alternatively, feedback can be taken from the highest-resolution features when a greater focus on local features is required. (d) The separate decoding strategy processes higher-resolution features and lower-resolution features separately to emphasize local and global information equally. Finally, (e) the pyramidal decoding strategy, where adjacent features are aggregated and decoded layer by layer in a progressive manner.

Following a multi-stage approach to address the COD task, MSCAF-Net~\cite{liu2023mscaf} extracted multi-resolution features, introduced a module to enhance resolution-specific features, and then employed progressive decoding to fuse them. SARNet~\cite{xing2023go} built a three-stage architecture that extracted multi-resolution features, applied adjacent- and cross-resolution feature fusion, and finally enhanced the features with background and foreground attention. HitNet~\cite{hu2023high} used an iterative feedback mechanism to refine feature representations across different resolutions. FSPNet~\cite{huang2023feature} progressively enhanced and decoded multi-resolution features with pyramidal shrinking. TPRNet~\cite{zhang2023tprnet} treated features separately, where progressive refinement was applied to high-resolution features and feature interactions were used for low-resolution features to improve the detection process. On the other hand, by implementing a joint training approach, Liu~\etal~\cite{liu2022boosting} built DTINet, an interactive transformer that detects camouflaged objects and their boundaries utilizing multi-head self-attention. Following the multi-scale feature aggregation approach, HDPNet~\cite{he2025hdpnet} employed an hourglass transformer encoder for feature extraction and focused on enhancing local-global feature interactions.

\begin{table}[t]
  \centering
  \caption{Summary of Transformer-based COD methods.}
  \label{tab:Transformer-based-Summary}
  \resizebox{1.0\textwidth}{!}
  {
  \begin{tabular}{c|c|p{8cm}|c}
    \hline
      \textbf{Approach Type} & \textbf{Models} & \multicolumn{1}{c}{\textbf{Primary features}} & \textbf{Decoding Strategy} \\
\hline      
Multi-Scale Feature Aggregation 
& ZoomNeXt & Merging multi-scale features with attention, then progressively enhancing and decoding them & Progressive \\
\hline

& MSCAF-Net & Enhancing resolution-specific features, then applying cross-resolution fusion & Progressive \\
\cline{2-4}

& SARNet & Three-stage architecture: Search (extraction), Amplify (fusion), Recognize (enhancement) & Dense + Feedback \\
\cline{2-4}

& CamoFormer & Progressively enhancing features using foreground, background, and full image attentions & Progressive \\
\cline{2-4}

Multi-Stage Techniques & HitNet & Applying iterative feature refinement with feedback from the high-resolution features to preserve fine details & Feedback \\
\cline{2-4}

& FSPNet & utilizing pyramidal shrinking to encode multi-resolution features & Pyramidal \\
\cline{2-4}

& TPRNet & Applying interactions across low-level features and progressive refinement on high-level features & Separate \\
\hline

Joint Training Approach
& DTINet & Applying COD and boundary detection utilizing multi-head self-attention & Progressive \\
\hline
\end{tabular}}
\end{table}

All models mentioned above have been built upon either CNNs or transformers. Some methods~\cite{yin2024camoformer, pang2024zoomnext} experimented with both backbones. CamoFormer~\cite{yin2024camoformer} was constructed on the multi-stage approach. It used masked separable attention to identify objects and a top-down decoder to progressively refine feature representations. ZoomNeXt~\cite{pang2024zoomnext} adopted a multi-scale feature aggregation approach and built a unified pyramid network for both static and dynamic COD. ZoomNeXt employed a multi-head scale-integration module and a feature-refinement mechanism. Both studies found that transformer-based models consistently outperform CNN-based models. Beyond the supervised COD methods discussed in this paper, some studies have also explored weakly-supervised~\cite{ni2026fcl} and unsupervised~\cite{chen2026beyond} COD. Moreover, a more recent study~\cite{han2026camoquery} has incorporated natural language queries to guide camouflaged object segmentation.

\Cref{tab:Transformer-based-Summary} summarizes the transformer-based COD methods categorized by their main approach and decoding strategy.

\begin{figure}[tb]
\centering
\includegraphics[height=6.5cm]{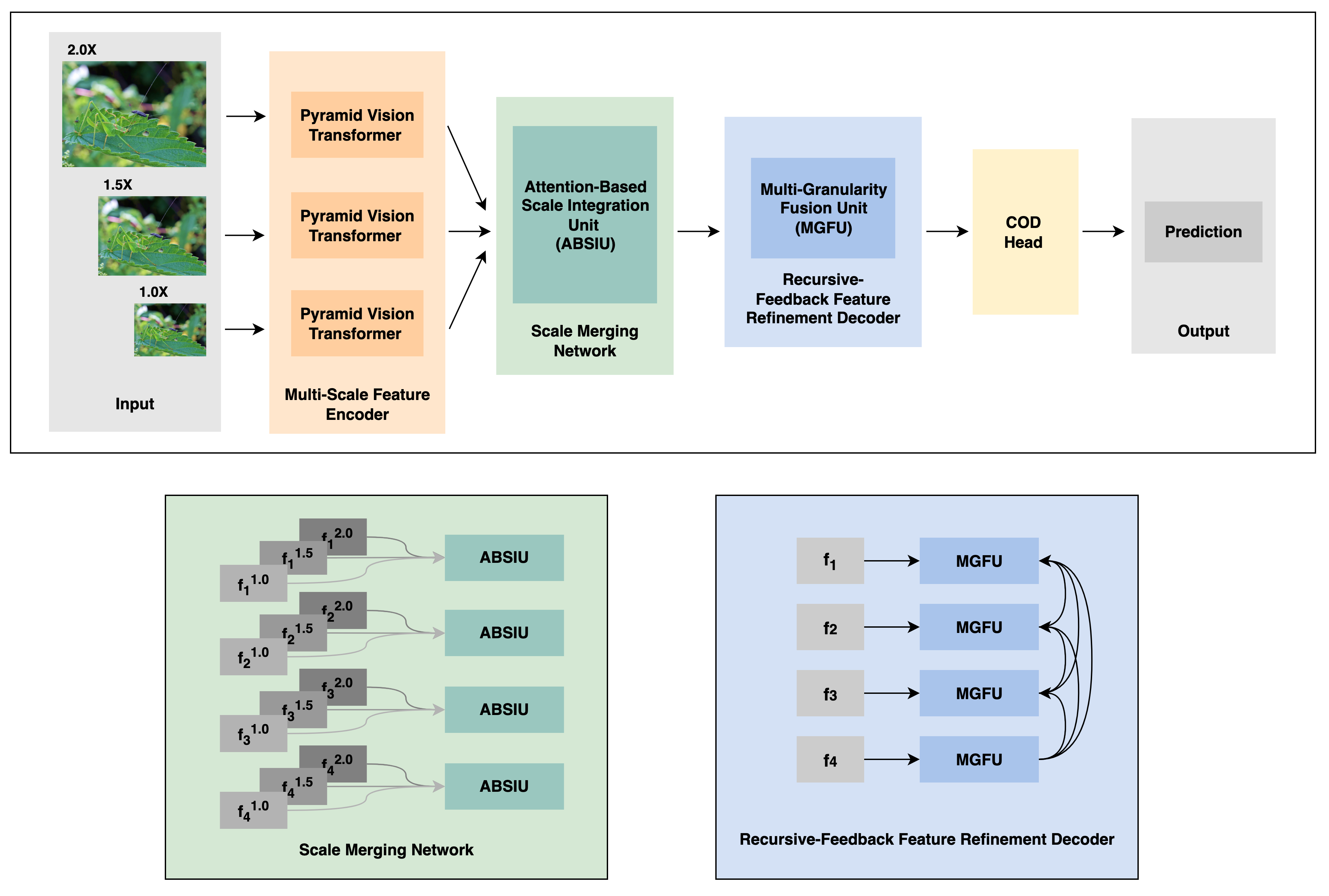}
\caption{The overall architecture of MSRNet consists of three scales of the original image, each of which is input into a PVT for feature extraction, generating four feature maps of different resolutions: $f_{1}$, $f_{2}$, $f_{3}$, and $f_{4}$. In the next stage, the feature maps of the same resolution across all scales are merged by the Attention-Based Scale Integration Unit (ABSIU). Each merged feature map is further refined inside the decoder using the Multi-Granularity Fusion Unit (MGFU). The Recursive-Feedback decoding strategy combines feedback from all lower resolutions with the current-resolution data processed by the MGFU.}
\label{fig:methodology}
\vspace{-4mm} \end{figure}

\section{Methodology}
\label{sec:methodology}
Our model takes a static image as input. This input then passes through several model components to produce a probability map ranging from 0 to 1 that represents the likelihood that a pixel belongs to the camouflaged object. Our approach utilizes an image pyramid that contains multiple scales of the input image. This multi-scale representation enables the extraction of diverse features at each scale, facilitating the detection of camouflaged objects. Our model components include a multi-scale feature encoder, a scale-merging network, and a recursive-feedback feature-refinement decoder. The multi-scale feature encoder extracts features at each scale. The scale-merging network is designed to integrate these features utilizing attention-based scale-integration units (ABSIUs). Moreover, multi-granularity fusion units (MGFUs) within the decoder refine feature representation, enhancing the model's accuracy in detecting camouflaged objects in complex scenes. The following subsections provide more details about the model's components. \Cref{fig:methodology} demonstrates the overall architecture of the model.


\begin{wrapfigure}{r}{0.41\textwidth}
\centering
\vspace{-0.4cm}
\includegraphics[height=4.5cm]{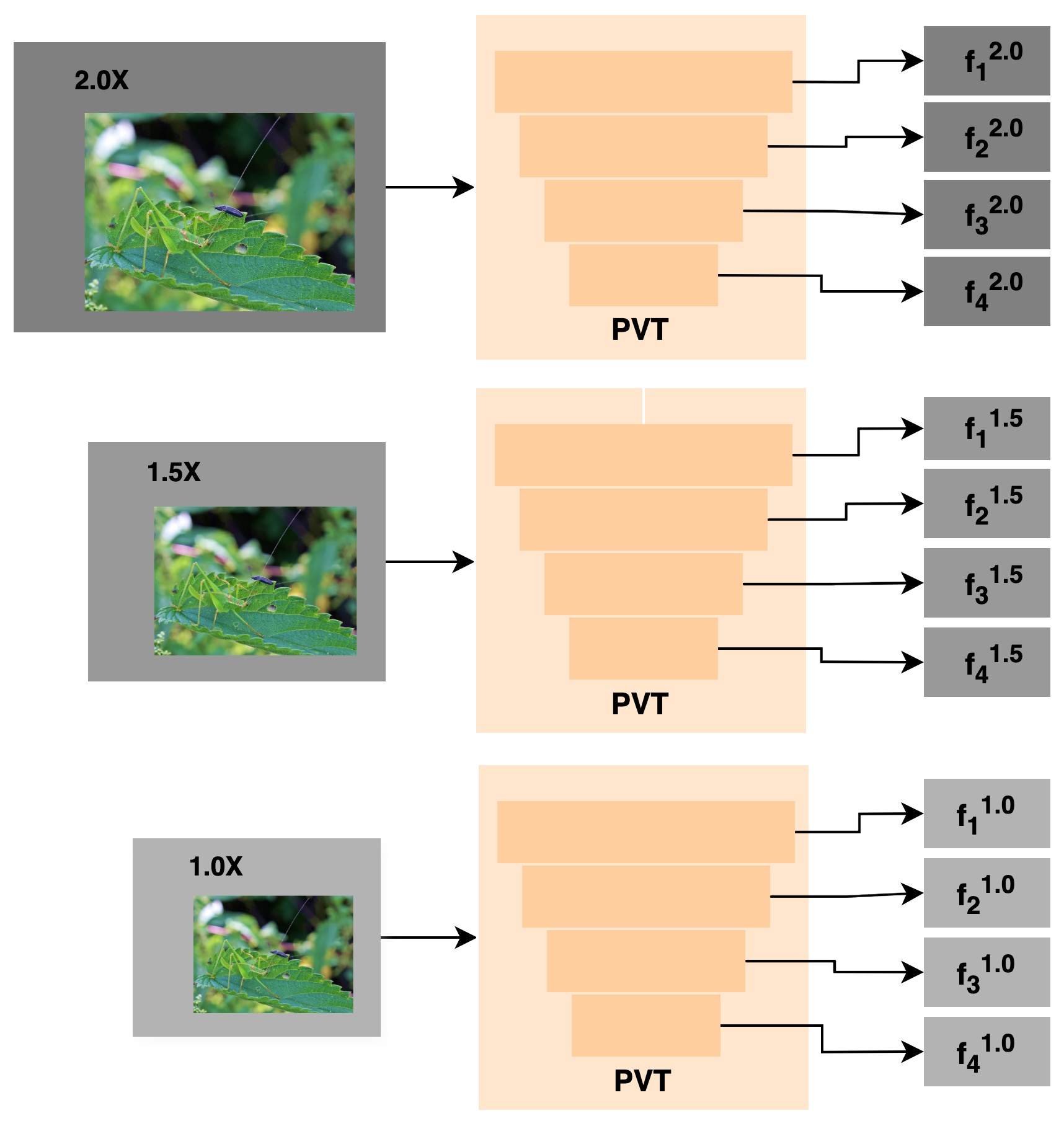}
\vspace{-2mm}
\caption{Feature Extraction Approach}
\vspace{-0.8cm}
\label{fig:feature_extraction_approach} 
\end{wrapfigure}

\noindent\textbf{Multi-Scale Feature Encoder}
\label{subsec: Multi-Scale Feature Encoder}
To extract deep features, we utilize the well-known Pyramid Vision Transformer (PVTv2)~\cite{wang2022pvt} as an encoder, excluding its classification head. Channel dimensionality reduction is applied to all extracted feature maps to enhance computational efficiency for subsequent processing. We extract features from three scales of the input image: the original-size input (1$\times$), the main scale of the input image, and 1.5$\times$ and 2$\times$, the two auxiliary scales. This choice of relatively large input scales enhances the network's ability to learn local features, therefore enabling the detection of small and tiny objects. As shown in \cref{fig:feature_extraction_approach}, this setup generates three feature map sets, each corresponding to an input scale and comprising four feature maps at different resolutions, corresponding to the number of encoder stages. These feature maps are denoted by {$f_{i}^{k}$}, where i ranges from 1 to 4, and k belongs to \{1.0, 1.5, 2.0\}, representing different resolutions and input scales, respectively. In the following stages, these features will be passed to the scale-merging network and then to the recursive-feedback feature-refinement decoder, where they will be integrated and refined.

\noindent\textbf{Scale Merging Network}
\label{subsec: Scale Merging Network}
The scale-merging network integrates features extracted from different input scales using the Attention-Based Scale Integration Unit (ABSIU). This unit employs an attention mechanism to merge features, emphasizing the most significant ones and capturing their relationships. Four ABSIUs are utilized, one for each resolution. For instance, the first ABSIU merges the highest-resolution features across the scales, namely, $f_{4}^{1.0}$, $f_{4}^{1.5}$, and $f_{4}^{2.0}$. This results in merged feature maps for each distinct resolution.

\noindent\textbf{Scale Alignment.} Before integration, the features of the auxiliary scales $f^{1.5}$ and $f^{2.0}$ are resized to align with the main scale feature $f^{1.0}$ by down-sampling them via a combination of max pooling and average pooling.

\begin{figure}[tb]
\centering
\includegraphics[height=6.5cm]{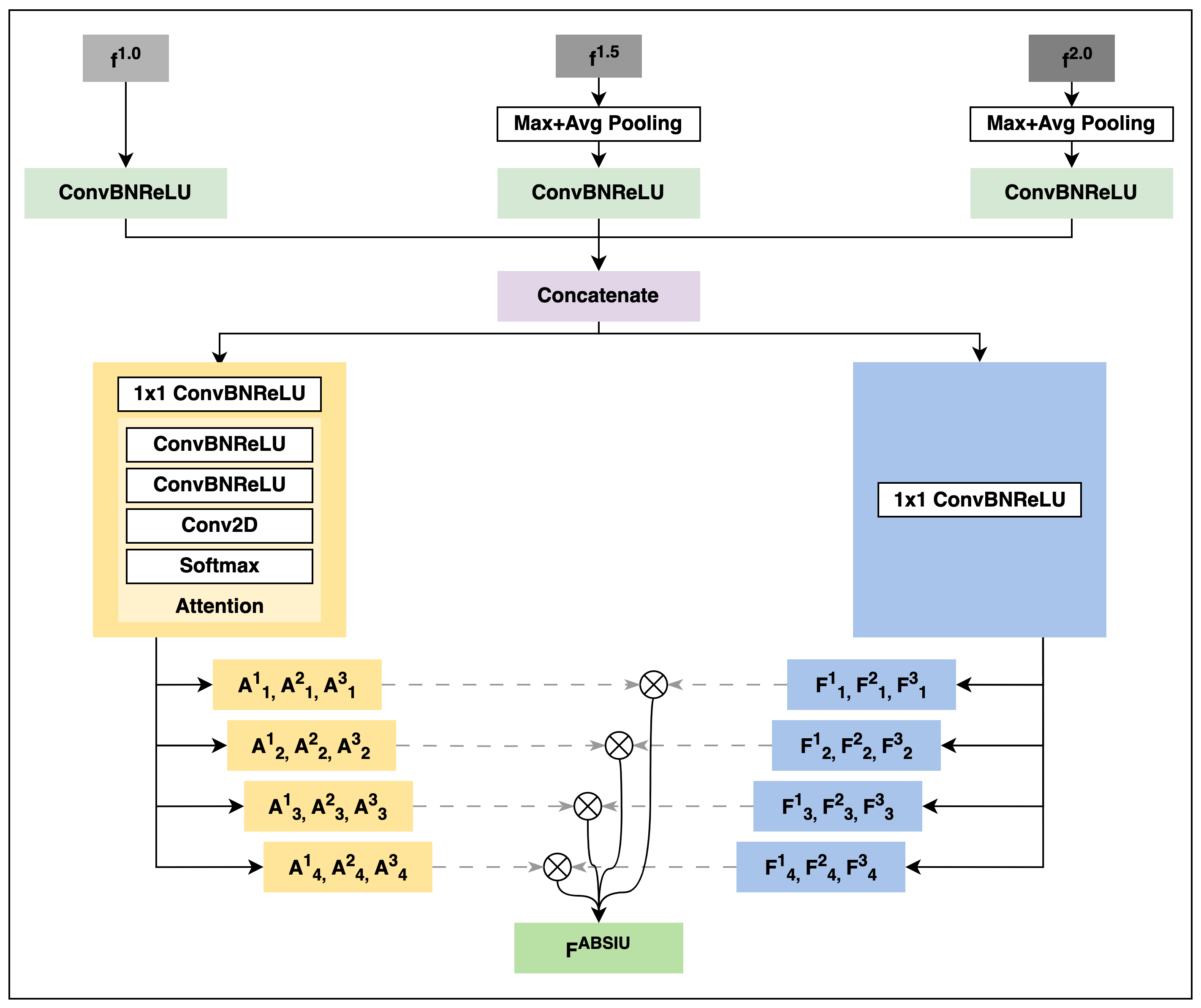}
\caption{The diagram illustrates the Attention-Based Scale Integration Unit (ABSIU) for multi-scale feature integration. Features from the three scales ($f^{1.0}$, $f^{1.5}$, $f^{2.0}$) are first aligned to a common resolution and concatenated. The attention mechanism then applies a series of convolutional layers followed by a Softmax activation layer to generate three-channel attention maps ($A^{1}_{i}$, $A^{2}_{i}$, $A^{3}_{i}$); each channel corresponds to a different scale. An element-wise multiplication $\otimes$ between the attention maps and their corresponding feature maps ($F^{1}_{i}$, $F^{2}_{i}$, $F^{3}_{i}$) is applied, resulting in three scale-grouped processed feature maps that are then summed to produce multi-scale feature maps. This process is repeated for each attention group, yielding four groups of multi-scale features. Lastly, a summation across groups merges features from all attention groups, producing the final output $F^{ABSIU}$.}
\label{fig:scale merging network}
\vspace{-4mm} \end{figure}

\noindent\textbf{Attention-Based Scale Integration Unit (ABSIU).} This unit is designed to integrate features from multiple scales by adopting a multi-head spatial attention mechanism. Spatial attention enables focusing on critical regions in feature maps while preserving location-specific information, which is essential for segmentation. Furthermore, the use of the multi-head attention mechanism enables the model to learn diverse attention patterns. This unit begins by independently processing each feature map from every scale to enhance scale-specific information. The processed feature maps are then concatenated along the channel dimension and passed through a 1$\times$1 convolutional layer to transform them into a common space, preparing them for subsequent processing. To generate the multi-head attention maps, the concatenated features are initially divided into four groups, each containing feature maps from all three scales. Each attention head processes a group by applying a series of convolutional layers, followed by a Softmax activation, yielding a three-channel attention map, one for each scale. 

The generated attention maps are applied to another copy of the concatenated features to produce the final fused multi-scale output. The concatenated features are first divided into four groups, as in the attention-generating step, ensuring that each attention map corresponds to its respective scale features. After alignment, an element-wise product is computed between each attention map and its corresponding scale features. This results in three groups of processed feature maps, one for each scale. These three scale-grouped feature maps are summed to produce multi-scale feature maps. This process is repeated for each attention group, yielding four groups of multi-scale features. Lastly, a summation across groups merges features from all attention groups. \Cref{fig:scale merging network} illustrates how this unit operates.

\begin{figure}[tb]
\centering
\includegraphics[height=4.5cm]{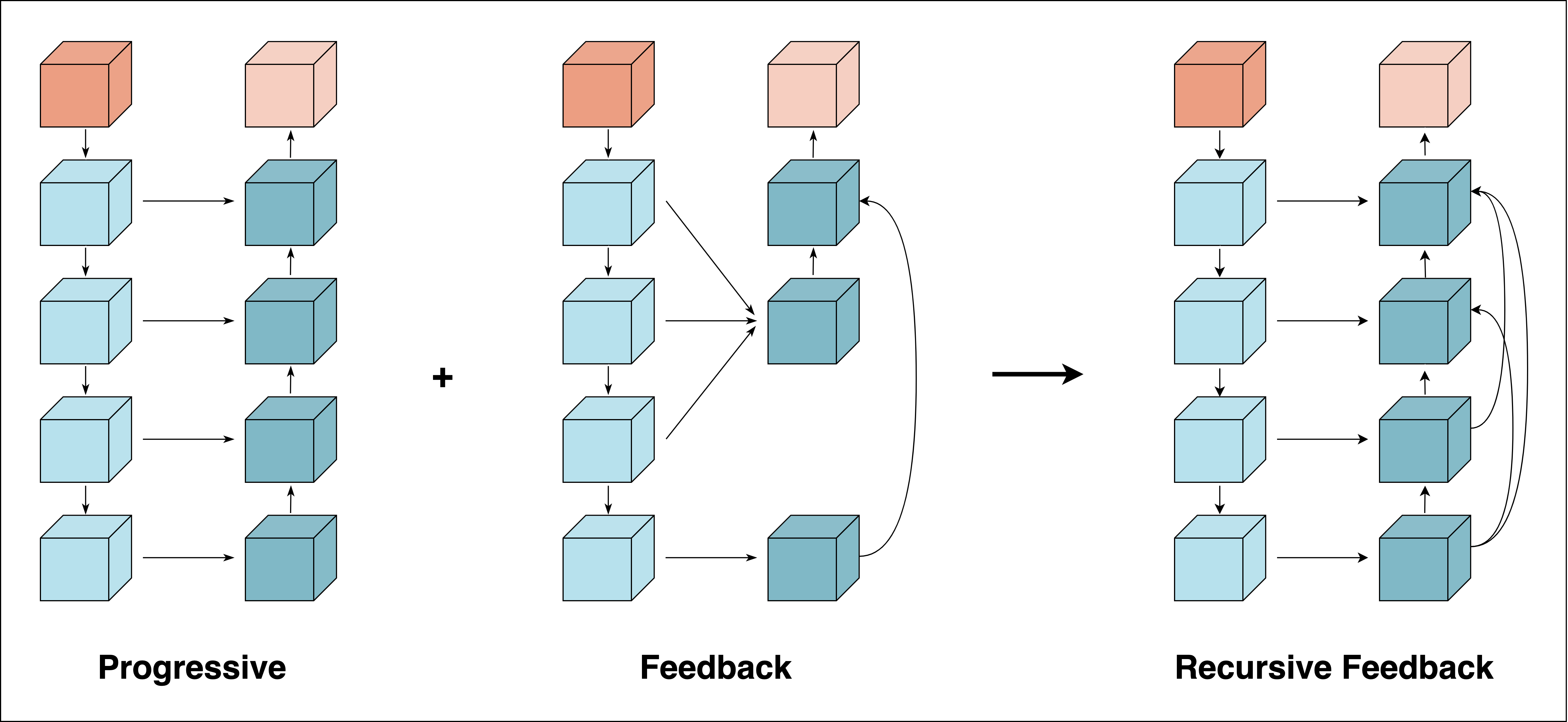}
\caption{This illustrates the proposed novel decoding strategy. This strategy combines progressive decoding and feedback decoding. The feedback decoding strategy is an advanced recursive mechanism that applies feedback from lower-resolution feature maps to all subsequent higher-resolution feature maps, enhancing the network's contextual learning.}
\label{fig:new_decoding_stratgy}

\end{figure}

\noindent\textbf{Recursive-Feedback Feature Refinement Decoder.}
\label{subsec:Recursive-Feedback Feature Refinement Decoder}
In this research, we employ a novel decoding strategy to enhance the global context learning in our network, thus enabling it to detect multiple objects in a scene. The new decoding strategy combines two known strategies: progressive decoding and an advanced version of the feedback strategy. The decoder progressively combines the multi-resolution feature maps from the lowest to the highest resolution. The advanced feedback decoding strategy is a recursive feedback mechanism that takes feedback from lower-resolution feature maps and propagates it to all subsequent higher-resolution feature maps. This recursively preserves global information from lower-resolution feature maps, enabling strong contextual learning within the network. The proposed novel decoding strategy is illustrated in \cref{fig:new_decoding_stratgy}.

The decoder is responsible for decoding features and refining their representation. In the proposed network, the feature representation refinement process is performed by the Multi-Granularity Fusion Unit (MGFU) within the decoder. As shown in \cref{fig:methodology}, this unit combines the multi-scale feature maps generated by the ABSIUs and all outputs of previous MGFUs while enhancing their representation. After aggregating all features in the last MGFU, they pass through a COD head to generate the final prediction map. This head applies up-sampling to restore the original spatial resolution, a 3$\times$3 convolutional layer to reduce the number of channels and refine features, and a 1$\times$1 convolutional layer to compress the channels into a single-channel feature map. This feature map contains raw model logits. A sigmoid activation function is applied to normalize these predictions to the [0, 1] range, representing the probability that each pixel belongs to the camouflaged object.

\noindent\textit{Multi-Granularity Fusion Unit (MGFU).} This unit is designed to enhance feature representations by analyzing and integrating features across multiple granularities. It processes features in groups and applies cross-channel interactions. This unit begins by expanding the feature space with a 1$\times$1 convolutional layer, thereby increasing the number of channels. Subsequently, the expanded features are partitioned into six-channel groups $\{g_{j}\}_{j=1}^{6}$, enabling specialized processing across contexts and facilitating the learning of distinct feature representations. 

A series of convolutional layers is employed to facilitate feature interactions across the various groups. The first group's features are processed directly through a convolutional layer to extract fundamental features. The output is partitioned into three parts: one designated for concatenation with the subsequent group $g^{1}_{1}$ (to propagate information), one for computing a gate value that weighs the significance of the features $g^{2}_{1}$, and one representing the features of this group $g^{3}_{1}$. In the intermediate groups, features from the current group are concatenated with those from the preceding group, then processed by a convolutional layer, enabling the model to learn more complex features. Each intermediate group's output is split into three parts, similar to the first group. The last group processes its features similarly, but its output is divided into two parts because there is no subsequent group.

\begin{figure}[tb]
\centering
\includegraphics[height=6.5cm]{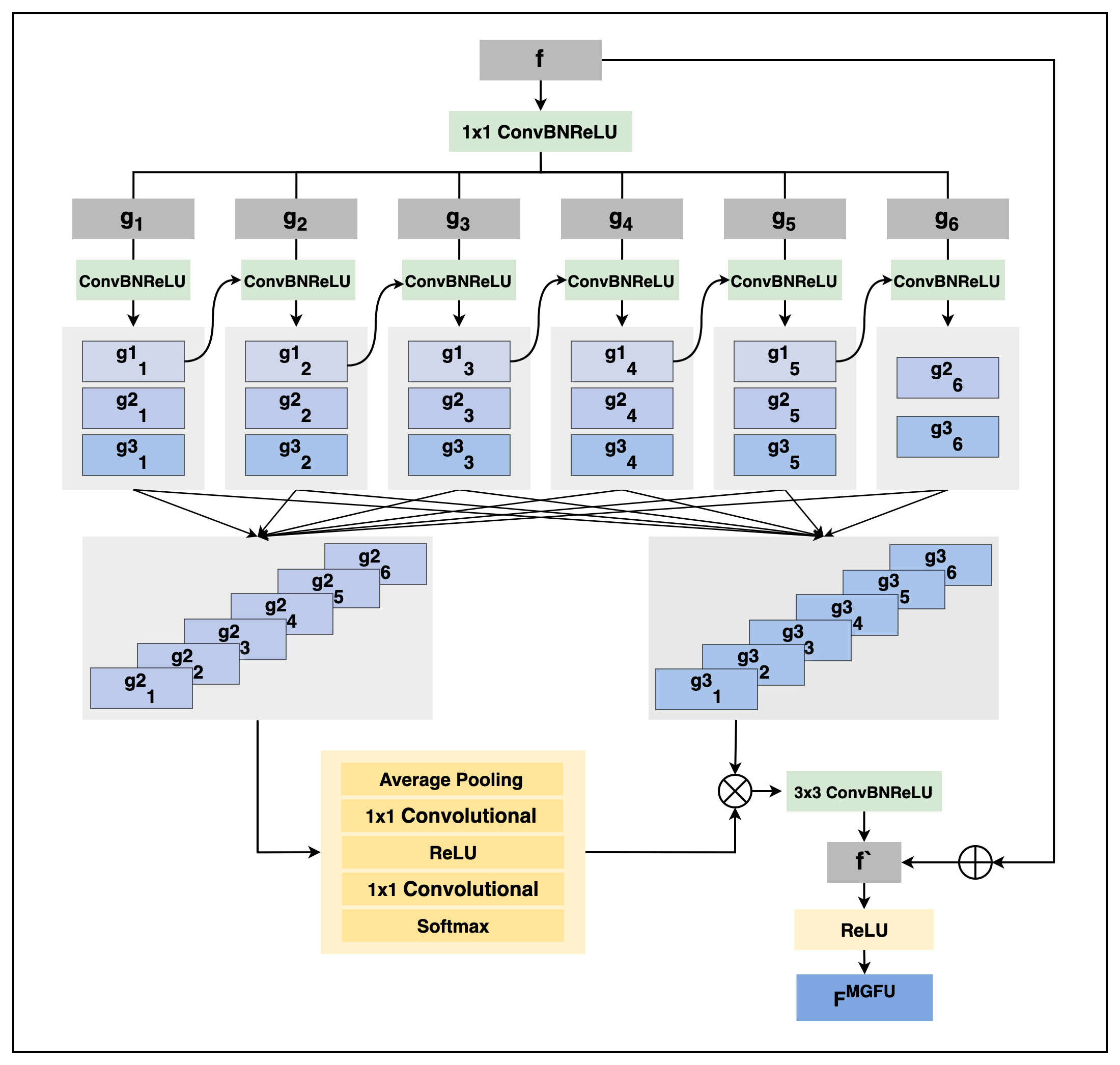}
\caption{Demonstration of all processes in the Multi-Granularity Fusion Unit (MGFU). The MGFU module enhances feature representations by analyzing and integrating features from multiple granularities. It processes features across groups with cross-channel interaction, then adaptively fuses them.}
\label{fig:MGFU}
\vspace{-4mm} \end{figure}

After processing all groups, the concatenated gate features $\{g^{2}_{j}\}_{j=1}^{6}$ are passed through a gating mechanism to generate channel-wise attention maps that highlight the most salient channels based on their global context. This is achieved by sequentially applying spatial and channel compression, a non-linearity, channel expansion, and normalization. The produced attention maps are then multiplied by another set of concatenated features $\{g^{3}_{j}\}_{j=1}^{6}$, yielding reweighted feature maps that enable the model to focus on the most relevant features. Ultimately, the output is refined, combined with the original input for residual learning, and then passed through a ReLU activation function. This process preserves essential information from the original input while ensuring non-linearity in the final output, as shown in \cref{fig:MGFU}. 

\noindent\textbf{Loss Function.}
\label{subsec:Loss Function}
The binary cross-entropy loss (BCE) is commonly used in binary image segmentation tasks. The BCE loss for a pixel at position (i,j) is defined as: 

\begin{equation}
\ell^{i,j}_{BCE} = -g_{i,j} \log p_{i,j} - (1 - g_{i,j}) \log (1 - p_{i,j}).
\end{equation}
where \( g_{i,j} \in \{0, 1\} \) represents the ground truth and \( p_{i,j} \in [0, 1] \) denotes the predicted value. However, relying solely on BCE during training can lead to ambiguous, uncertain predictions due to the task's inherent complexity. To address this, we use an additional loss, the Uncertainty Awareness Loss (UAL)~\cite{pang2024zoomnext}, which enhances model confidence by penalizing predictions with high uncertainty. The UAL is expressed as:
\begin{equation}
\ell^{i,j}_{\text{UAL}} =  1 - \|2p_{i,j} - 1\|^2 
\end{equation}
The total loss function combines both terms as follows: 
\begin{equation}
\mathcal{L} = \mathcal{L}_{\text{BCE}} + \lambda \mathcal{L}_{\text{UAL}},
\end{equation}
where \(\lambda\) is a balancing factor that controls the contribution of the UAL and increases gradually using the cosine strategy. 

\section{Experiments}
\label{sec:experiments}
\subsection{Experiment Setup}
\noindent\textbf{Datasets.}
We utilized four datasets for camouflaged object detection: CAMO~\cite{le2019anabranch}, CHAMELEON~\cite{przemyslaw2017animal}, COD10K~\cite{fan2020camouflaged}, and NC4K~\cite{lv2021simultaneously}. While a subset of these datasets includes images intended for various tasks, we focused specifically on the COD images. We employed a total of 10,513 images from the following datasets: 1,250 from CAMO, 76 from CHAMELEON, 5,066 from COD10K, and 4,121 from NC4K. Consistent with benchmark practices~\cite{pang2024zoomnext, ji2023deep, pang2022zoom, fan2020camouflaged}, we allocated 1,000 images from the CAMO dataset and 3,040 images from the COD10K dataset for training, and reserved the remainder for testing.

\input{main-table/main-table}

\noindent\textbf{Evaluation Metrics.}
To evaluate the performance of our image COD model, we employ five widely recognized metrics: (1) The Structure-measure ($S_{m}$)~\cite{fan2017structure}, which assesses the spatial structure of the detected object; (2) The F-measure ($F_{\beta}$), provides a balanced measure of precision and recall; (3) The Weighted F-measure ($F^{\omega}_{\beta}$)~\cite{margolin2014evaluate}, an enhanced version that offers more reliable evaluation outcomes; (4) The Mean Absolute Error (MAE), which calculates the element-wise difference between the predicted map and the ground truth; (5) The E-measure ($E_{m}$)~\cite{fan2018enhanced}, which evaluates pixel-level matchings and image-level statistics simultaneously. 

\noindent\textbf{Implementation Details.}
The proposed model was built in PyTorch~\cite{fan2018enhanced} on an NVIDIA RTX A6000 GPU. The training configurations are consistent with the current best practices~\cite{pang2024zoomnext, ji2023deep, pang2022zoom, fan2020camouflaged}. The encoder's parameters were initialized with those of the PVTv2 encoder pre-trained on ImageNet, while the other model components were initialized randomly. The Adam optimizer was utilized to update the model parameters, with betas set to (0.9, 0.999). The learning rate was set to 0.0001, with stepwise decay. The model was trained for 150 epochs, with a batch size of 8. During training, the input and ground truth images were bilinearly interpolated to 384$\times$384. During testing, input images, prediction maps, and ground-truth images were interpolated to 384$\times$384. To conduct a fair comparison with other methods, we experimented with different input sizes to match their settings. Data augmentation techniques were applied to the training dataset by random flipping, rotation, and color jittering.

\begin{figure}[tb]
\centering
\begin{tabular}{cccccccc}
\includegraphics[width=0.118\textwidth]{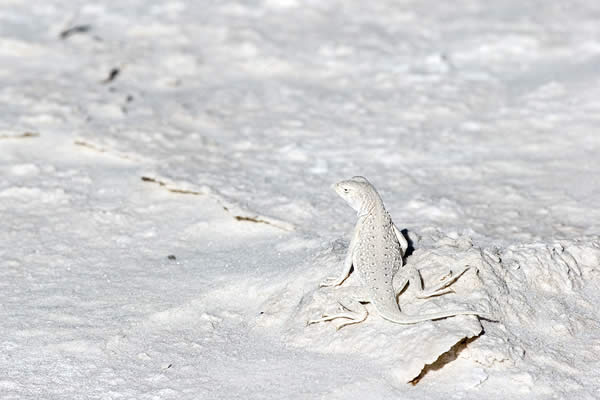} & 
\includegraphics[width=0.118\textwidth]{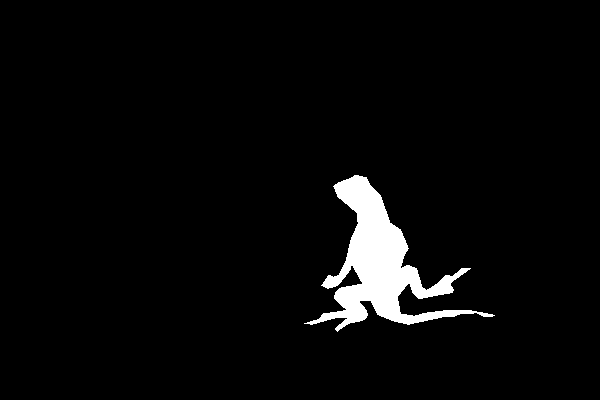} & 
\includegraphics[width=0.118\textwidth]{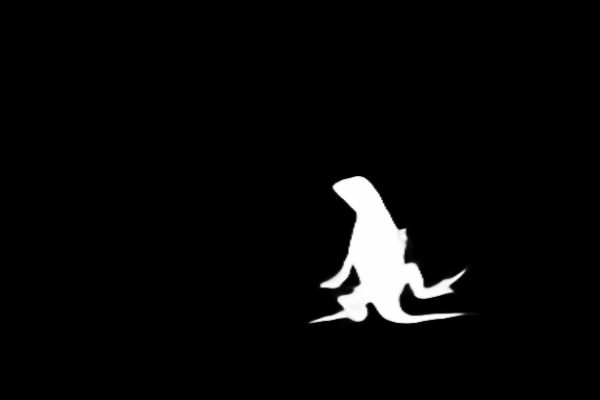} & 
\includegraphics[width=0.118\textwidth]{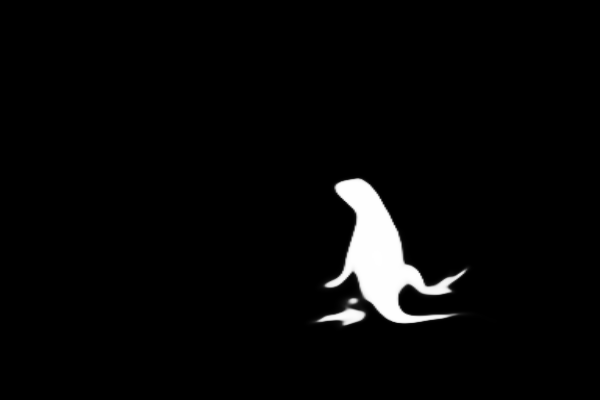} & 
\includegraphics[width=0.118\textwidth]{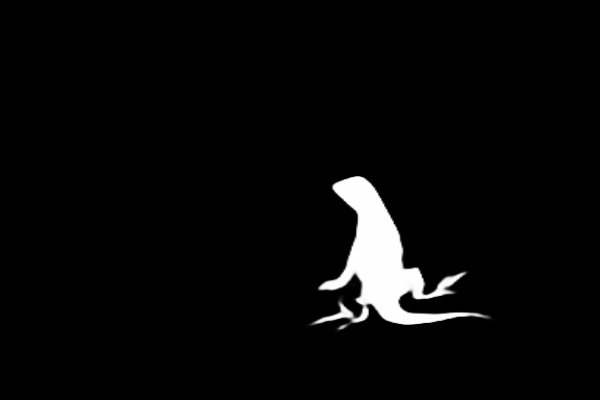} & 
\includegraphics[width=0.118\textwidth]{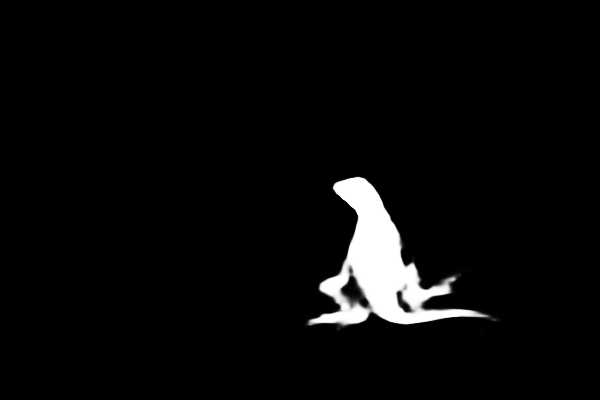} & 
\includegraphics[width=0.118\textwidth]{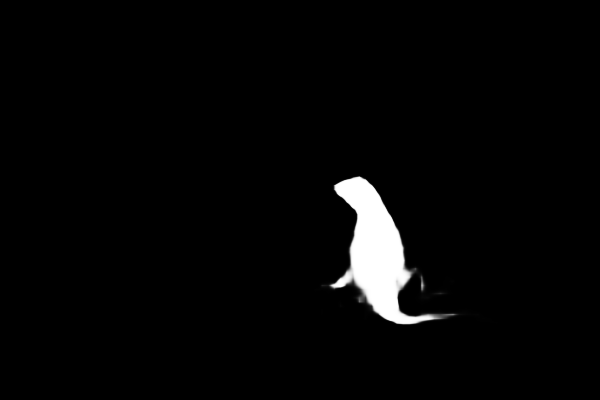} & 
\includegraphics[width=0.118\textwidth]{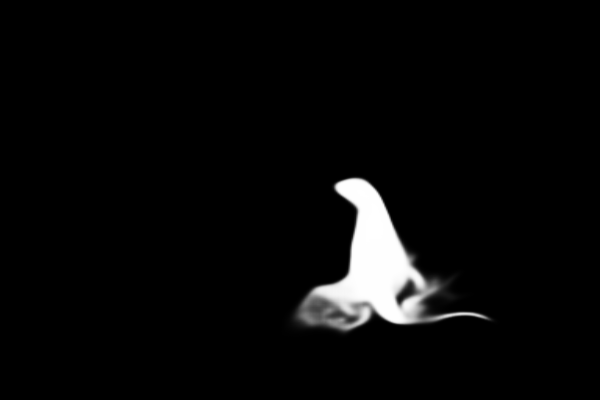} \\

\includegraphics[width=0.118\textwidth]{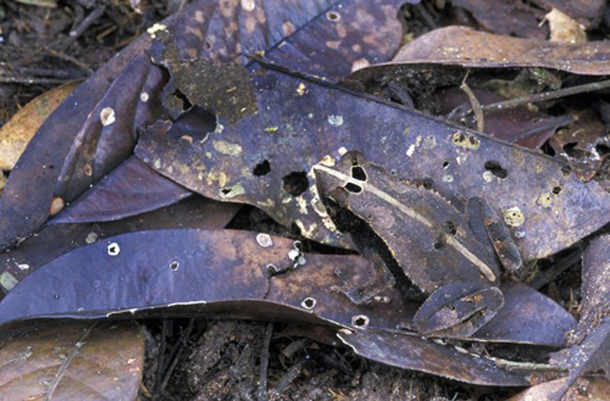} & 
\includegraphics[width=0.118\textwidth]{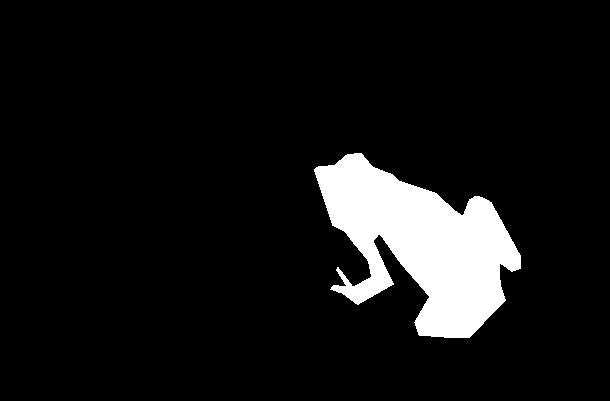} & 
\includegraphics[width=0.118\textwidth]{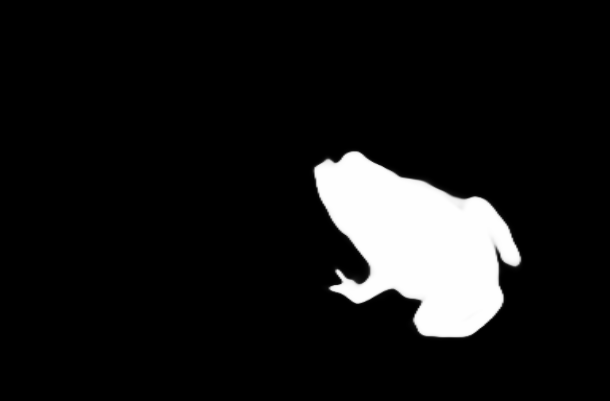} & 
\includegraphics[width=0.118\textwidth]{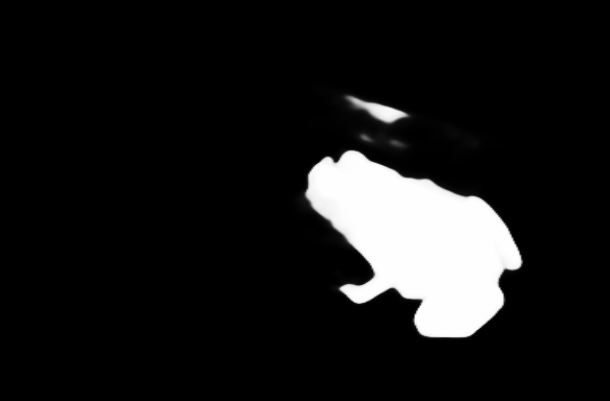} & 
\includegraphics[width=0.118\textwidth]{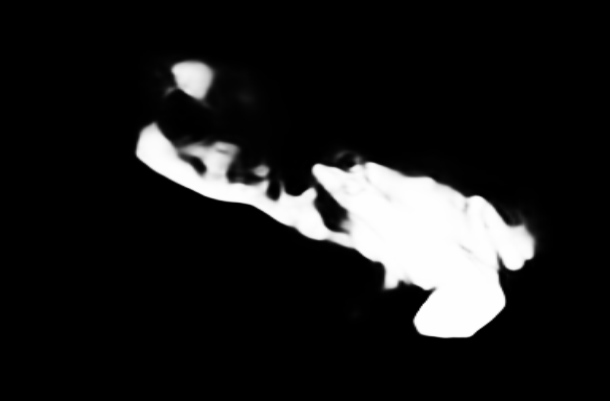} & 
\includegraphics[width=0.118\textwidth]{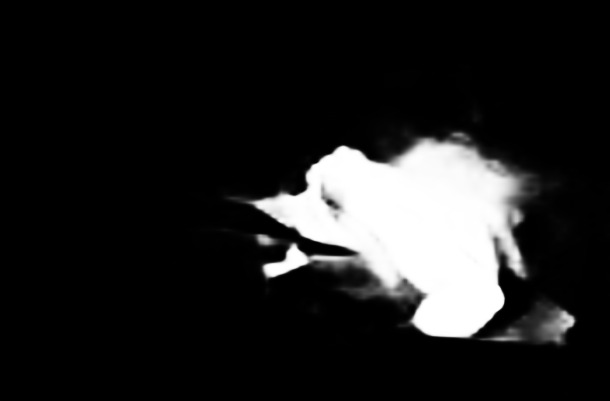} & 
\includegraphics[width=0.118\textwidth]{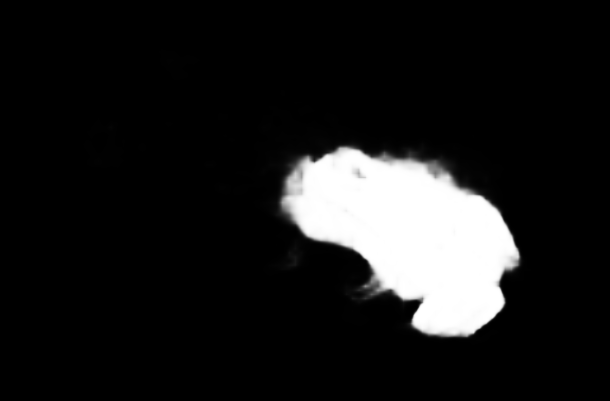} & 
\includegraphics[width=0.118\textwidth]{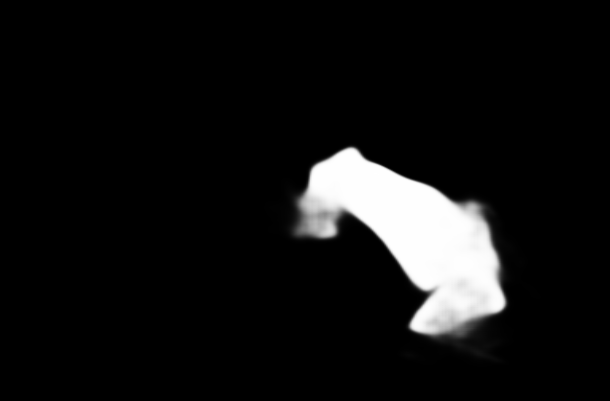} \\

\includegraphics[width=0.118\textwidth]{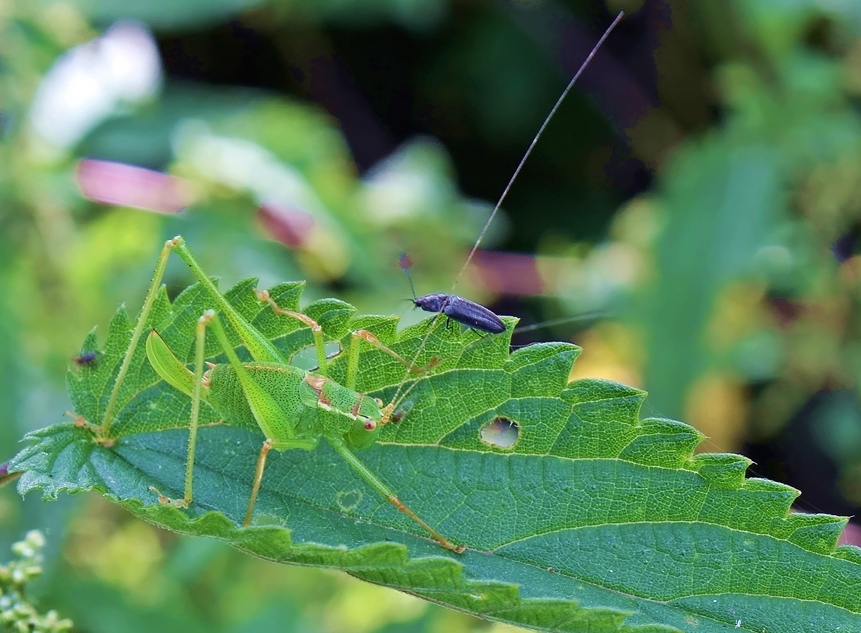} & 
\includegraphics[width=0.118\textwidth]{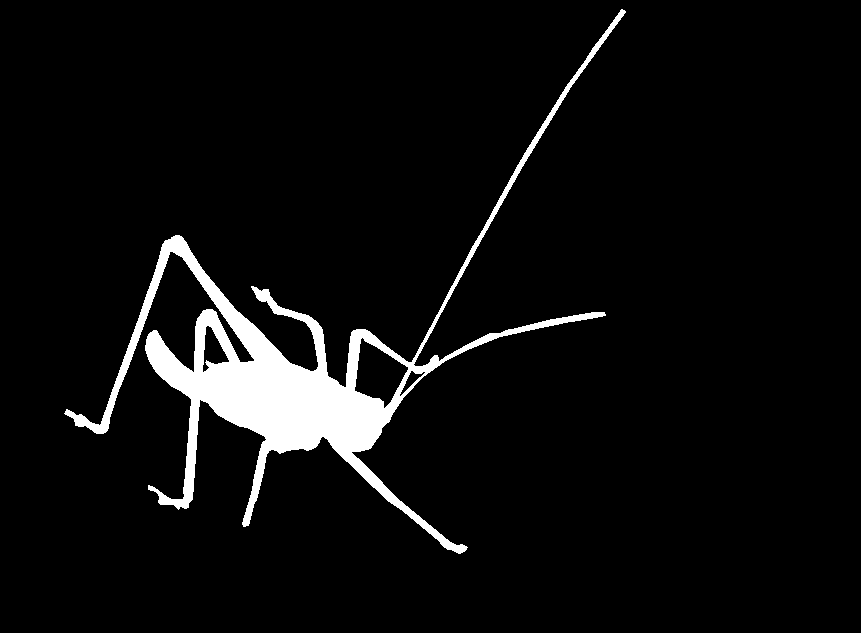} & 
\includegraphics[width=0.118\textwidth]{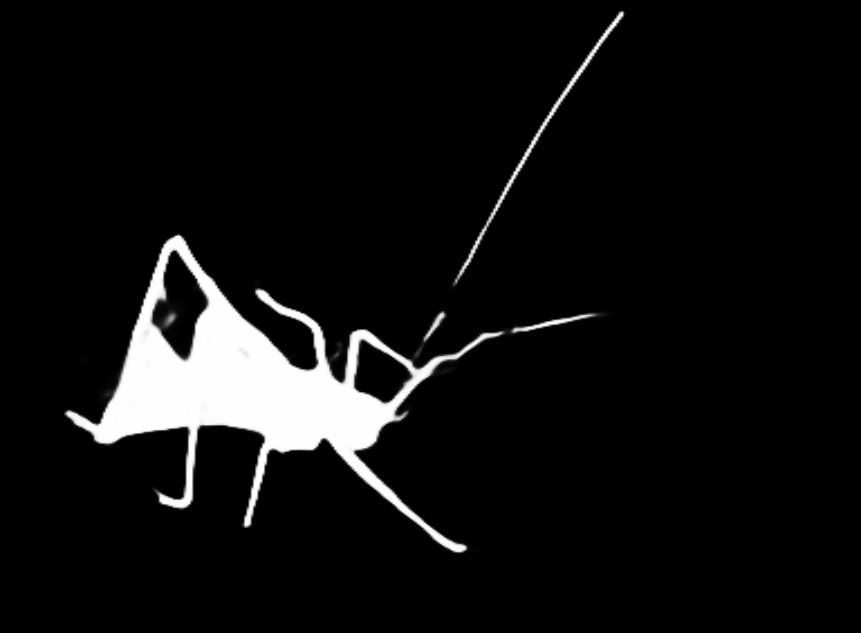} & 
\includegraphics[width=0.118\textwidth]{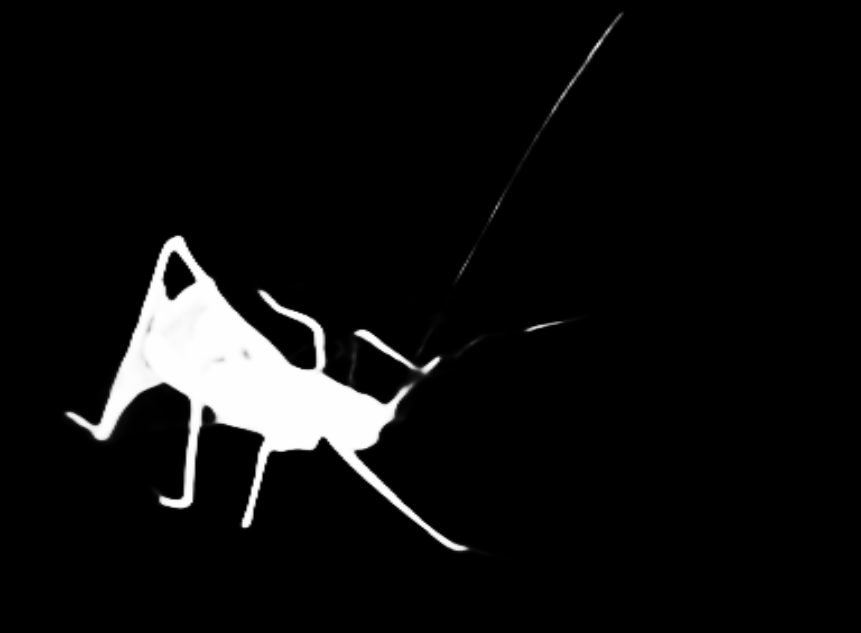} & 
\includegraphics[width=0.118\textwidth]{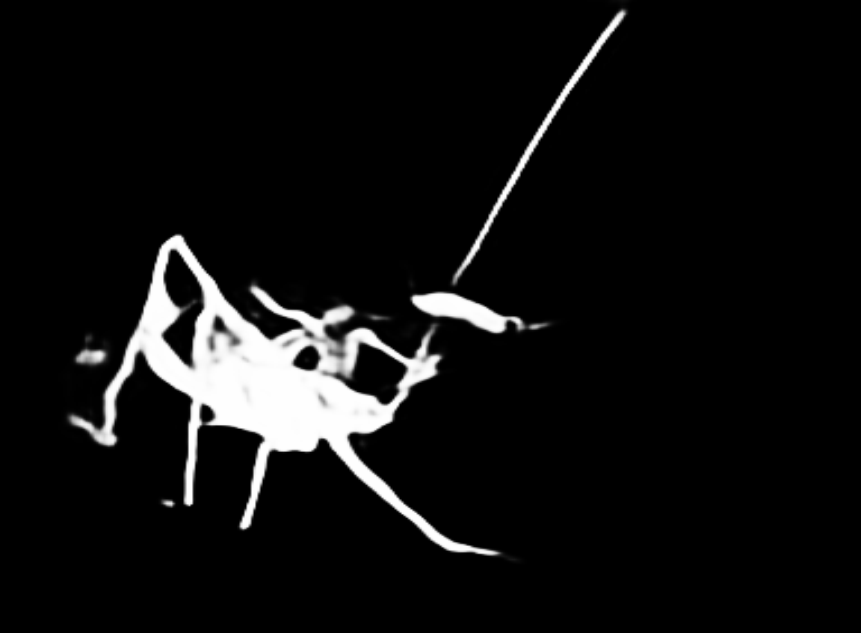} & 
\includegraphics[width=0.118\textwidth]{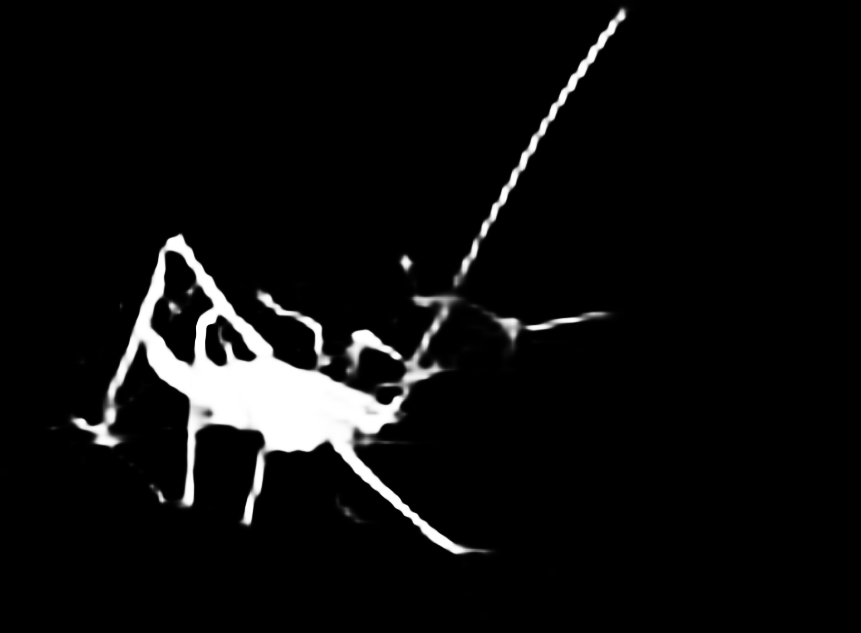} & 
\includegraphics[width=0.118\textwidth]{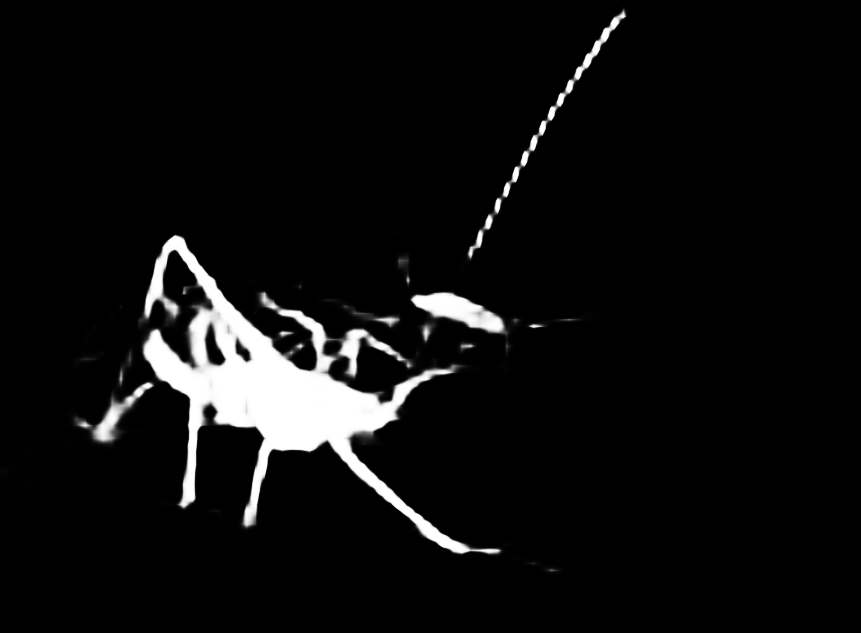} & 
\includegraphics[width=0.118\textwidth]{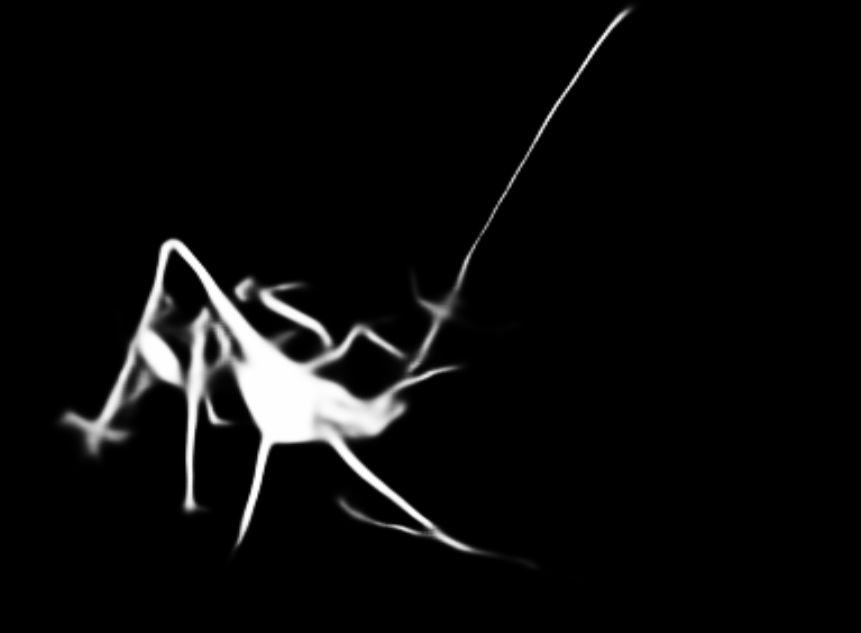} \\

\includegraphics[width=0.118\textwidth]{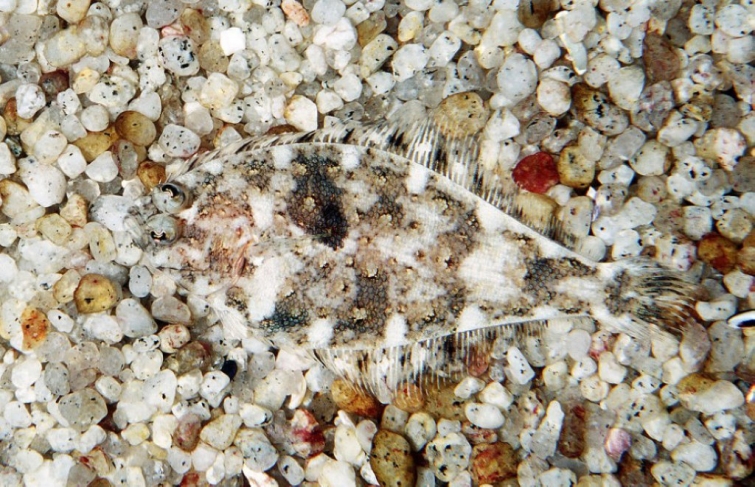} & 
\includegraphics[width=0.118\textwidth]{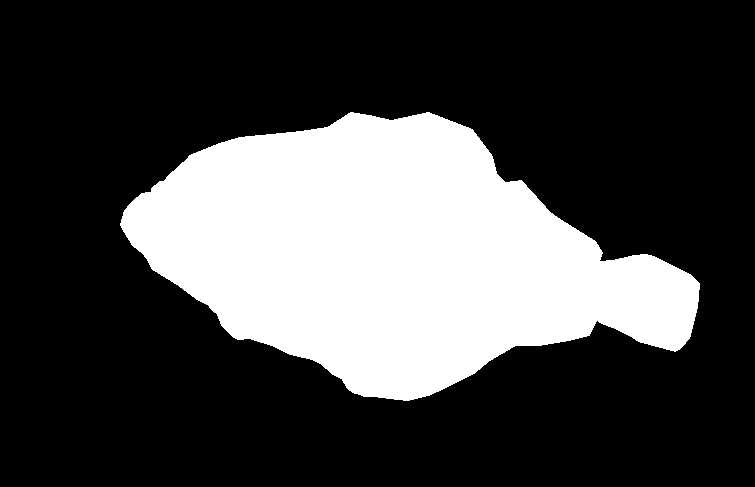} & 
\includegraphics[width=0.118\textwidth]{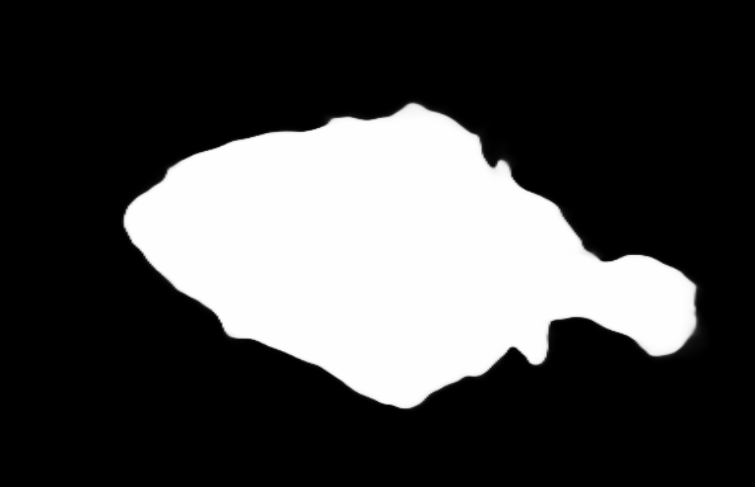} & 
\includegraphics[width=0.118\textwidth]{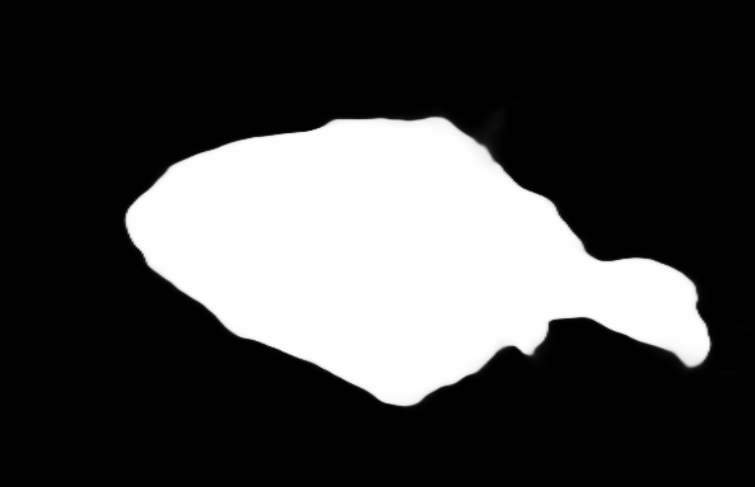} & 
\includegraphics[width=0.118\textwidth]{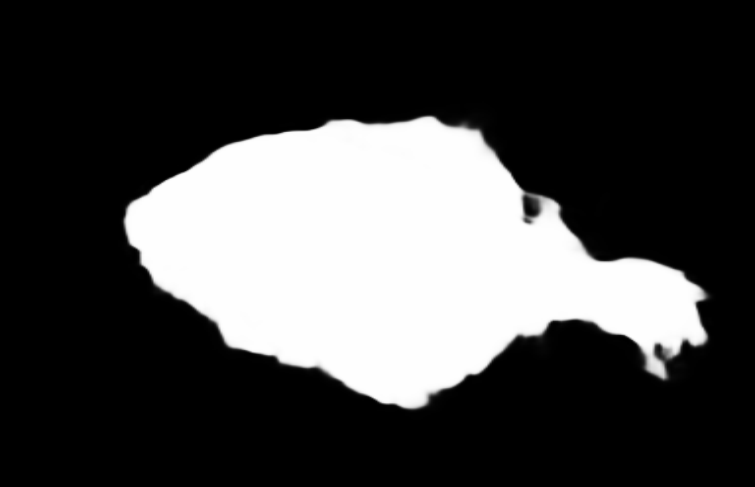} & 
\includegraphics[width=0.118\textwidth]{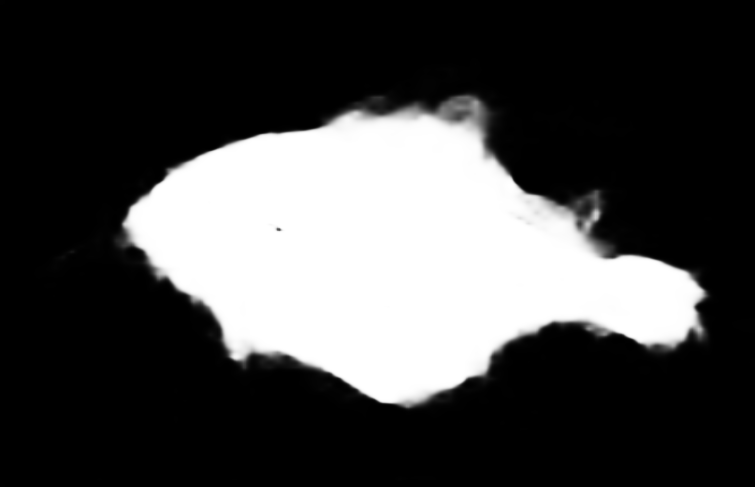} & 
\includegraphics[width=0.118\textwidth]{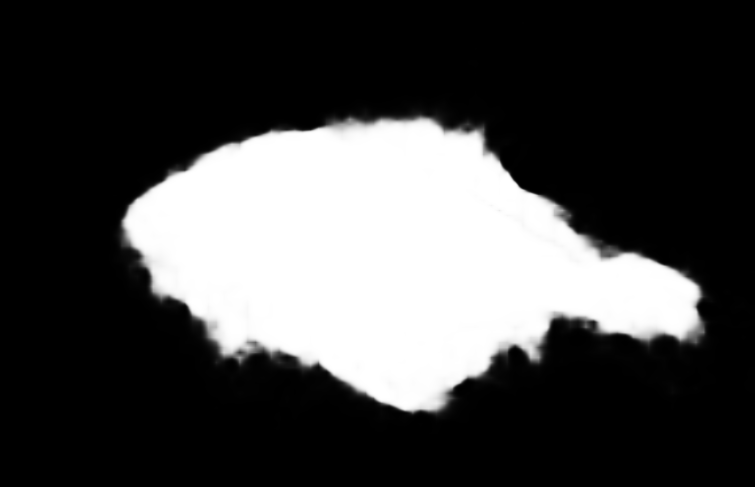} & 
\includegraphics[width=0.118\textwidth]{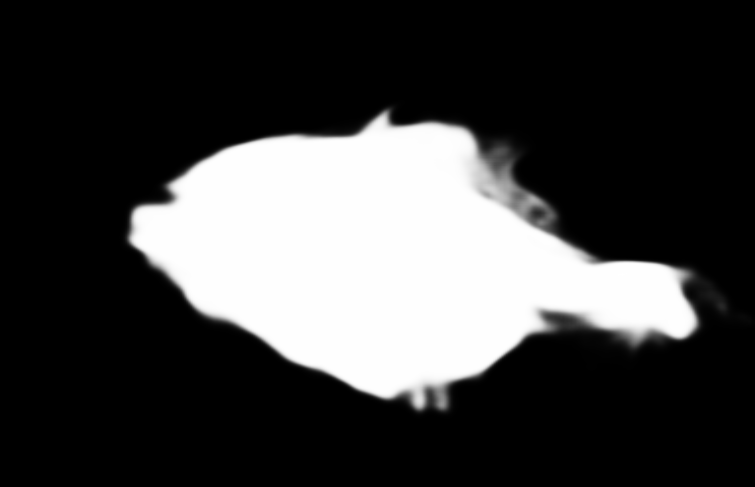} \\

\includegraphics[width=0.118\textwidth]{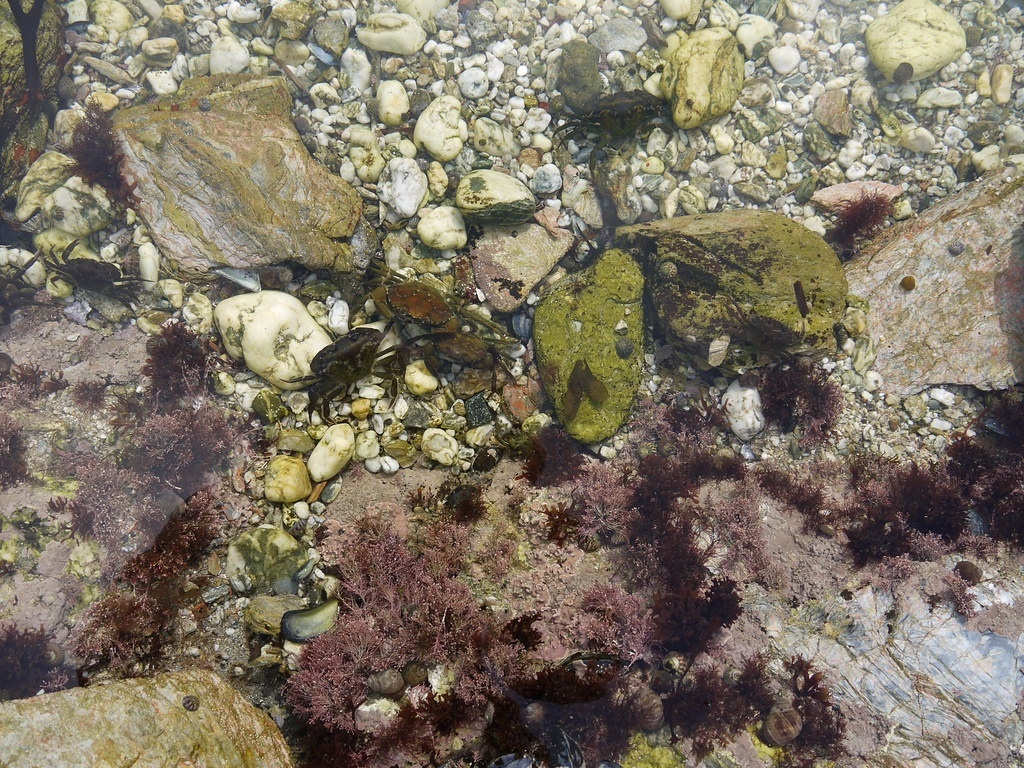} & 
\includegraphics[width=0.118\textwidth]{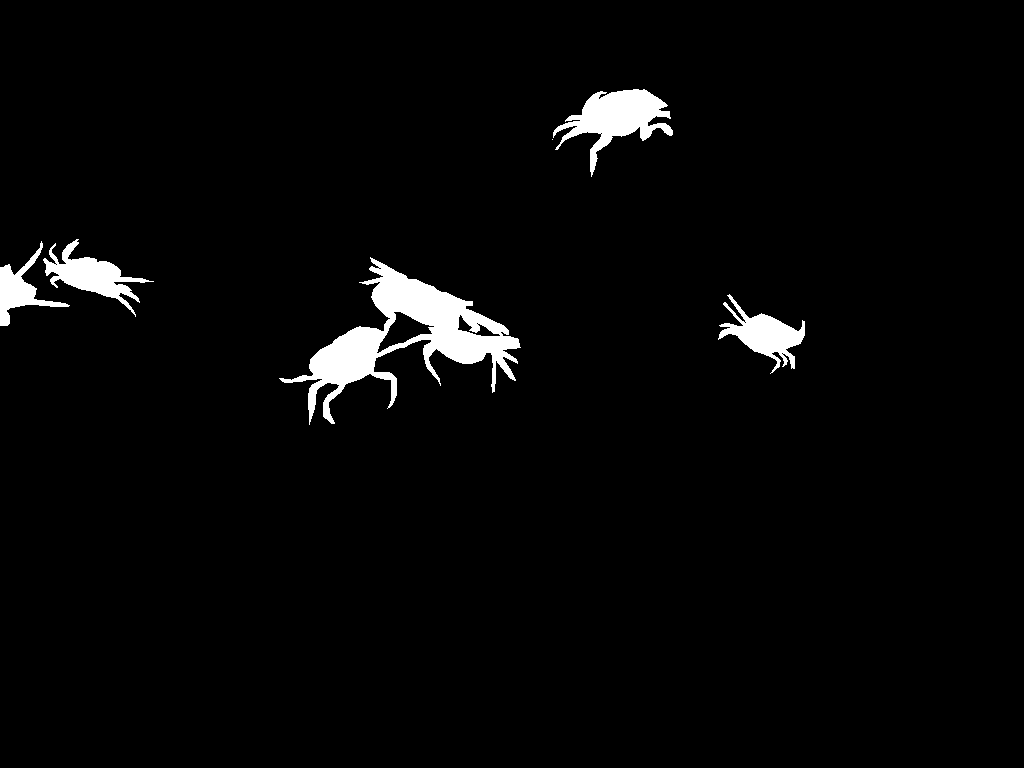} & 
\includegraphics[width=0.118\textwidth]{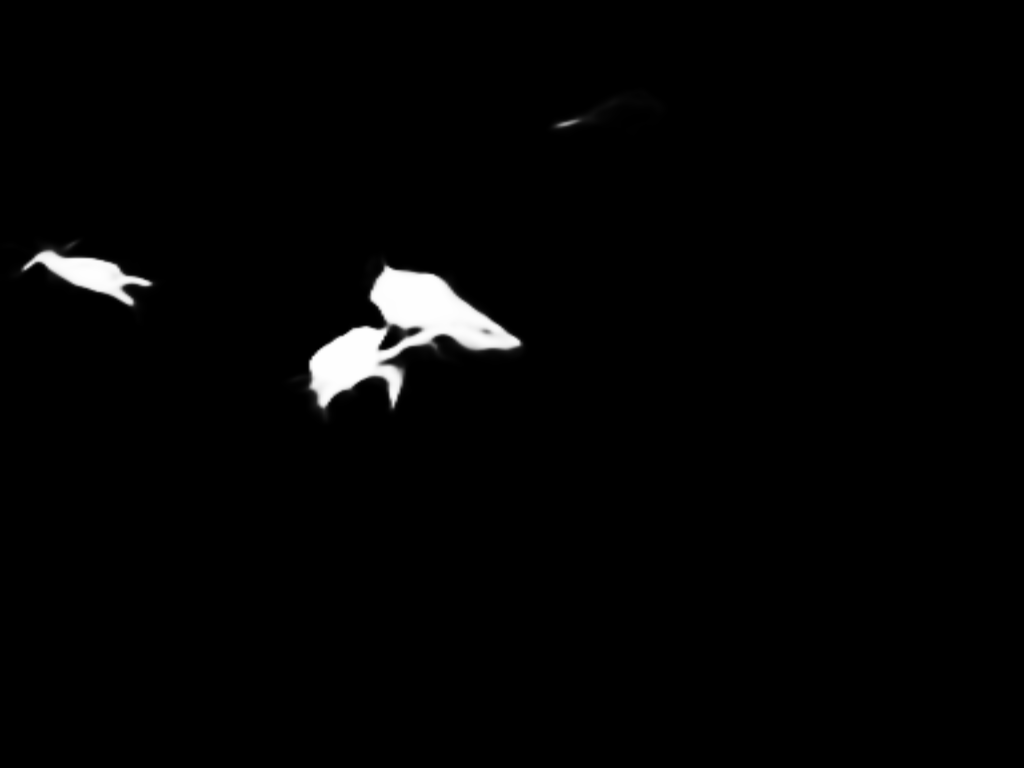} & 
\includegraphics[width=0.118\textwidth]{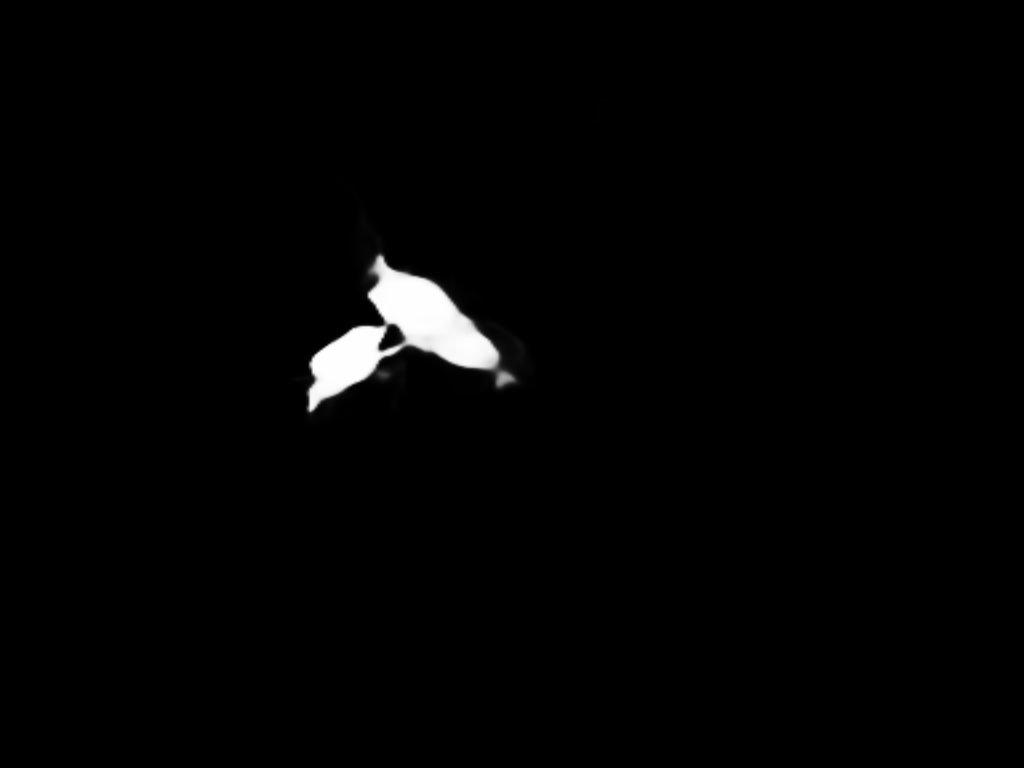} & 
\includegraphics[width=0.118\textwidth]{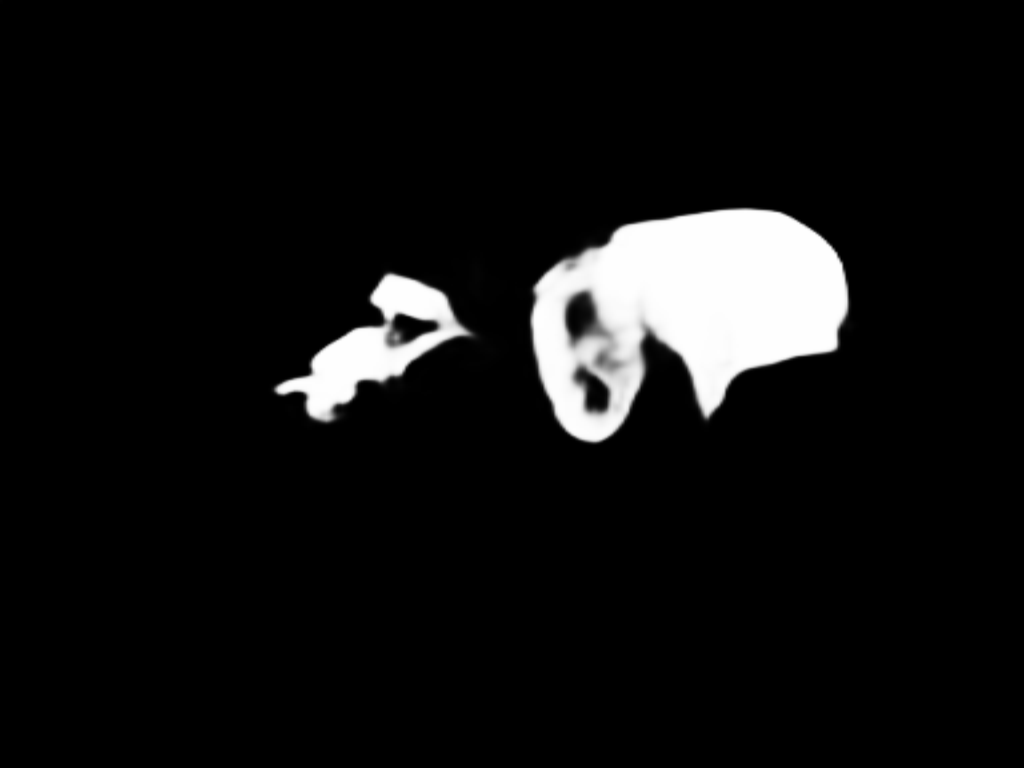} & 
\includegraphics[width=0.118\textwidth]{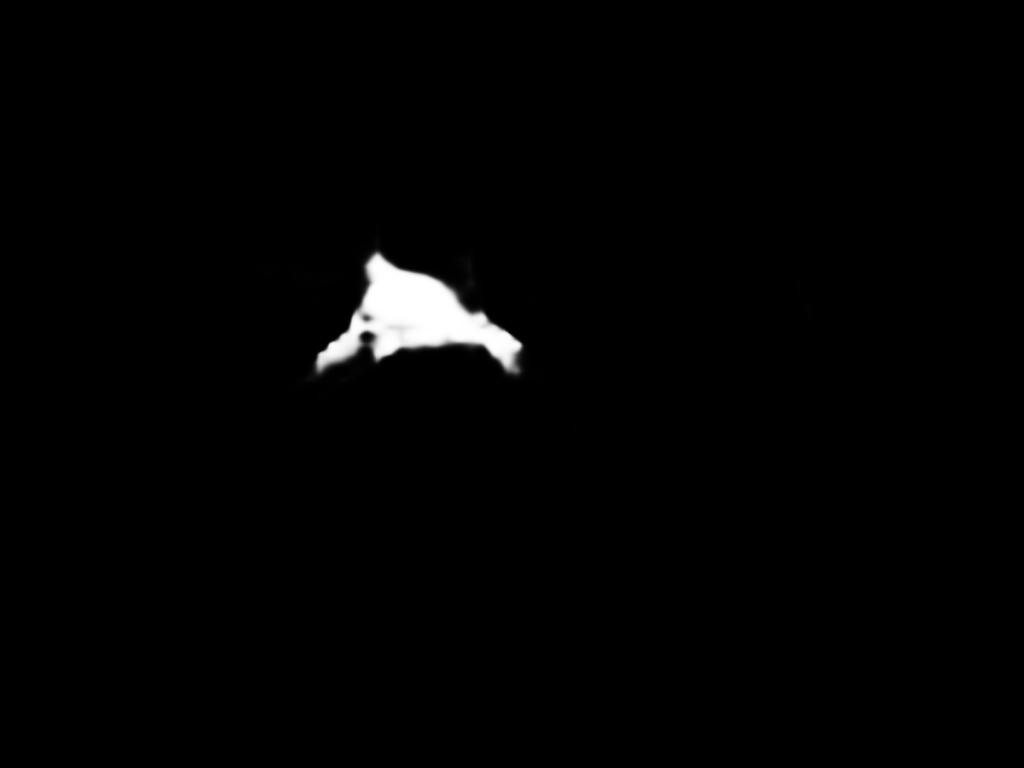} & 
\includegraphics[width=0.118\textwidth]{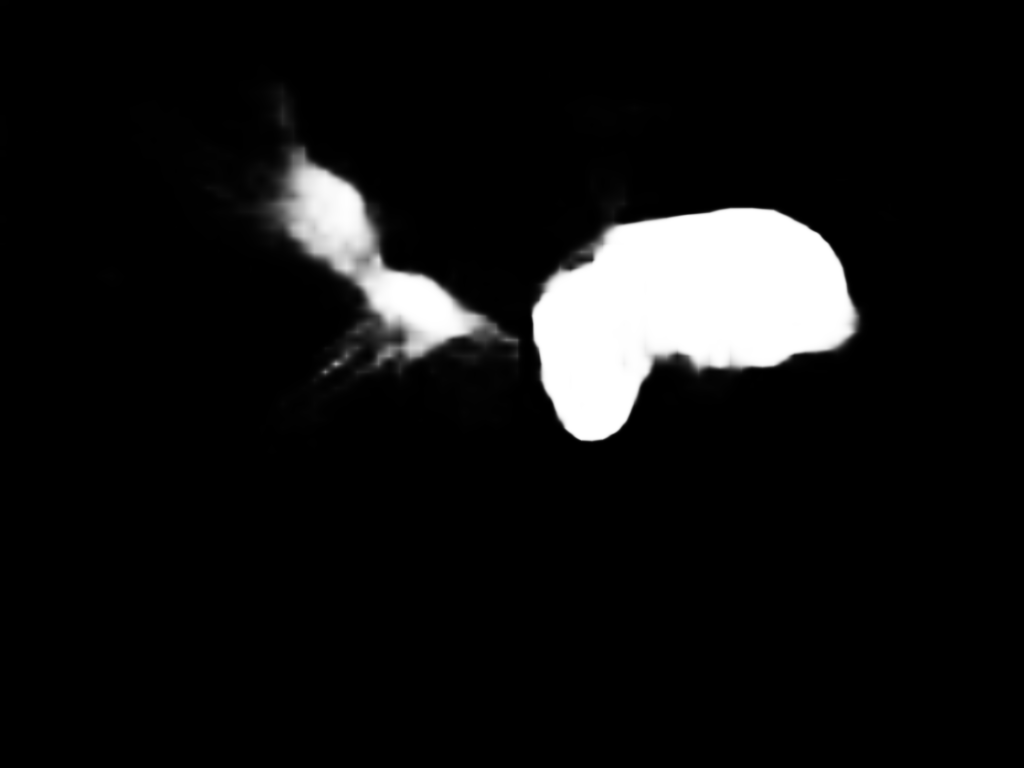} & 
\includegraphics[width=0.118\textwidth]{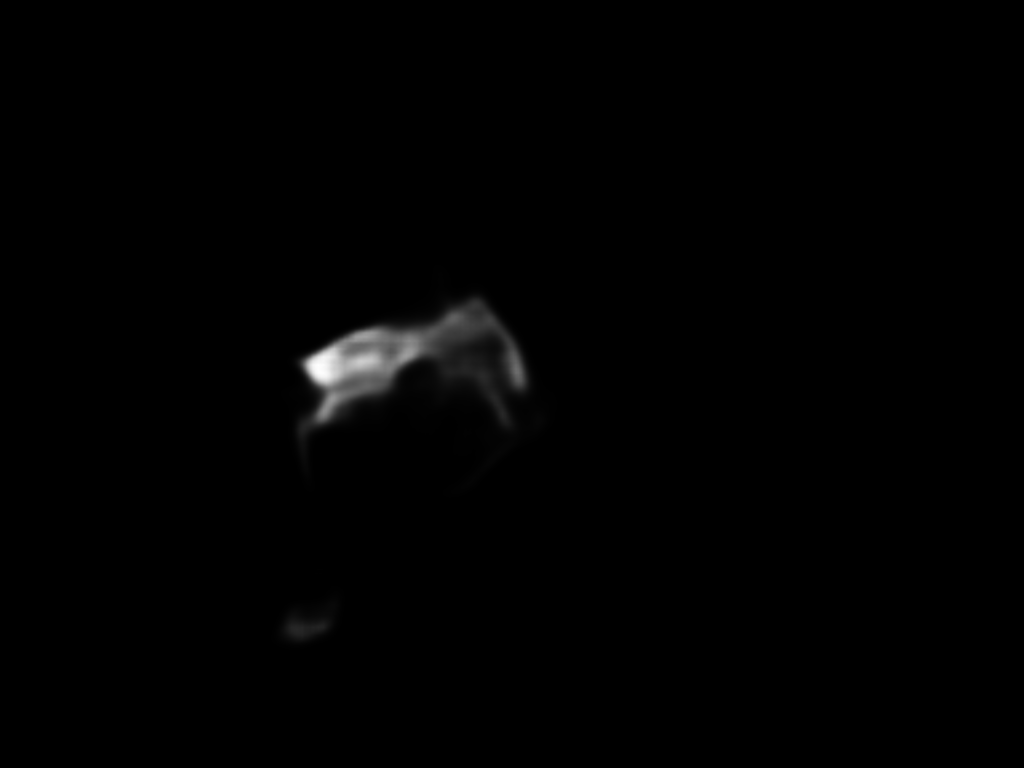} \\

\includegraphics[width=0.118\textwidth]{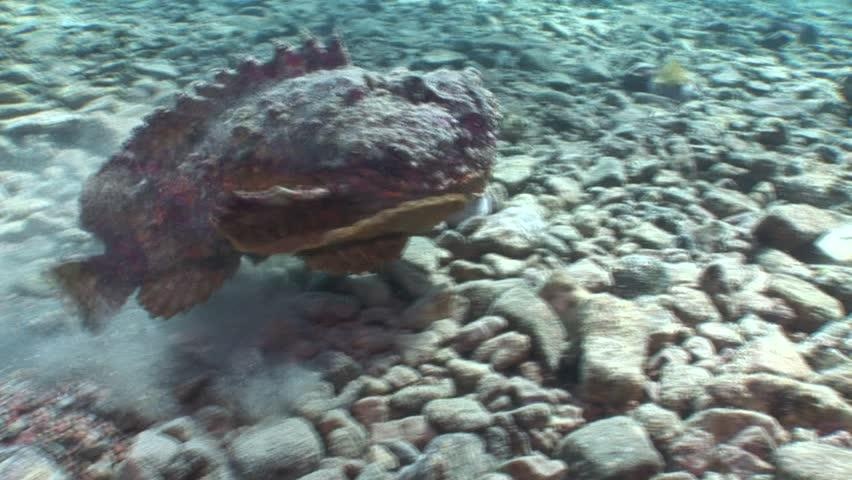} & 
\includegraphics[width=0.118\textwidth]{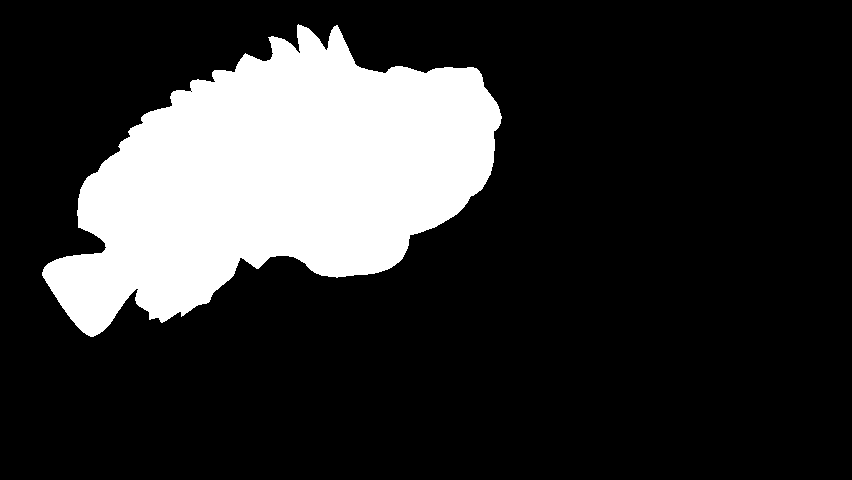} & 
\includegraphics[width=0.118\textwidth]{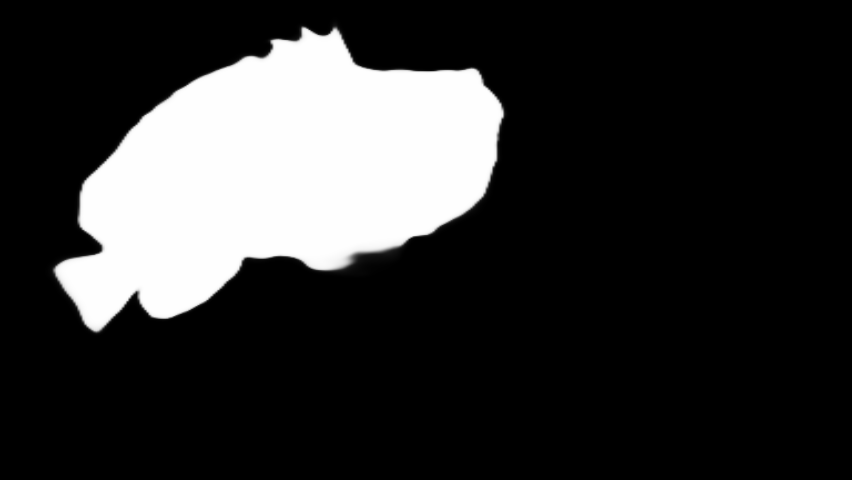} & 
\includegraphics[width=0.118\textwidth]{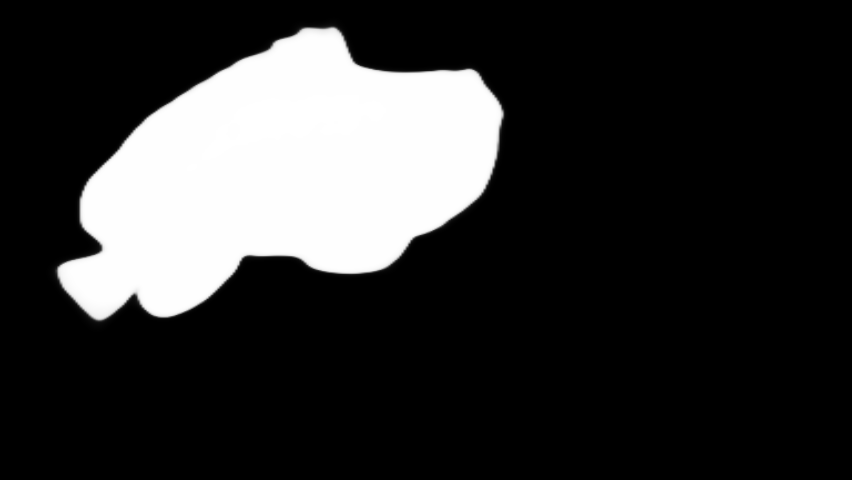} & 
\includegraphics[width=0.118\textwidth]{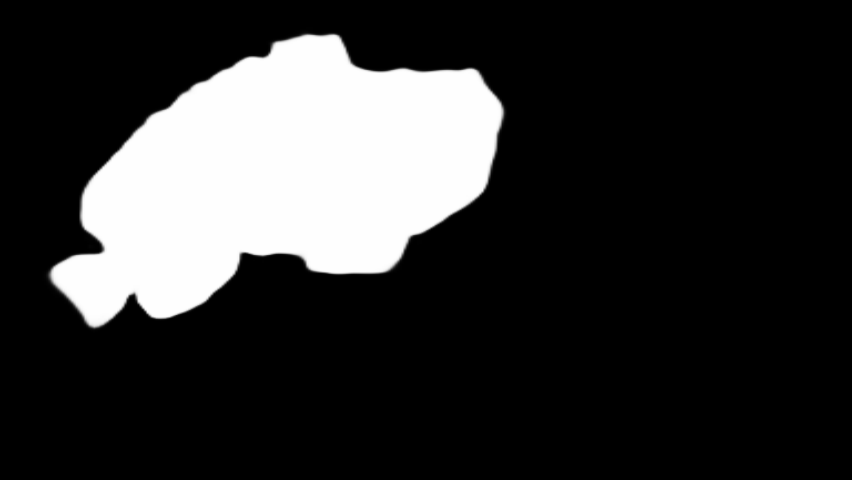} & 
\includegraphics[width=0.118\textwidth]{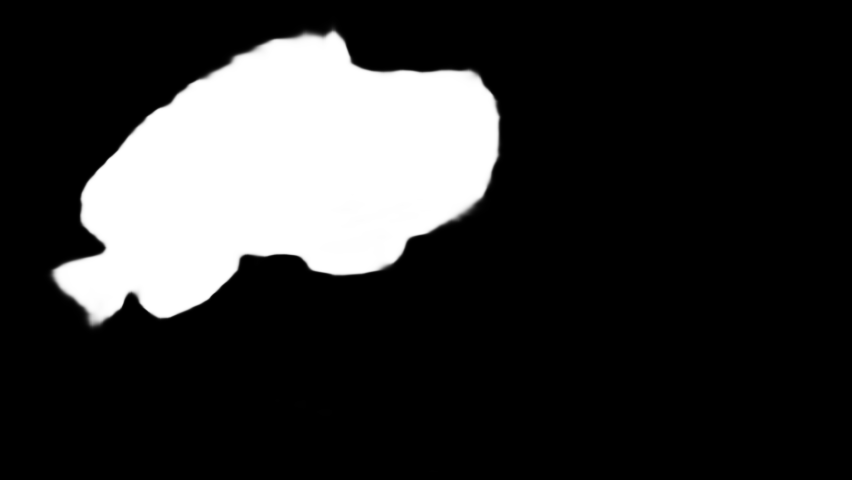} & 
\includegraphics[width=0.118\textwidth]{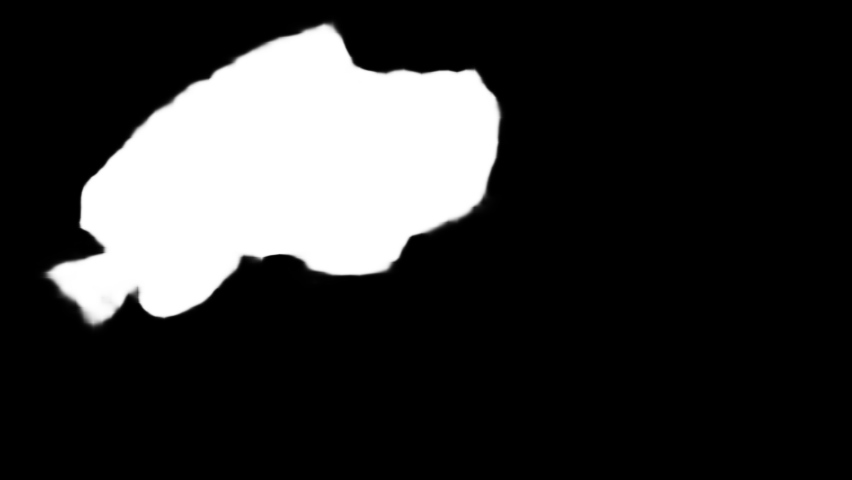} & 
\includegraphics[width=0.118\textwidth]{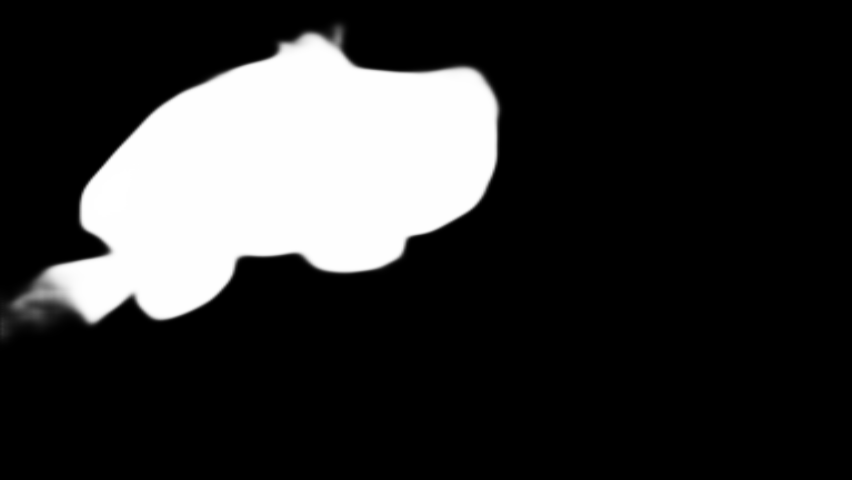} \\

\includegraphics[width=0.118\textwidth]{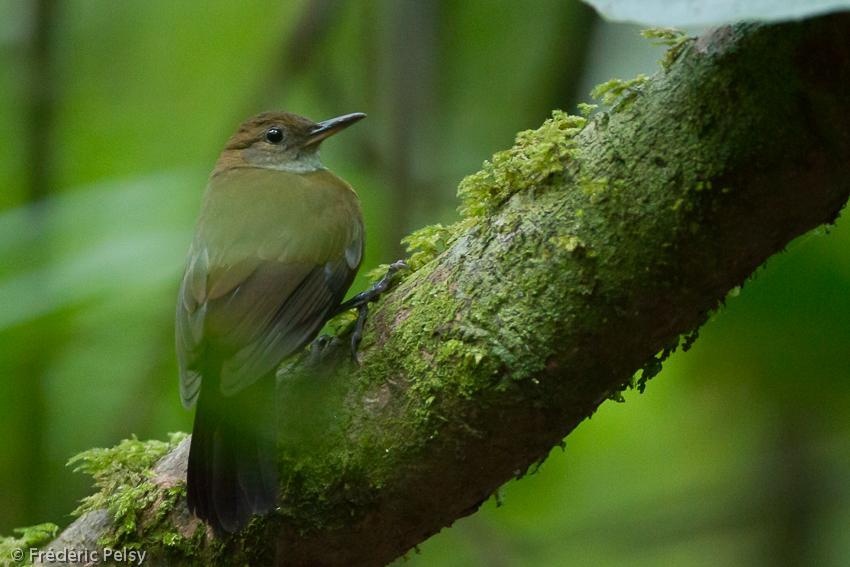} & 
\includegraphics[width=0.118\textwidth]{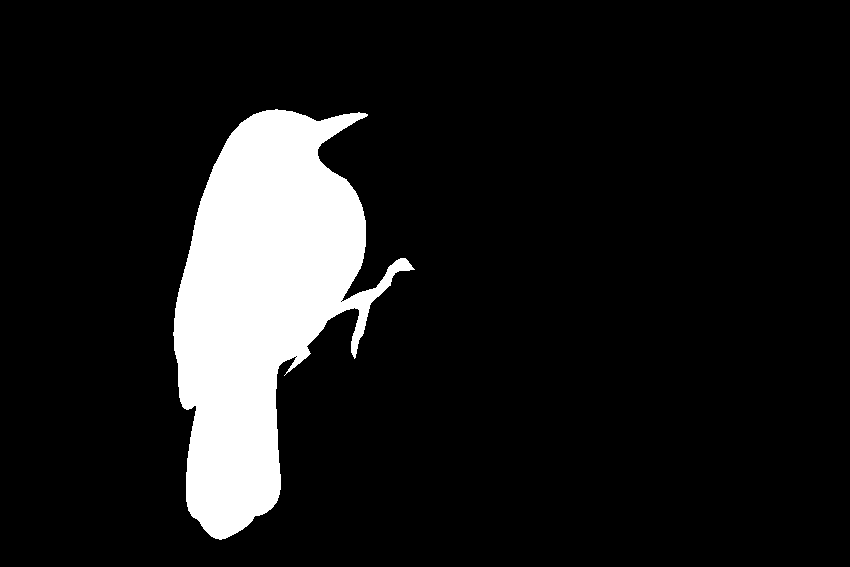} & 
\includegraphics[width=0.118\textwidth]{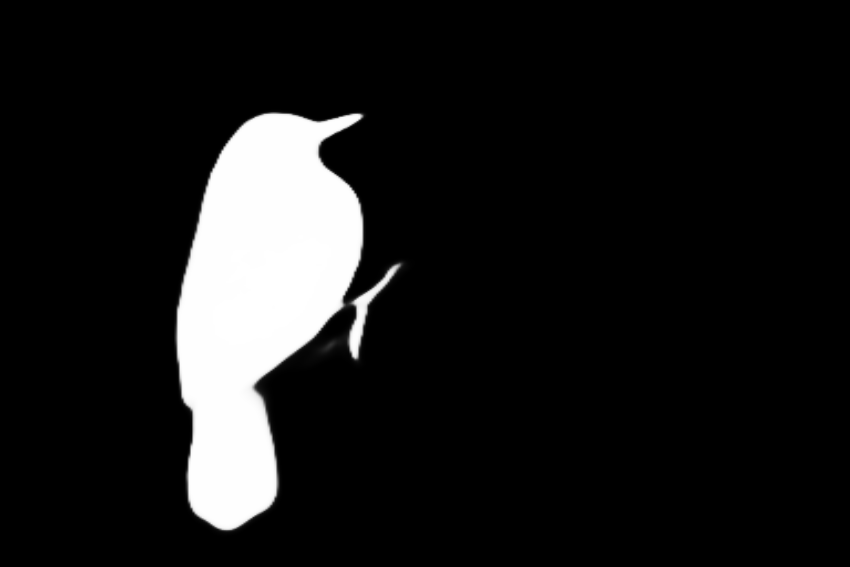} & 
\includegraphics[width=0.118\textwidth]{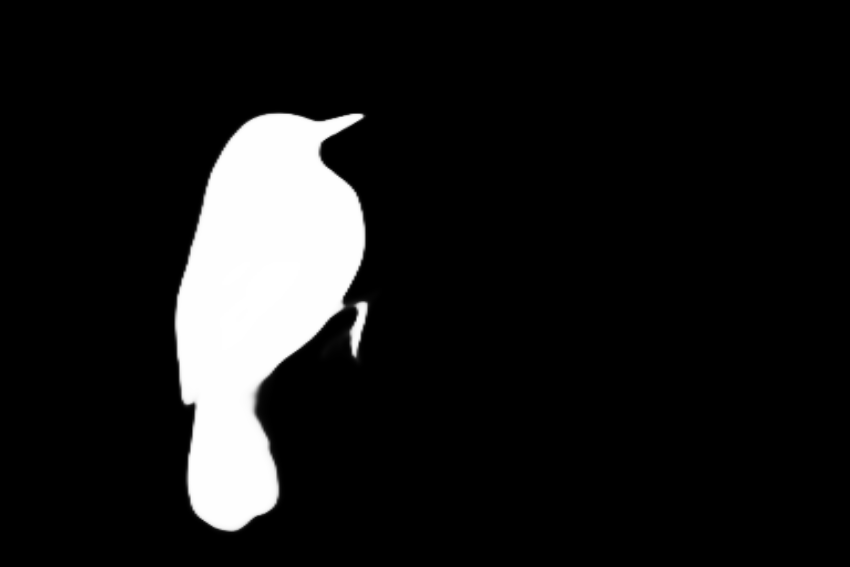} & 
\includegraphics[width=0.118\textwidth]{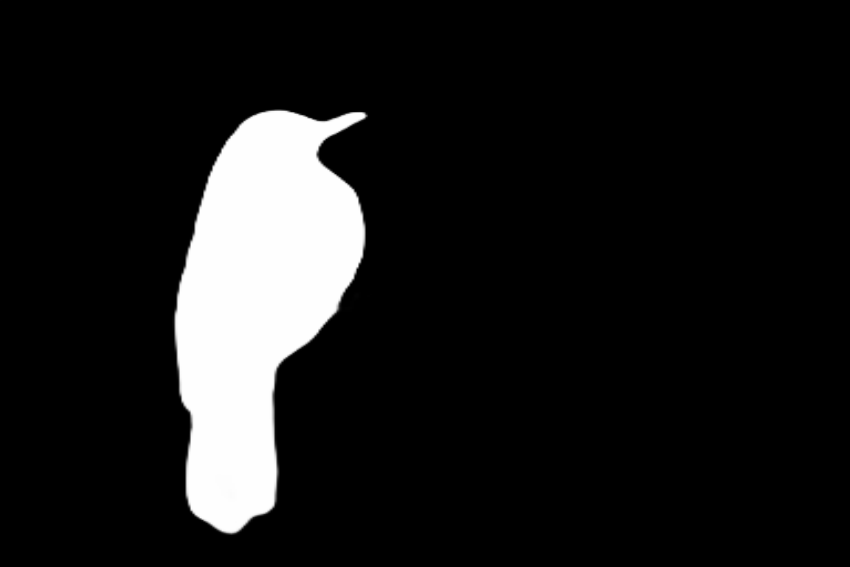} & 
\includegraphics[width=0.118\textwidth]{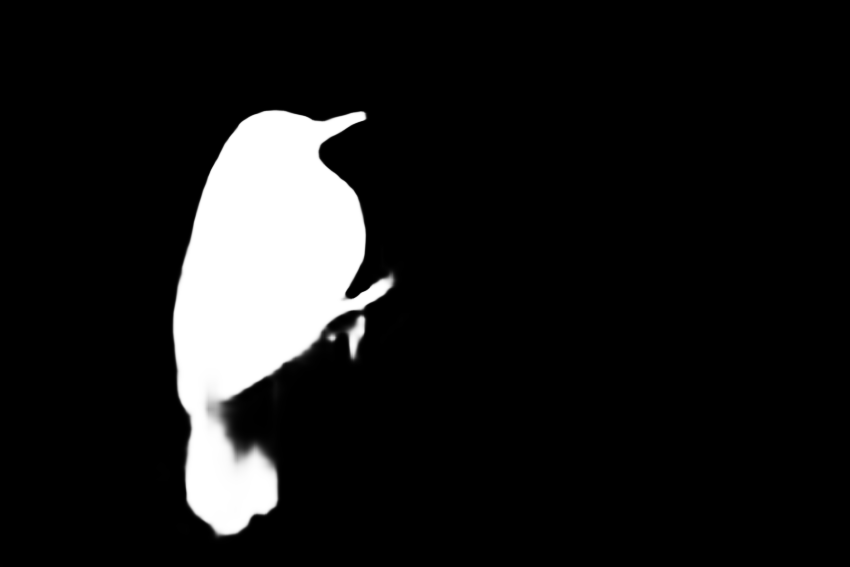} & 
\includegraphics[width=0.118\textwidth]{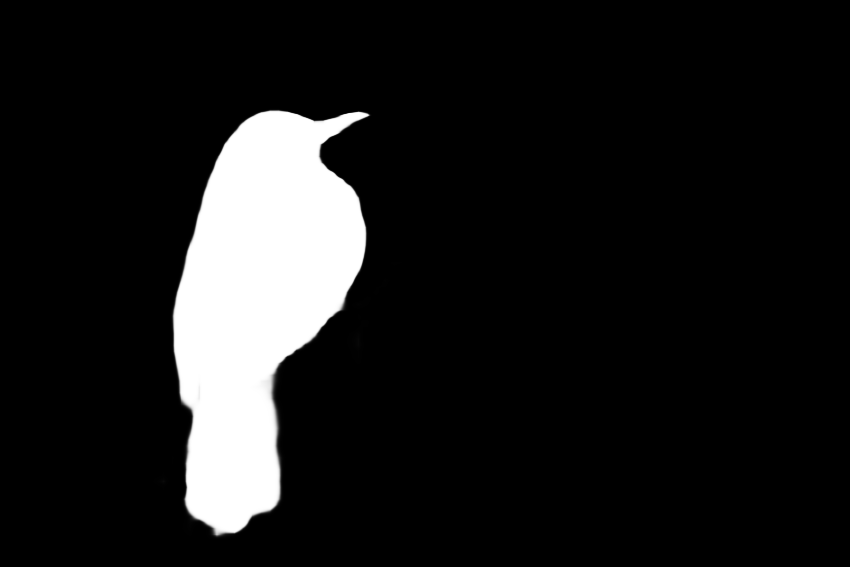} & 
\includegraphics[width=0.118\textwidth]{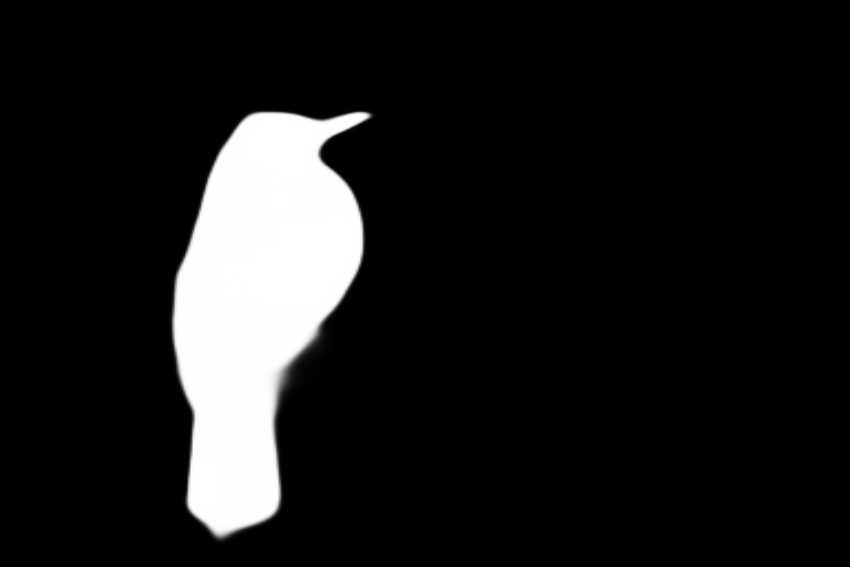} \\
Input & GT & Ours & \cite{pang2024zoomnext} & \cite{xing2023go} & \cite{liu2023mscaf} & \cite{yin2024camoformer} & \cite{huang2023feature}\\ 

\end{tabular}
\caption{Visual Comparison between our model and competing SOTA methods.} 
\label{fig:Visual_comparison}
\end{figure}
\subsection{Results and Comparison}
In this section, we present comprehensive quantitative and qualitative comparisons between our model and the current state-of-the-art methods (SOTA). All results for SOTA methods are taken from their respective publications.

\noindent\textbf{Quantitative Comparisons.} We compare our model with 20 SOTA models. \Cref{tab:results} presents the results obtained from all methods applied to the mentioned COD datasets. It includes the employed backbones, input image dimensions, and the total number of parameters utilized. In the context of CNN-based methodologies, EfficientNet-based models outperform ResNet-based models, particularly for feature extraction tasks. Additionally, our CNN-based model outperforms all SOTA models across all examined datasets.

Although CNN-based methods achieve good performance on this complex task, vision-based methods consistently outperform them. In particular, our vision-based model achieves state-of-the-art results on the CAMO10K and NC4K datasets without needing any additional training data. Furthermore, our model achieves second place on both the CAMO and CHAMELEON datasets, trailing only ZoomNeXt~\cite{pang2024zoomnext} and SARNet~\cite{xing2023go}, respectively. Moreover, experimental results show that our model typically has fewer parameters than models using the same backbone and configurations, indicating a reduced computational burden.


\noindent\textbf{Qualitative Comparisons.} 
\Cref{fig:Visual_comparison} provides a comparative analysis of our model against the top five SOTA methods, using sample images drawn from different datasets. These samples thoroughly illustrate the challenging aspects of the COD task, including variations in object sizes, differing proportions of fine details, and objects with indistinguishable boundaries seamlessly integrated into their backgrounds. The visual results indicate that our model demonstrates superior performance compared to other methods in multiple aspects, such as the ability to capture finer details (all rows), better-defined object areas (rows 1, 2, 5 and 6), more precise corners (rows 3 and 6), detection of small-sized objects and the identification of multiple objects present within a single scene (row 5).

\begin{figure}[tb]
\centering
\begin{tabular}{cccccccc}

\includegraphics[width=0.118\textwidth]{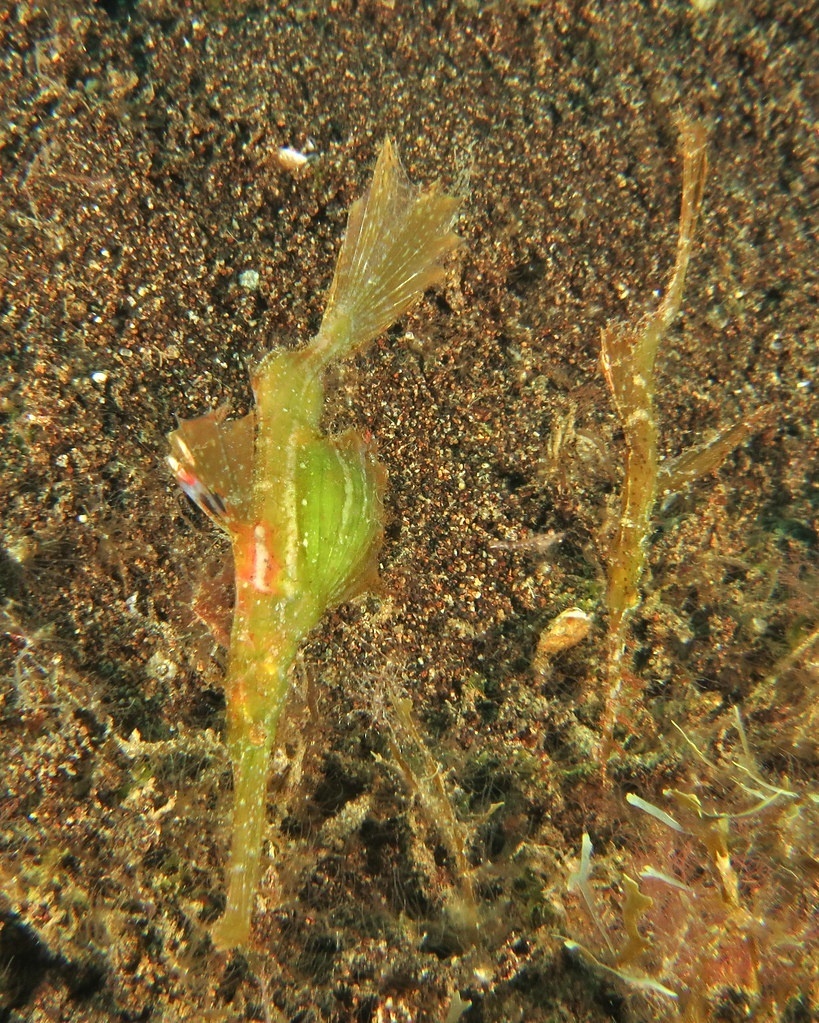} & 
\includegraphics[width=0.118\textwidth]{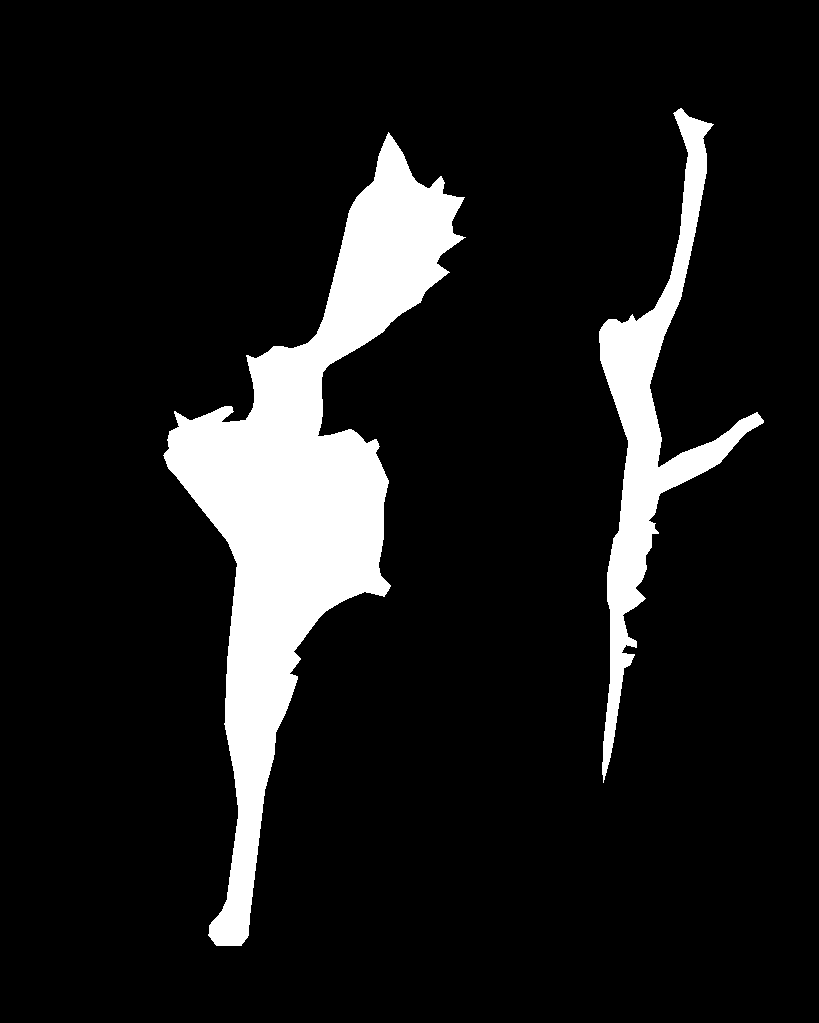} & 
\includegraphics[width=0.118\textwidth]{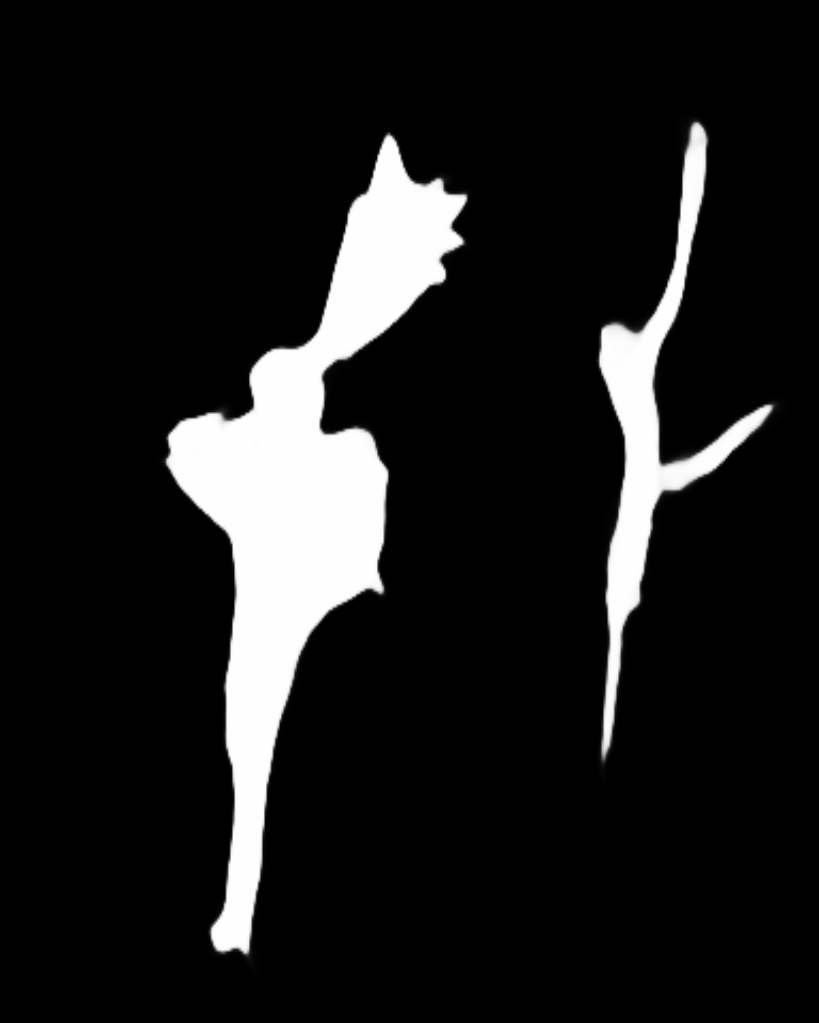} & 
\includegraphics[width=0.118\textwidth]{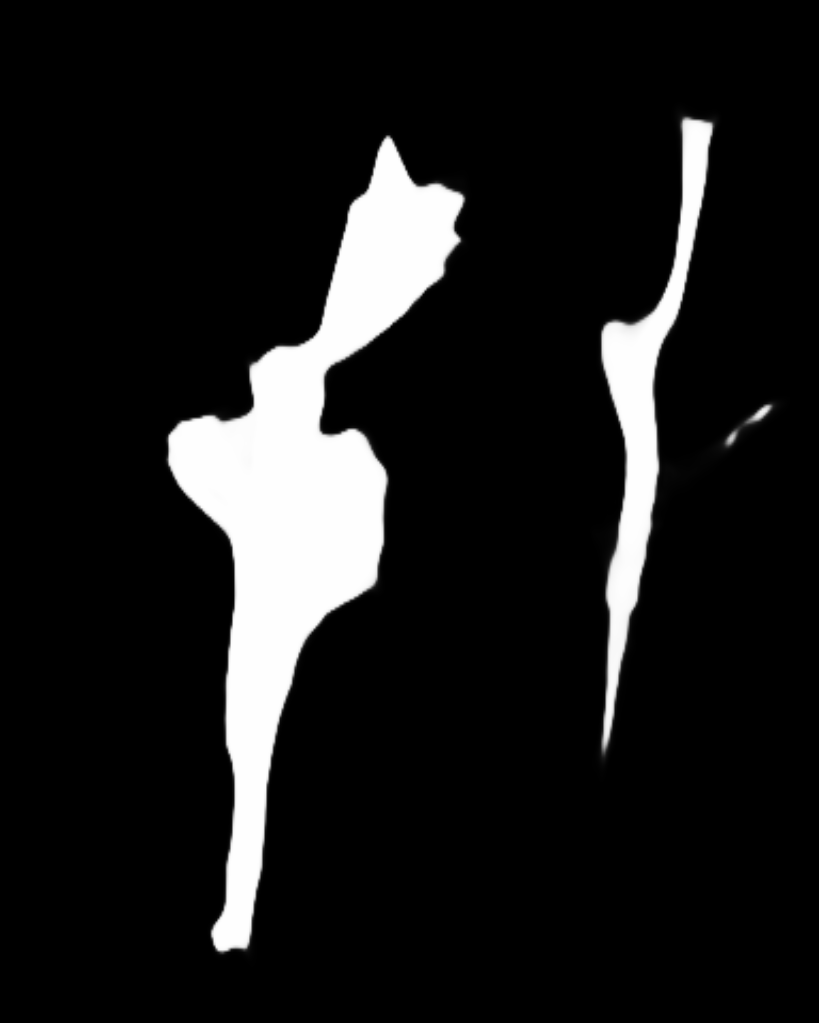} & 
\includegraphics[width=0.118\textwidth]{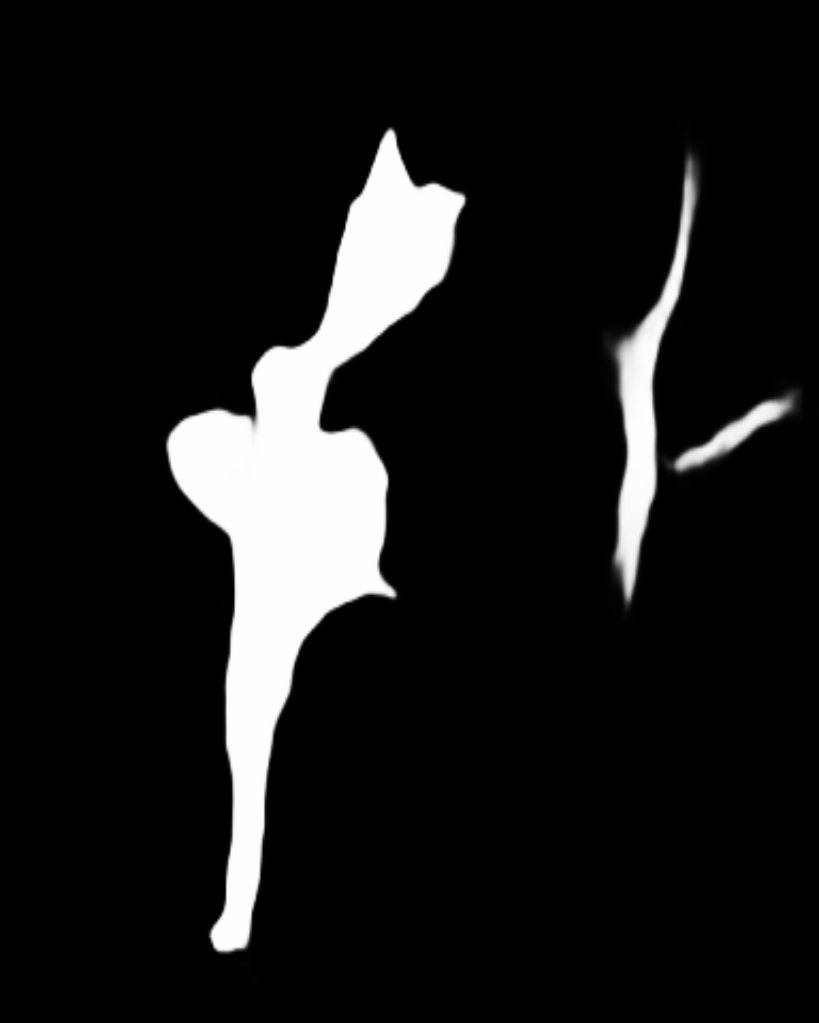} & 
\includegraphics[width=0.118\textwidth]{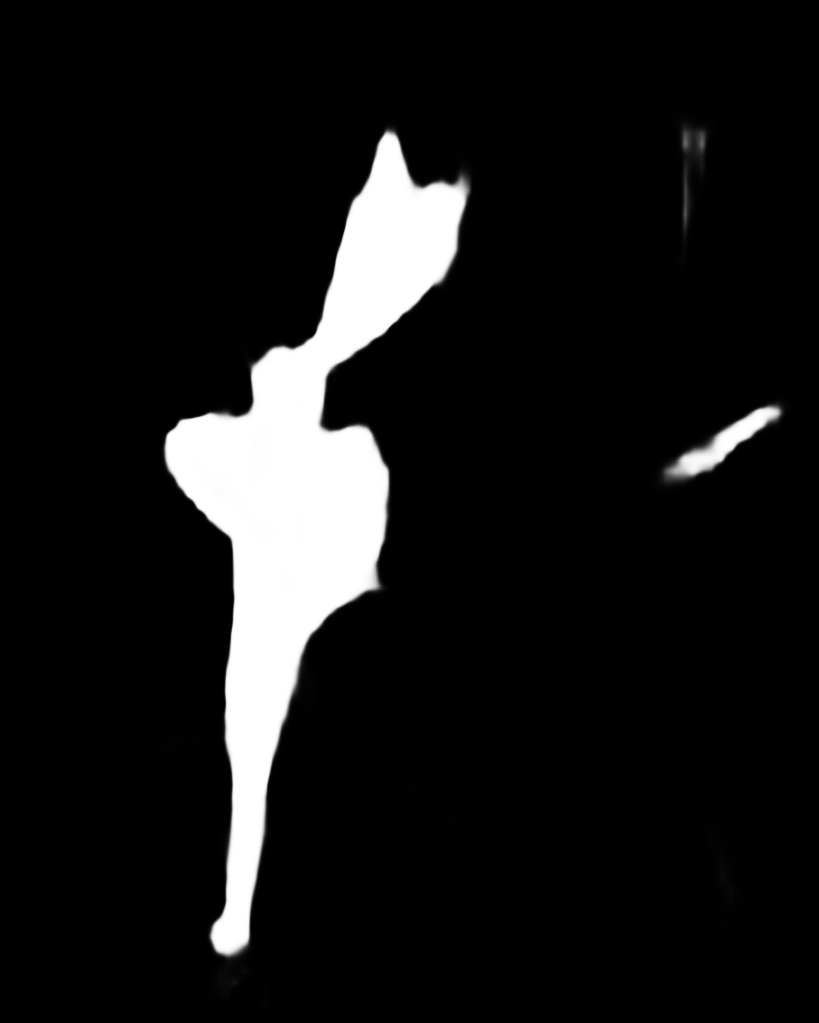} & 
\includegraphics[width=0.118\textwidth]{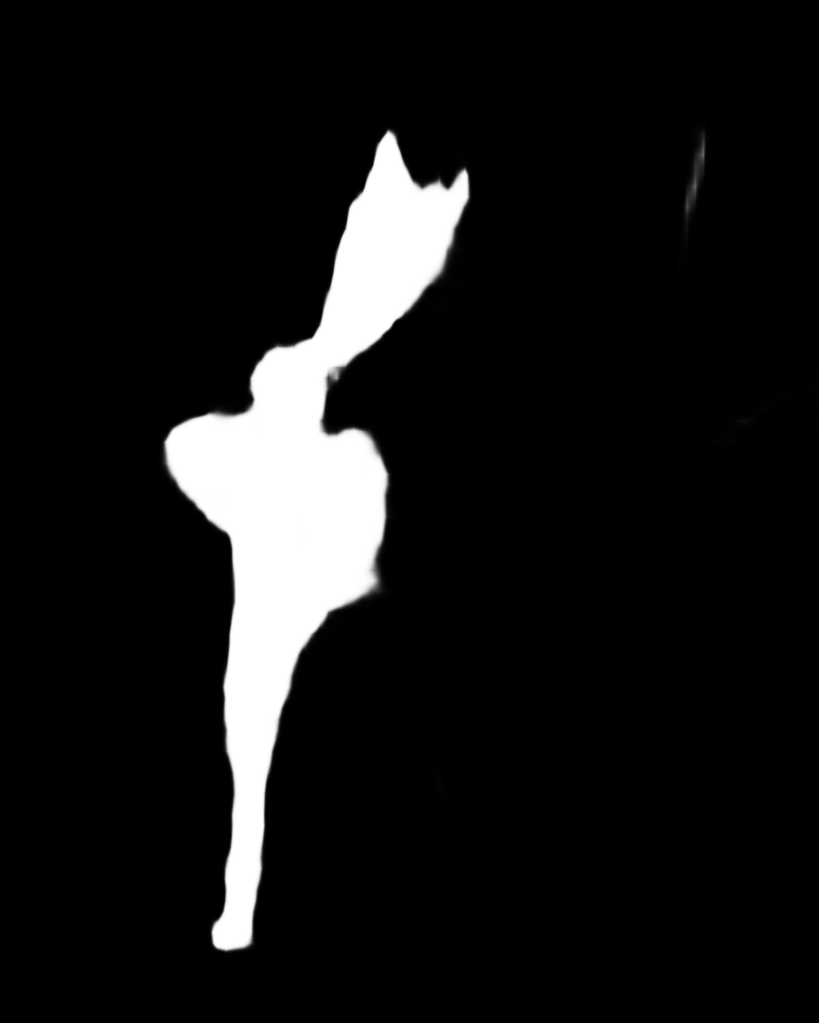} & 
\includegraphics[width=0.118\textwidth]{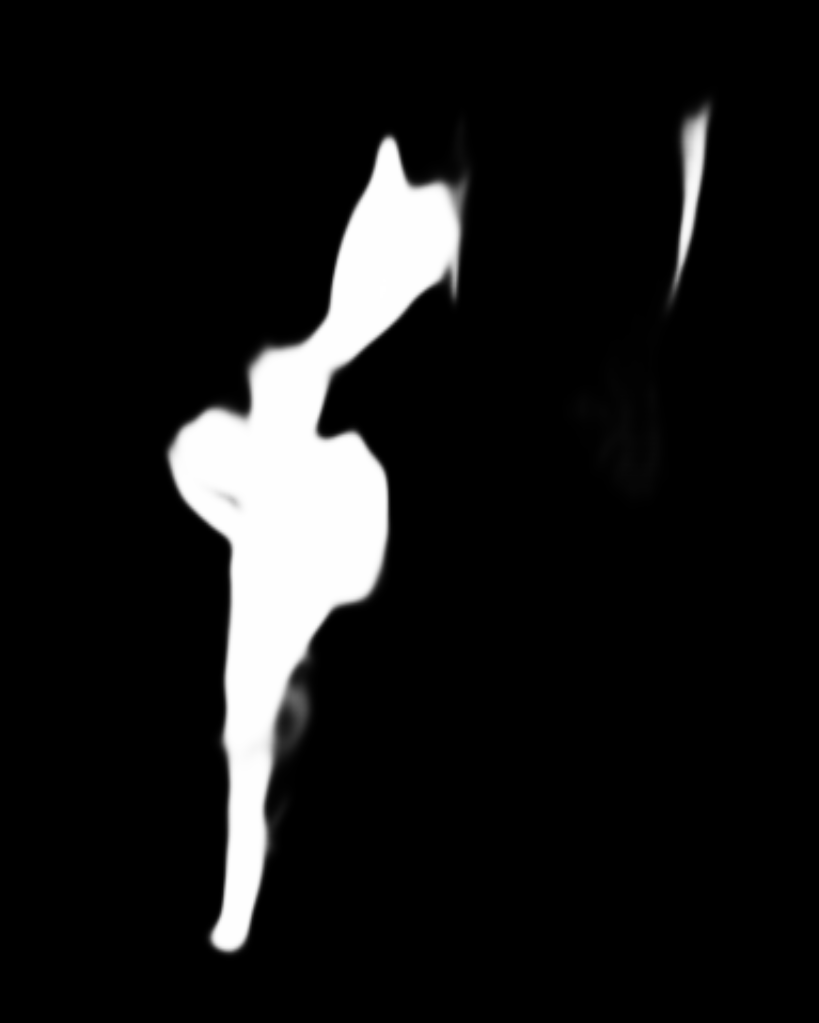} \\

\includegraphics[width=0.118\textwidth]{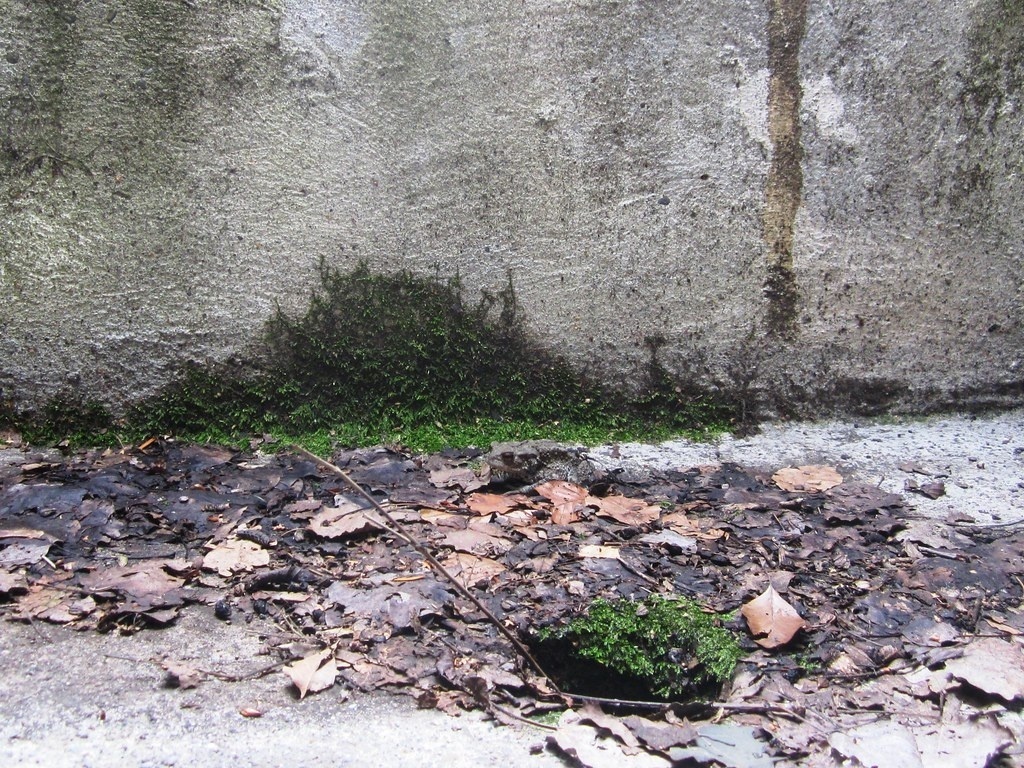} & 
\includegraphics[width=0.118\textwidth]{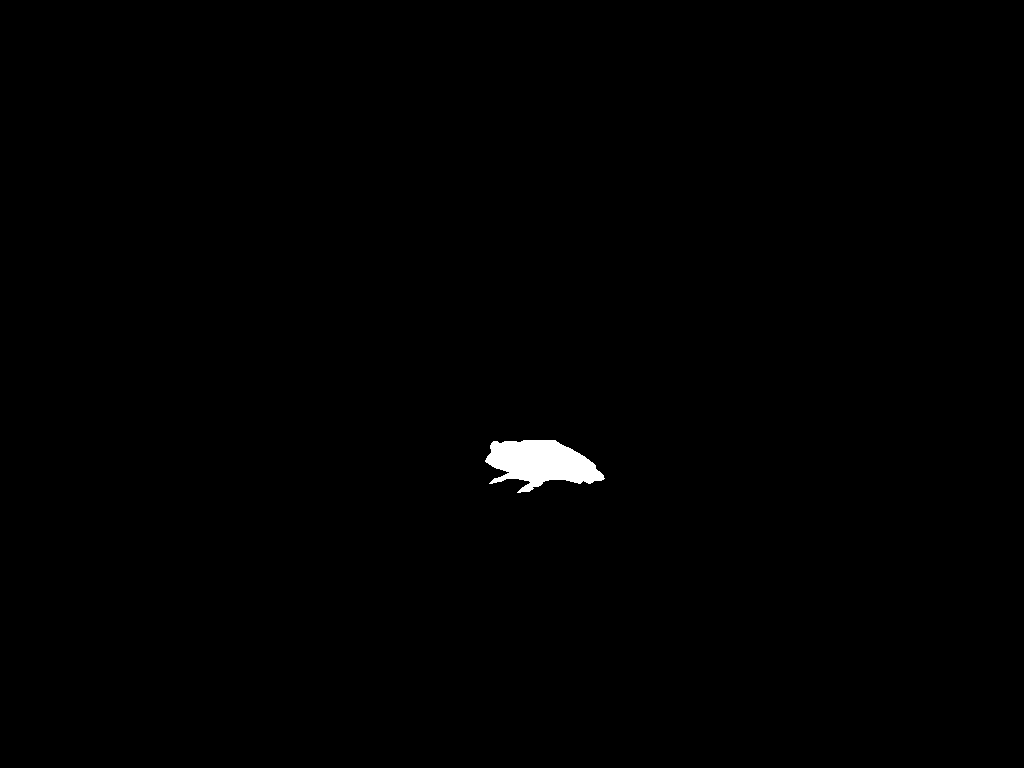} & 
\includegraphics[width=0.118\textwidth]{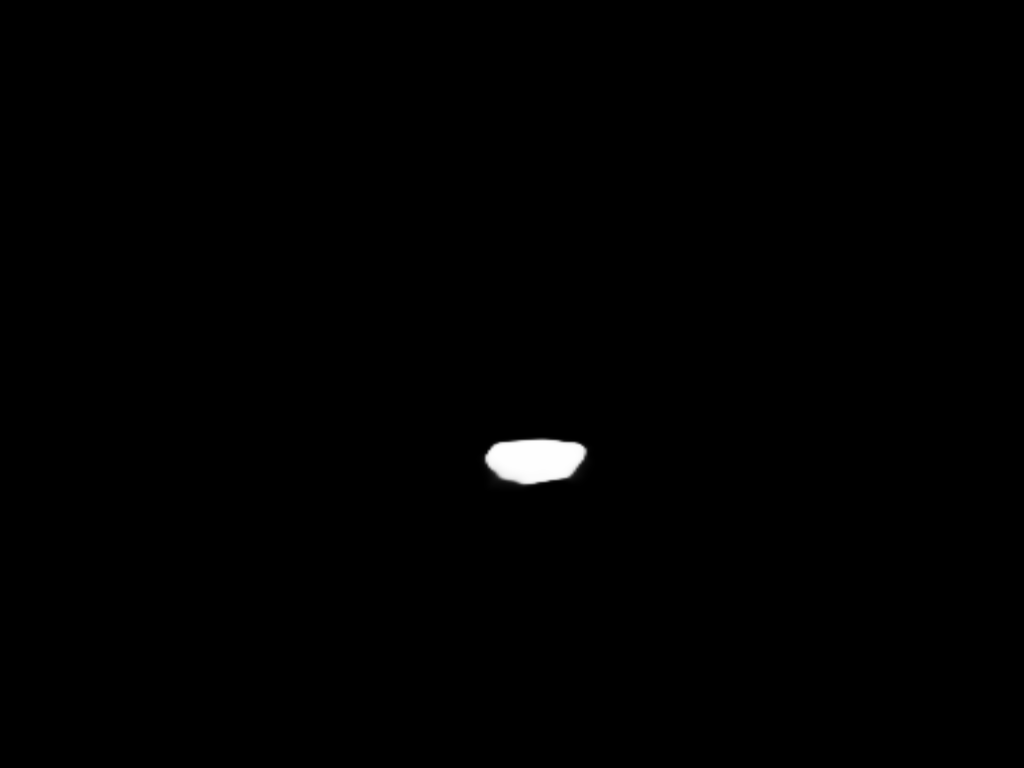} & 
\includegraphics[width=0.118\textwidth]{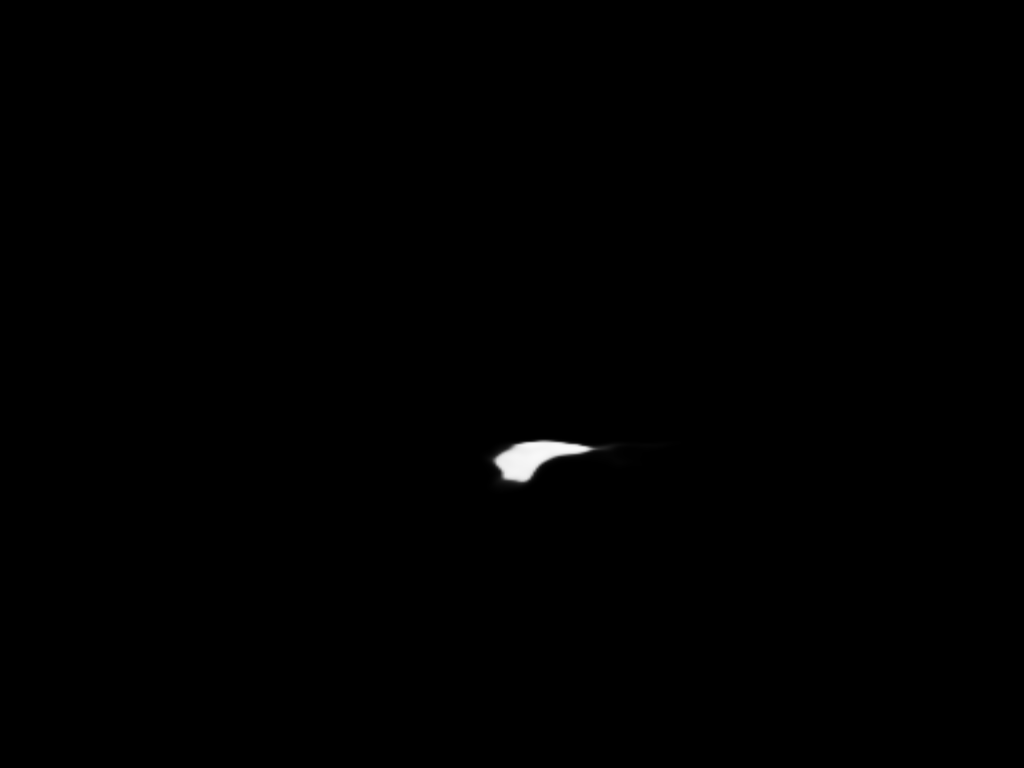} & 
\includegraphics[width=0.118\textwidth]{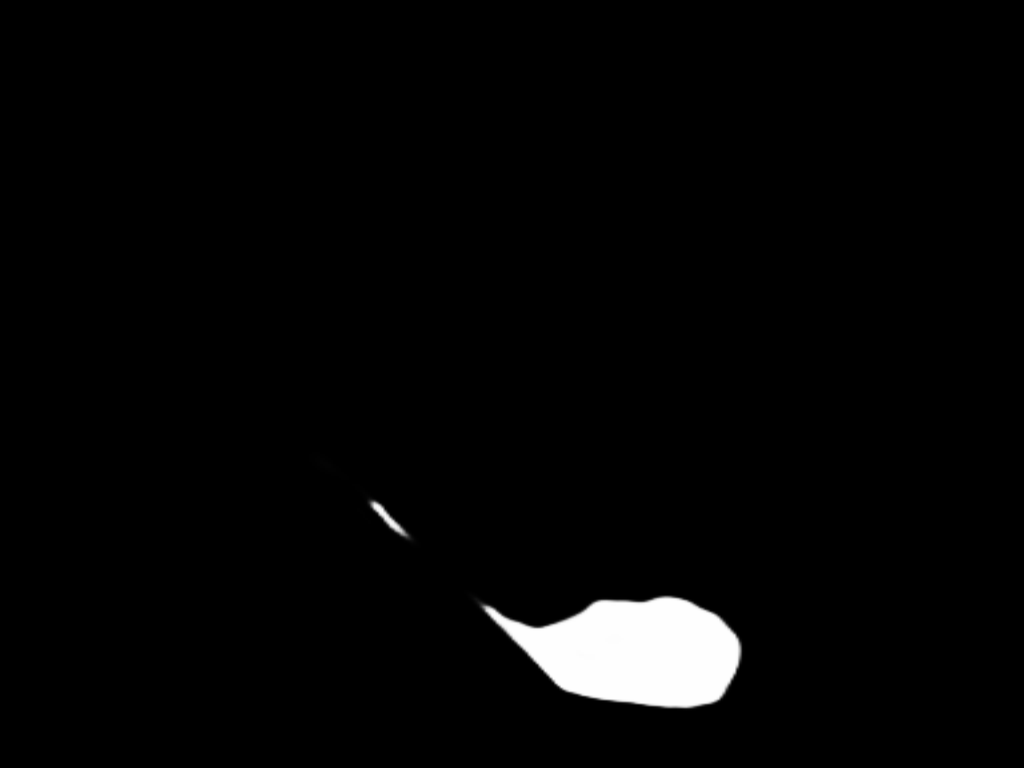} & 
\includegraphics[width=0.118\textwidth]{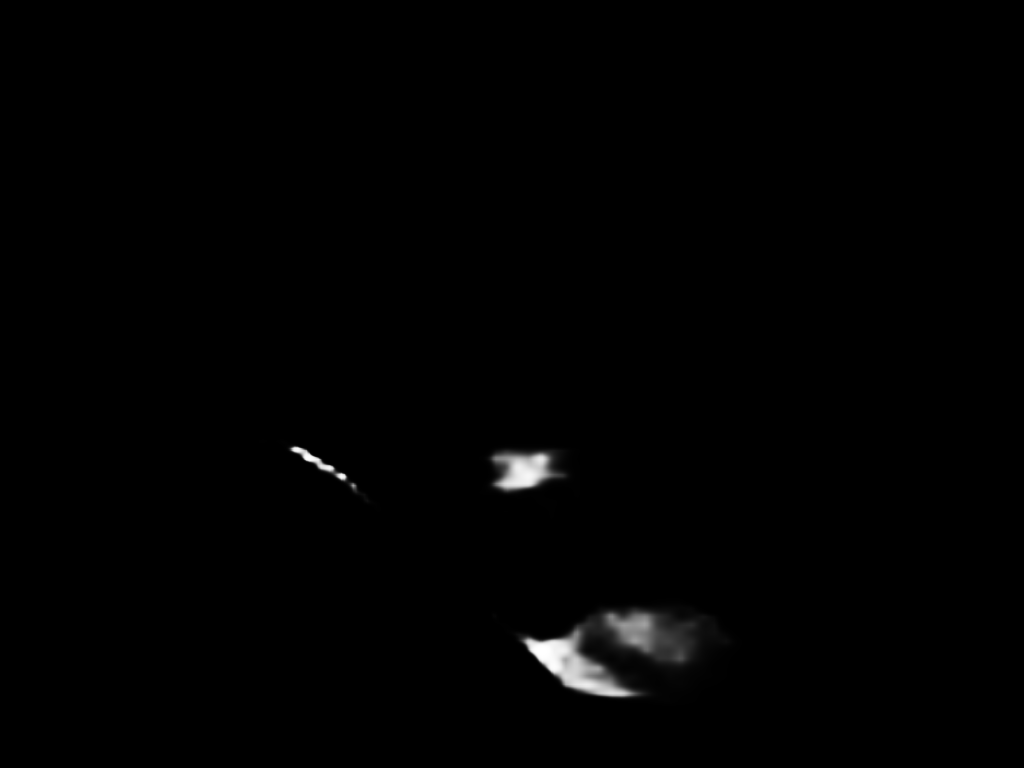} & 
\includegraphics[width=0.118\textwidth]{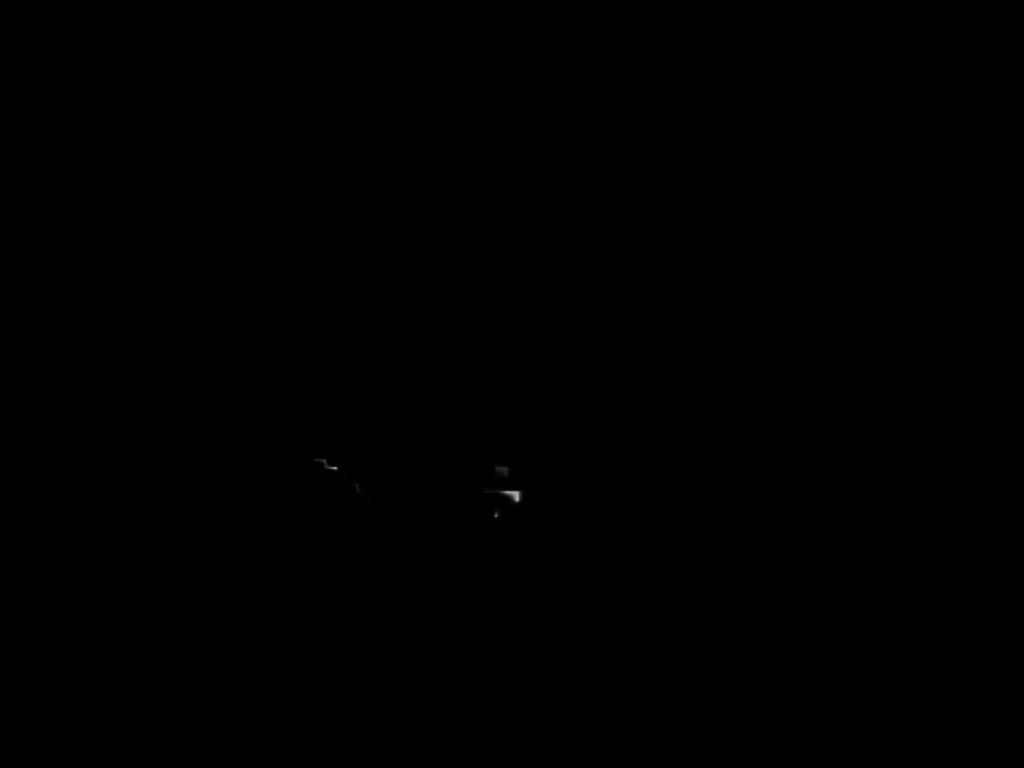} & 
\includegraphics[width=0.118\textwidth]{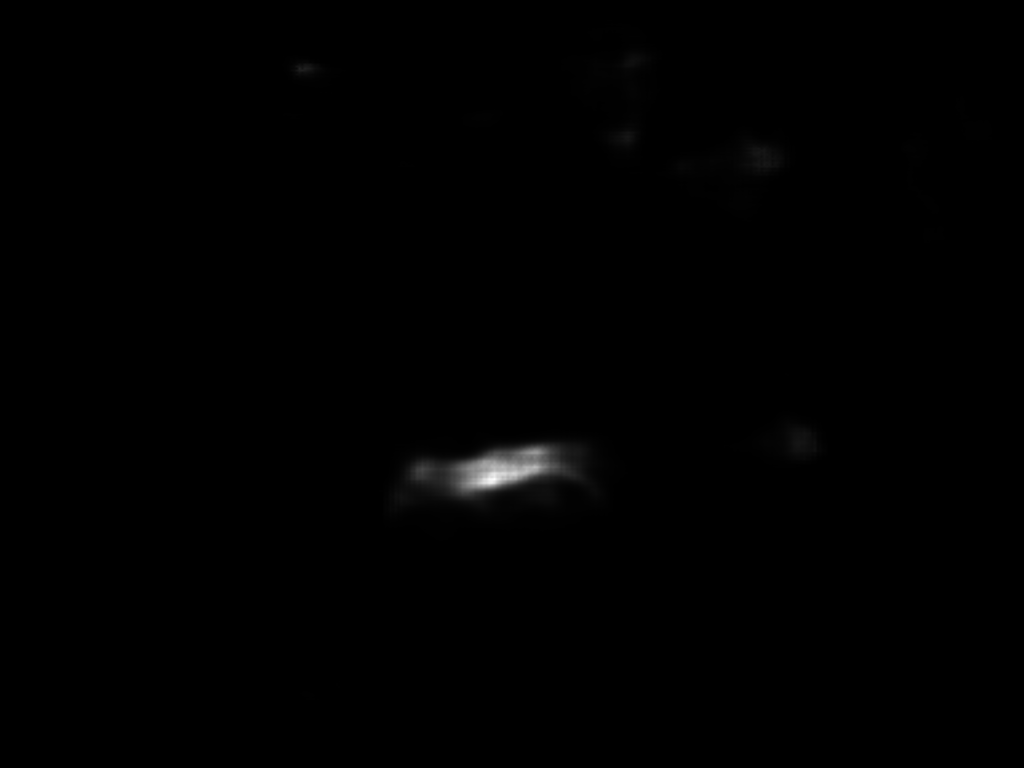} \\

\includegraphics[width=0.118\textwidth]{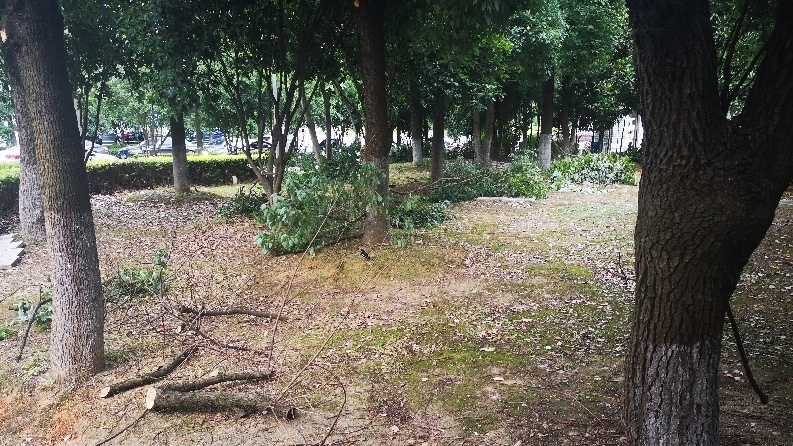} & 
\includegraphics[width=0.118\textwidth]{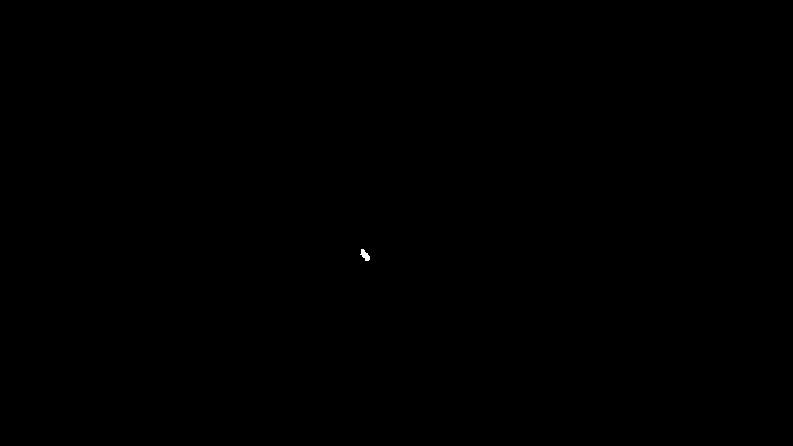} & 
\includegraphics[width=0.118\textwidth]{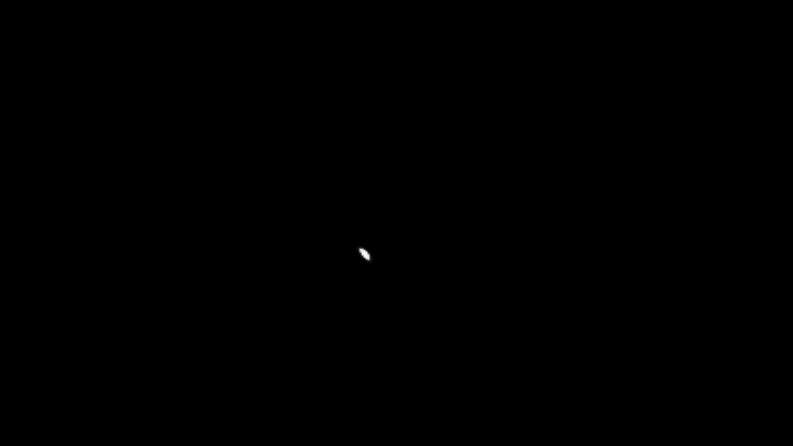} & 
\includegraphics[width=0.118\textwidth]{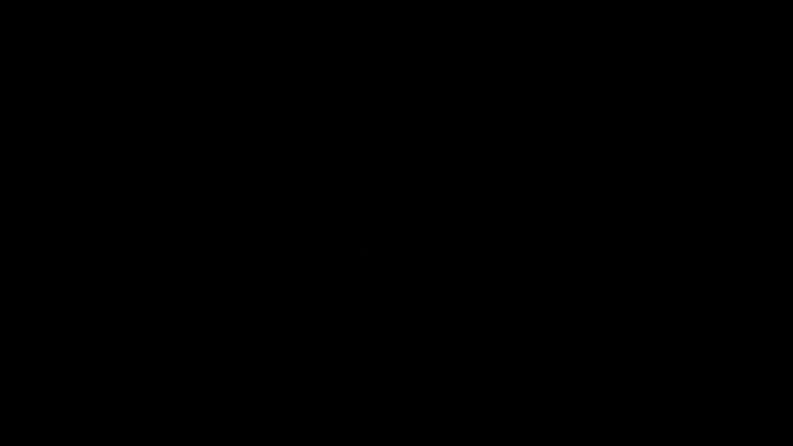} & 
\includegraphics[width=0.118\textwidth]{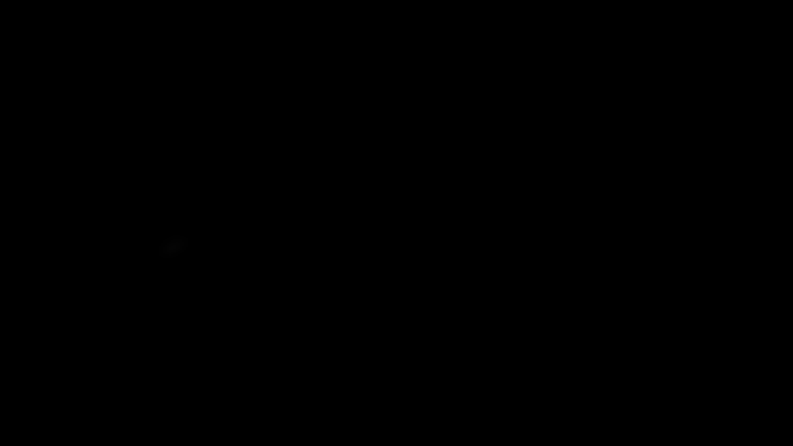} & 
\includegraphics[width=0.118\textwidth]{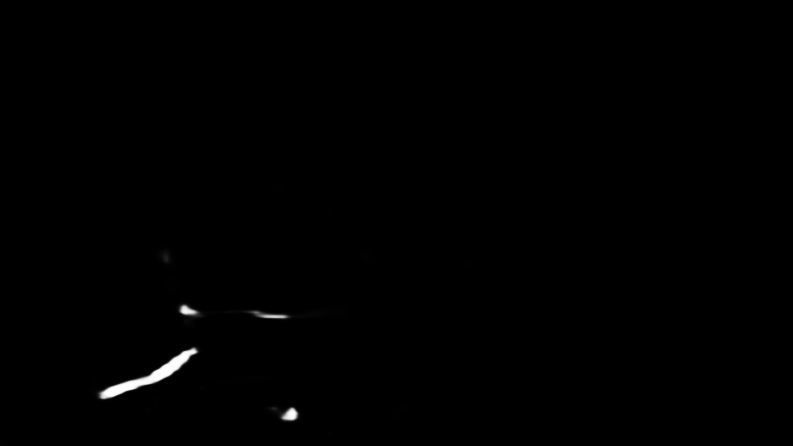} & 
\includegraphics[width=0.118\textwidth]{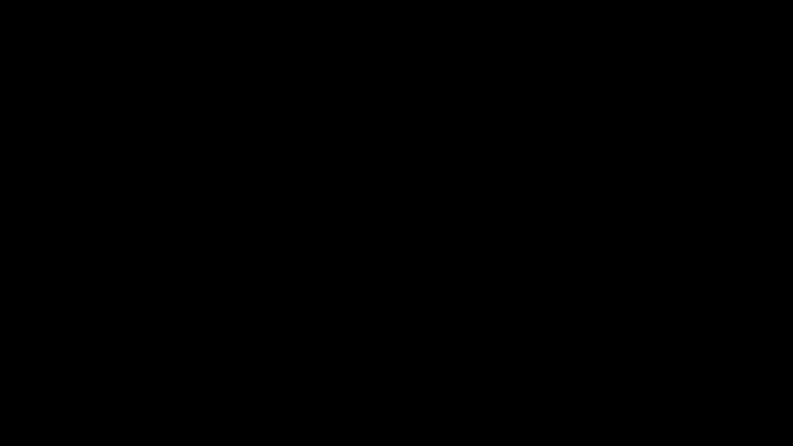} & 
\includegraphics[width=0.118\textwidth]{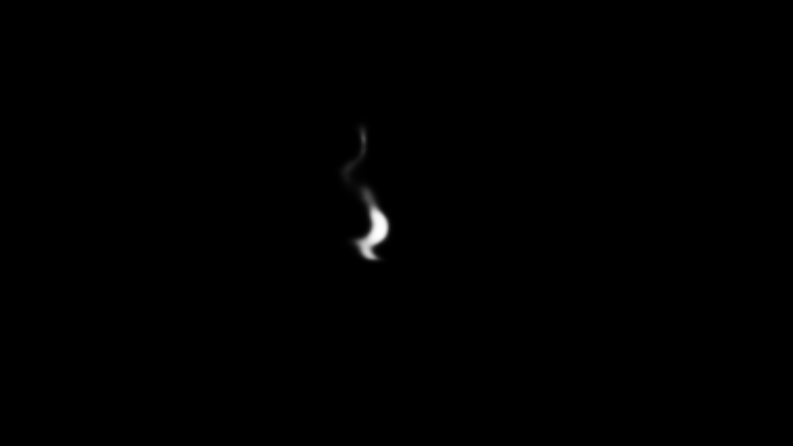} \\
Input & GT & Ours & \cite{pang2024zoomnext} & \cite{xing2023go} & \cite{liu2023mscaf} & \cite{yin2024camoformer} & \cite{huang2023feature}\\ 
\end{tabular}
\caption{Visual Comparison illustrating the superiority of our model in detecting multiple (row 1), small (row 2), and tiny (row 3) camouflaged objects.}
\label{fig:more_comparison_results}
\vspace{-4mm} \end{figure}


Our model was built with the primary objective of overcoming the limitations of SOTA models in detecting small, multiple camouflaged objects. To improve the detection of small objects, we used larger input scales in our multi-scale feature extraction module, thereby enhancing local feature learning. Furthermore, we introduced a novel recursive feedback decoding strategy to strengthen global context learning, enabling our model to better detect multiple objects. \Cref{fig:more_comparison_results} shows the success of our methods, allowing the model to detect multiple objects (row 1), small objects (row 2), and even tiny objects (row 3). This proves the superiority of our model, as it succeeded in complex scenarios where competing SOTA models failed, either wholly or partially.

\begin{figure}[tb]
\centering
\begin{tabular}{cccccccc}
\includegraphics[width=0.118\textwidth]{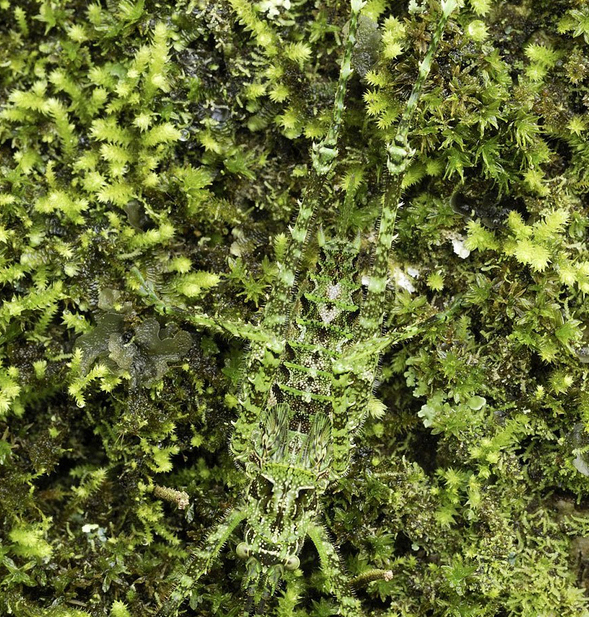} & 
\includegraphics[width=0.118\textwidth]{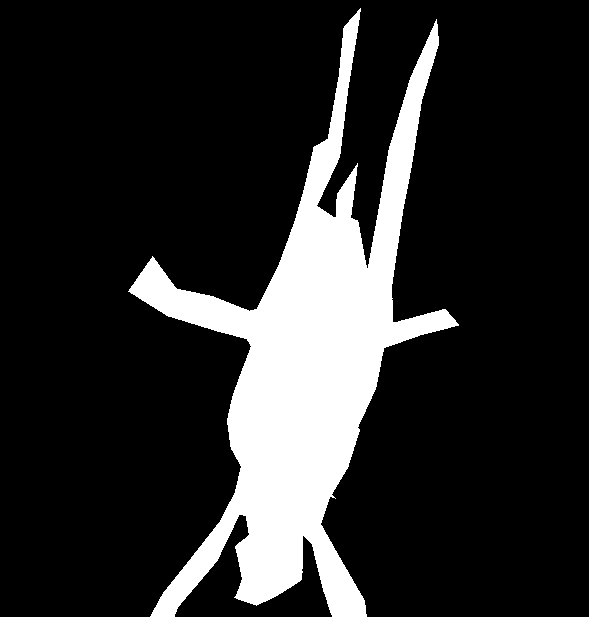} & 
\includegraphics[width=0.118\textwidth]{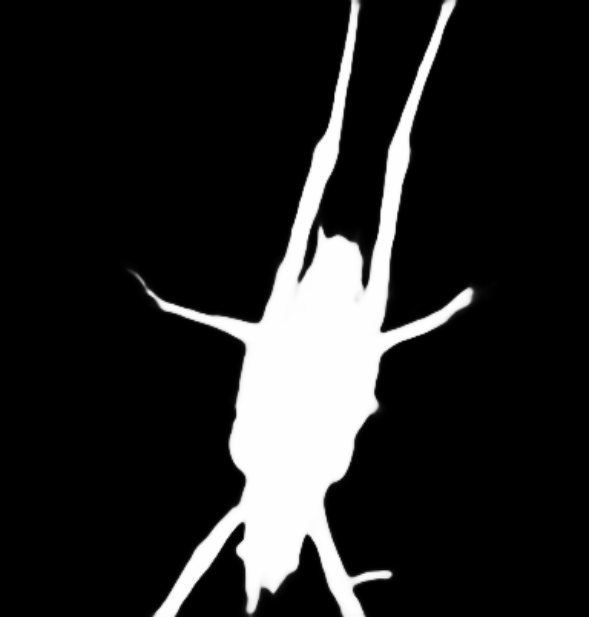} & 
\includegraphics[width=0.118\textwidth]{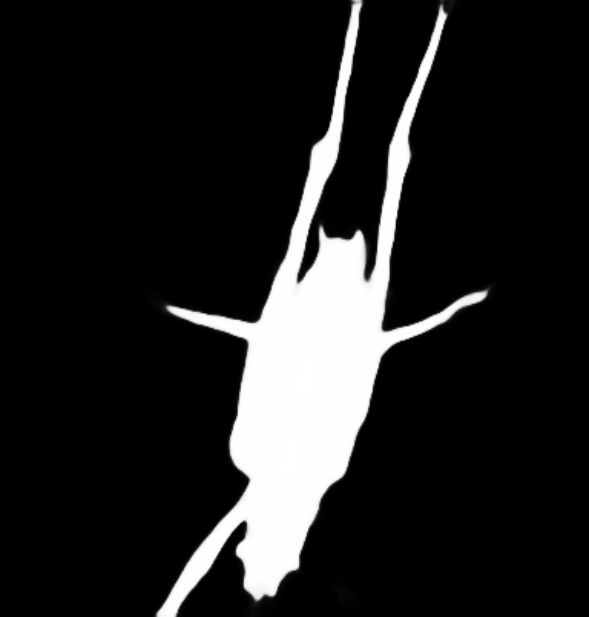} & 
\includegraphics[width=0.118\textwidth]{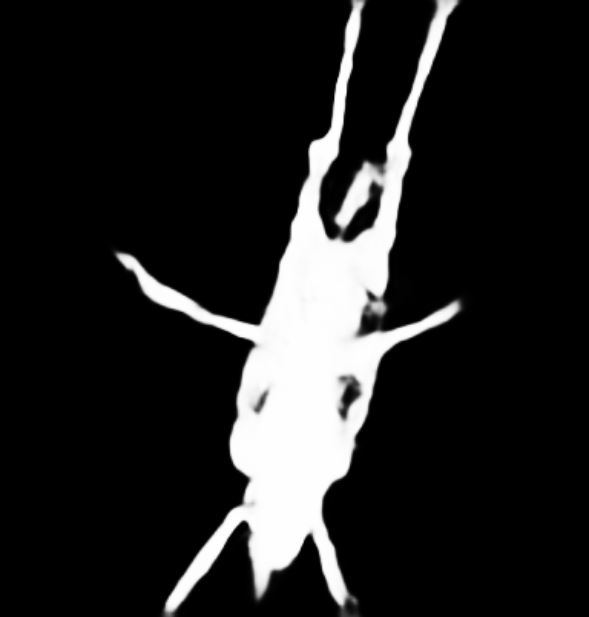} & 
\includegraphics[width=0.118\textwidth]{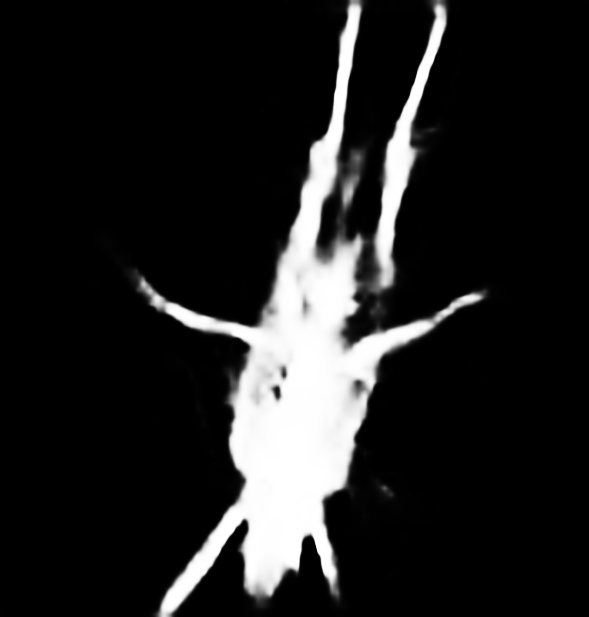} & 
\includegraphics[width=0.118\textwidth]{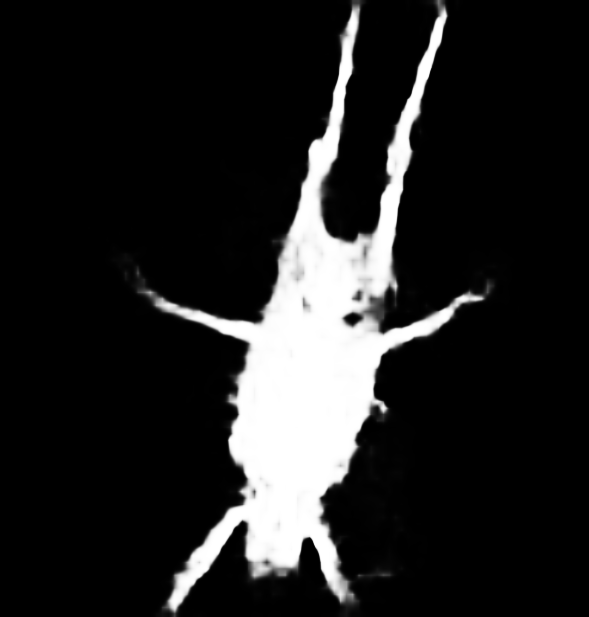} & 
\includegraphics[width=0.118\textwidth]{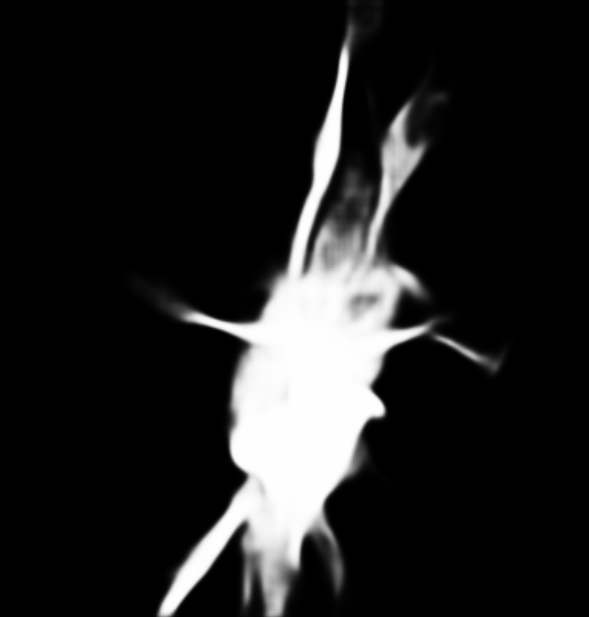} \\

\includegraphics[width=0.118\textwidth]{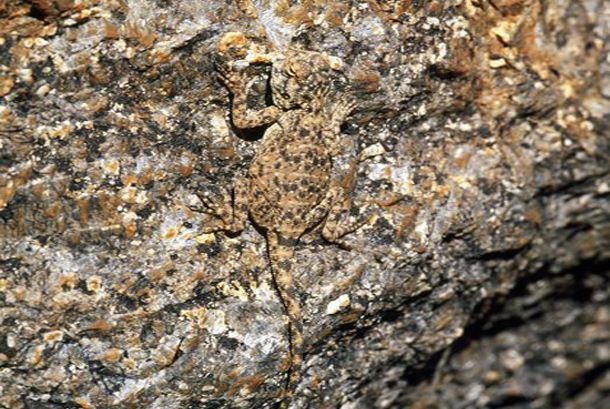} & 
\includegraphics[width=0.118\textwidth]{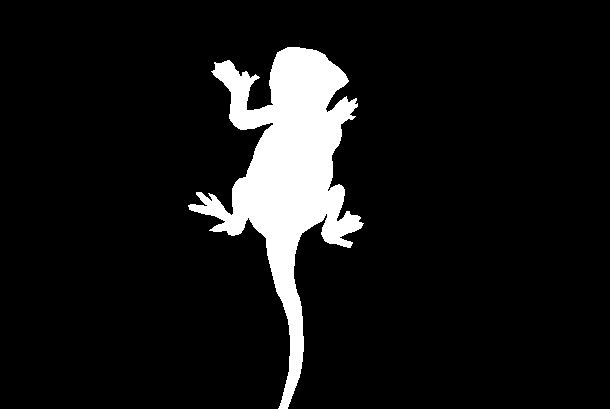} & 
\includegraphics[width=0.118\textwidth]{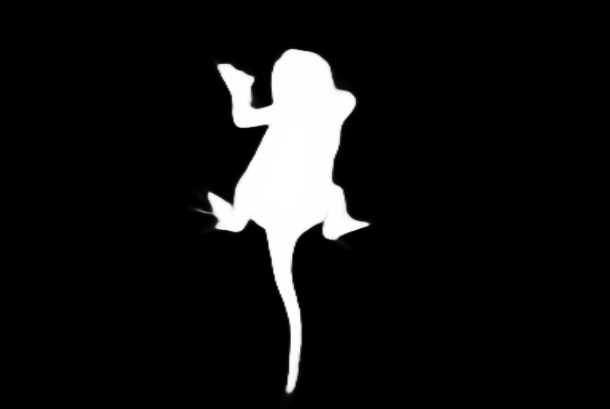} & 
\includegraphics[width=0.118\textwidth]{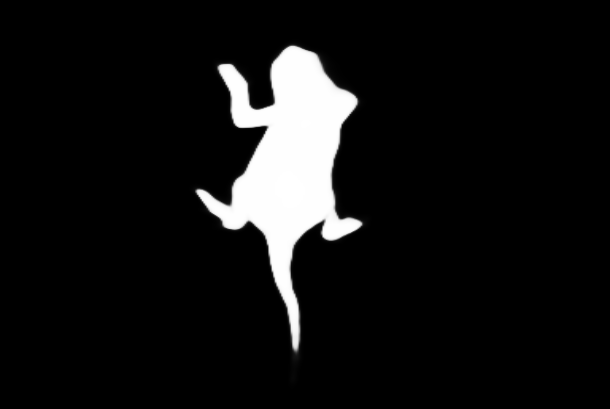} & 
\includegraphics[width=0.118\textwidth]{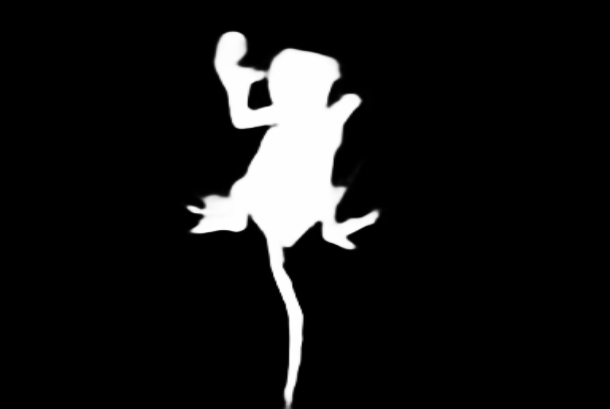} & 
\includegraphics[width=0.118\textwidth]{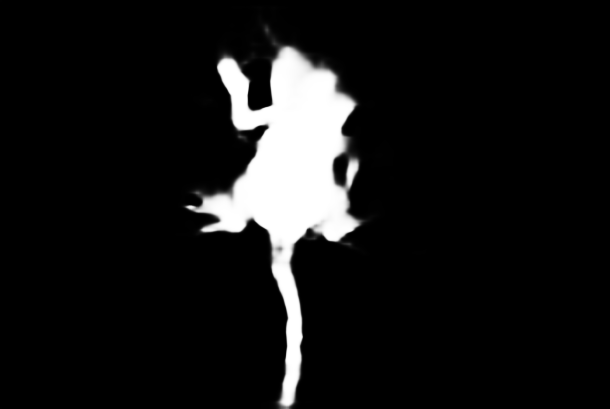} & 
\includegraphics[width=0.118\textwidth]{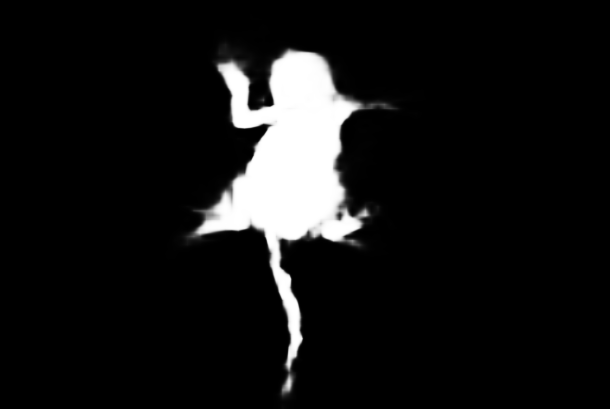} & 
\includegraphics[width=0.118\textwidth]{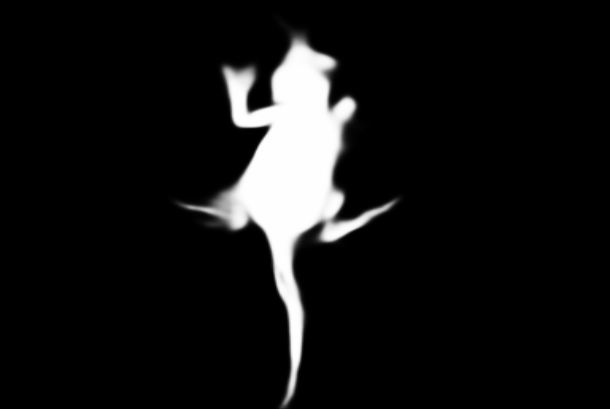} \\

\includegraphics[width=0.118\textwidth]{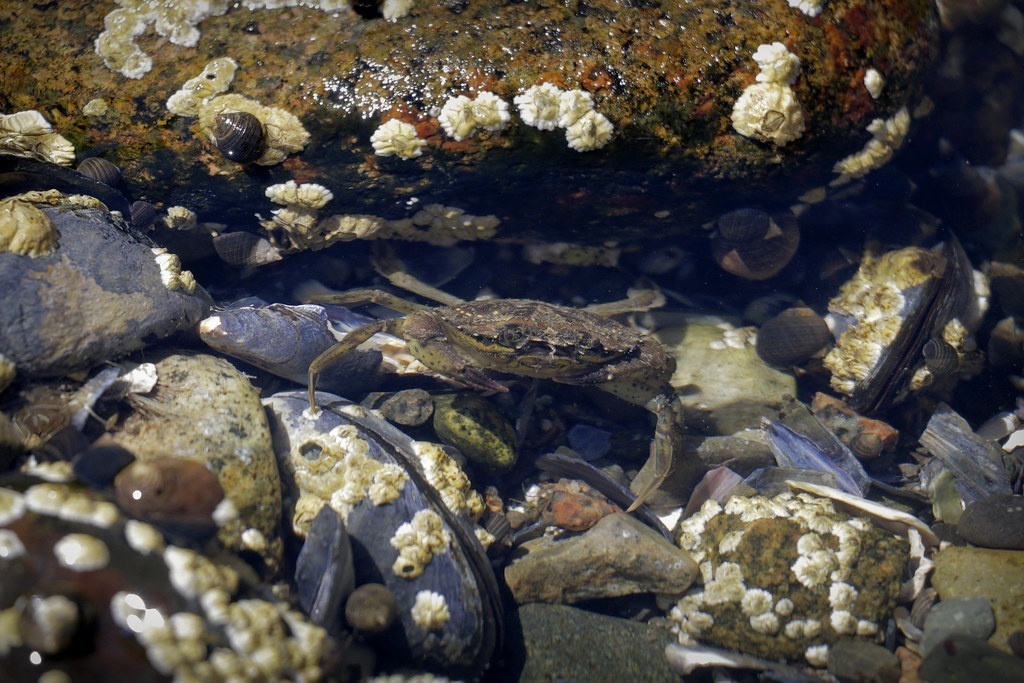} & 
\includegraphics[width=0.118\textwidth]{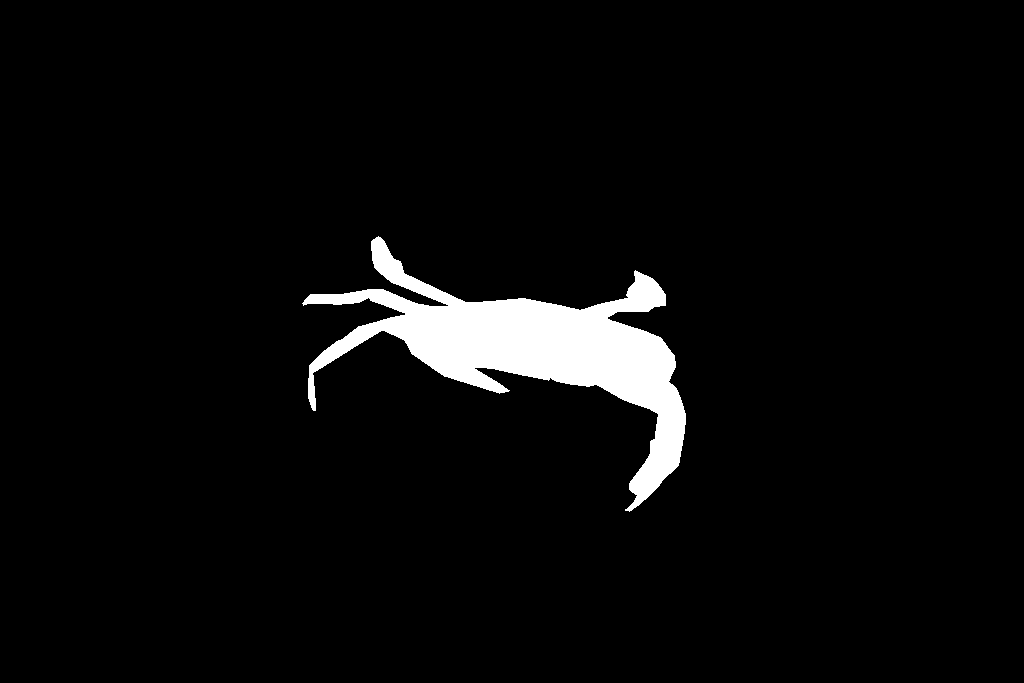} & 
\includegraphics[width=0.118\textwidth]{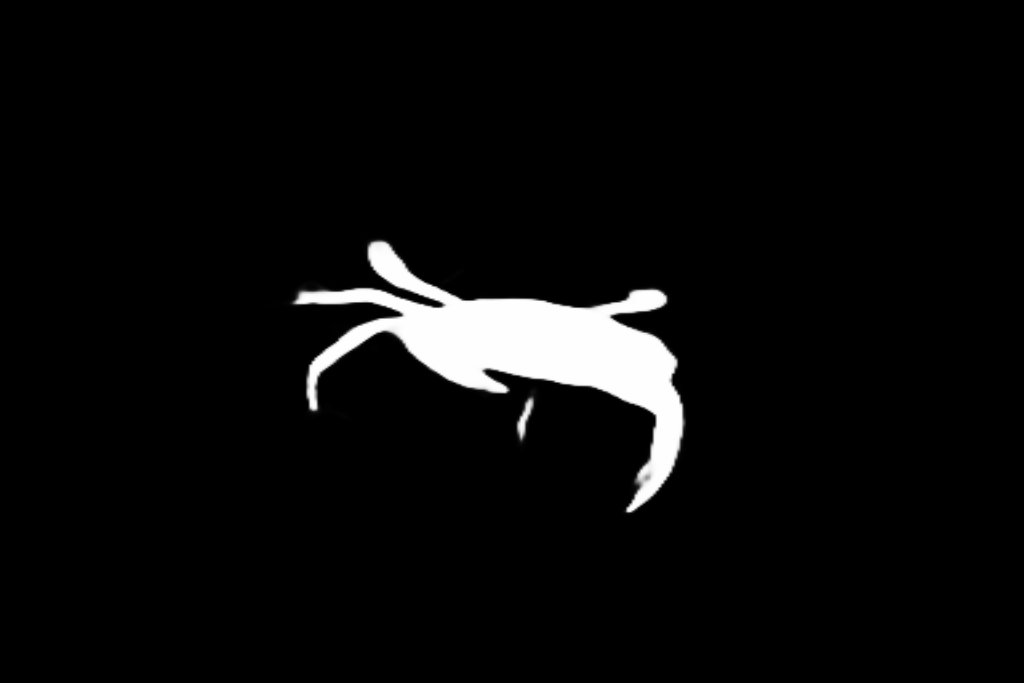} & 
\includegraphics[width=0.118\textwidth]{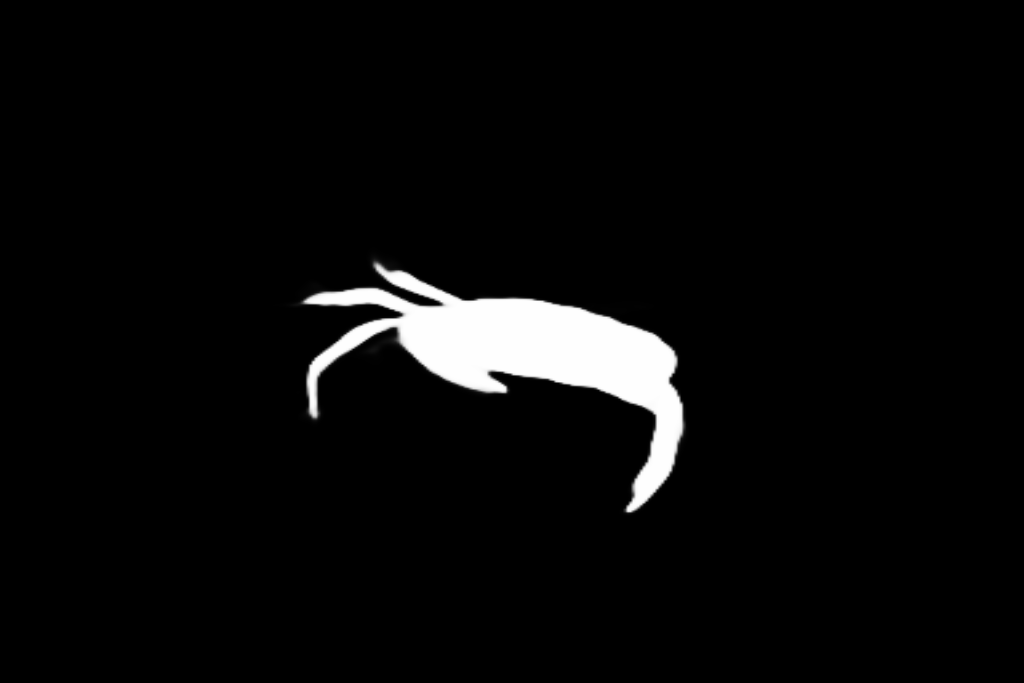} & 
\includegraphics[width=0.118\textwidth]{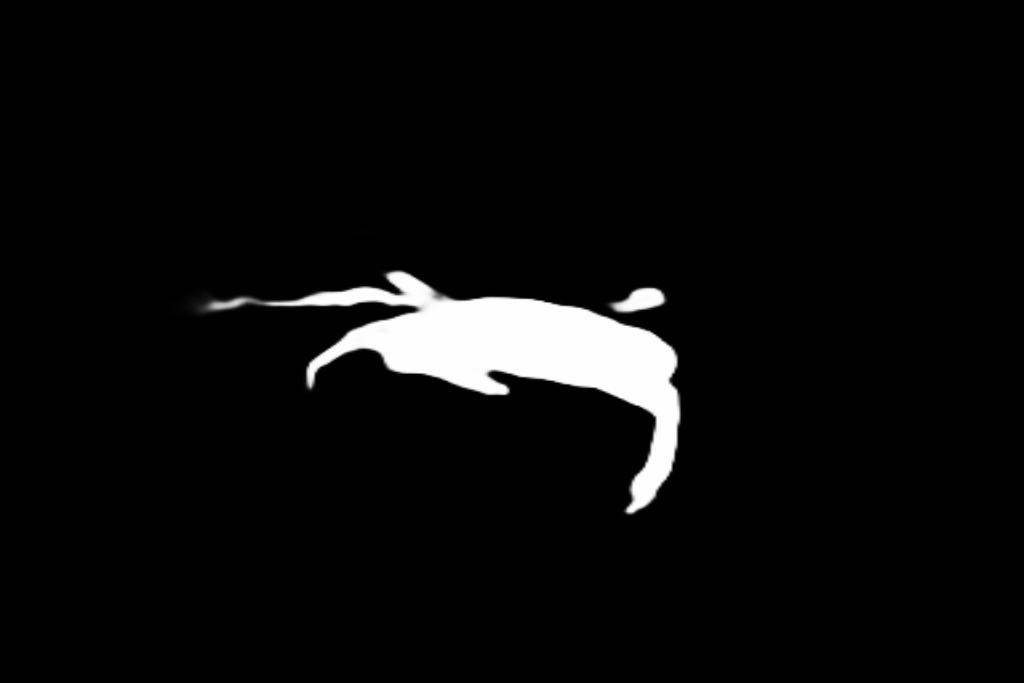} & 
\includegraphics[width=0.118\textwidth]{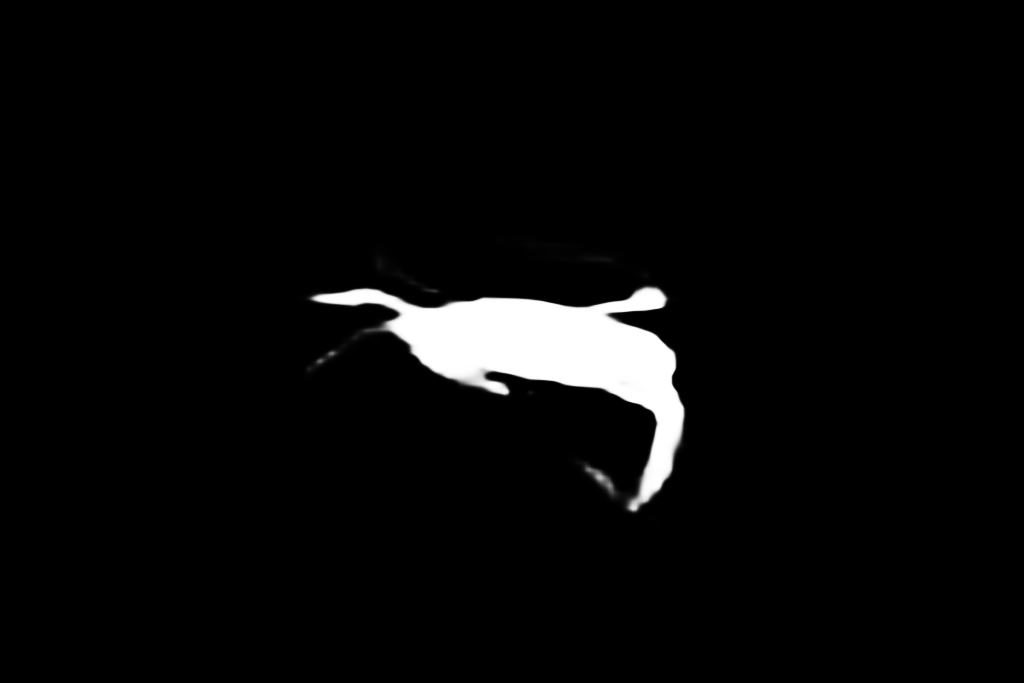} & 
\includegraphics[width=0.118\textwidth]{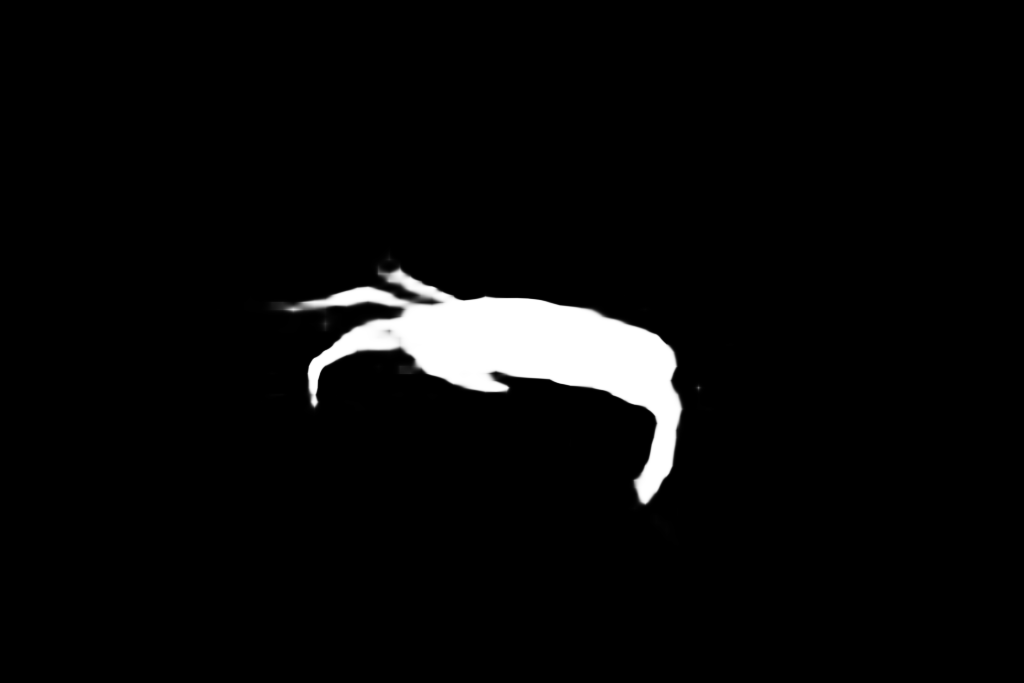} & 
\includegraphics[width=0.118\textwidth]{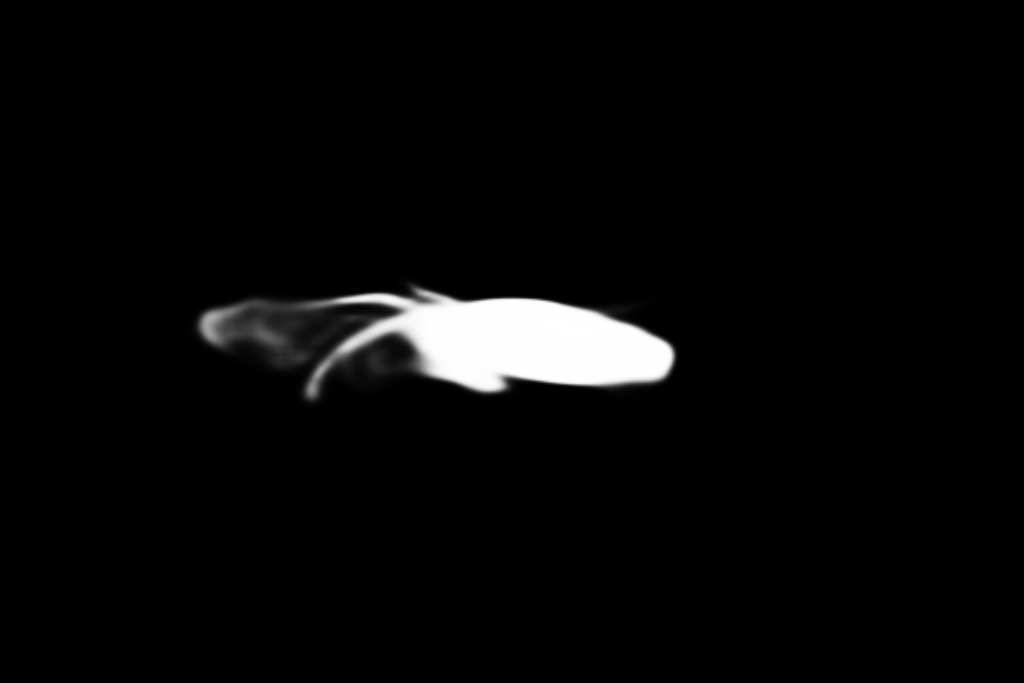} \\

\includegraphics[width=0.118\textwidth]{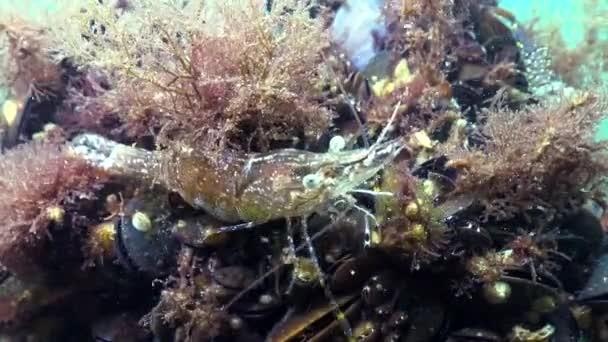} & 
\includegraphics[width=0.118\textwidth]{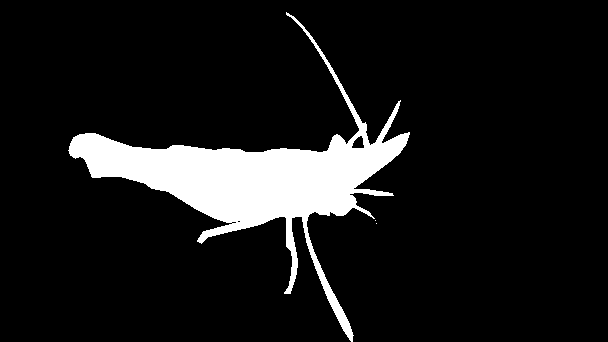} & 
\includegraphics[width=0.118\textwidth]{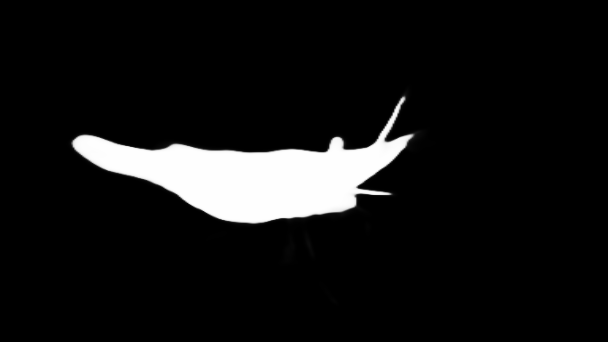} & 
\includegraphics[width=0.118\textwidth]{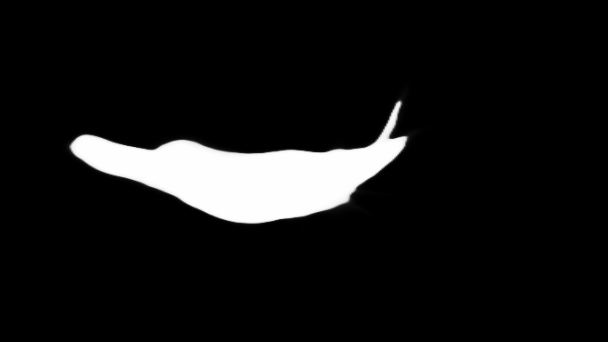} & 
\includegraphics[width=0.118\textwidth]{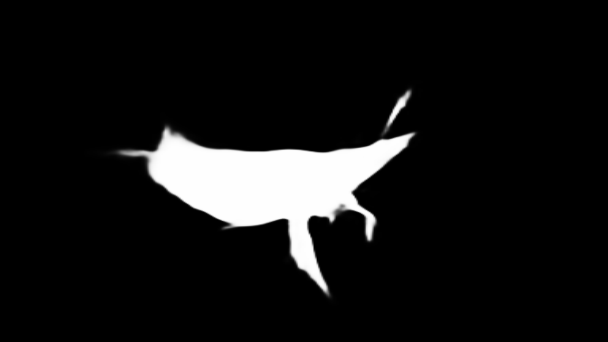} & 
\includegraphics[width=0.118\textwidth]{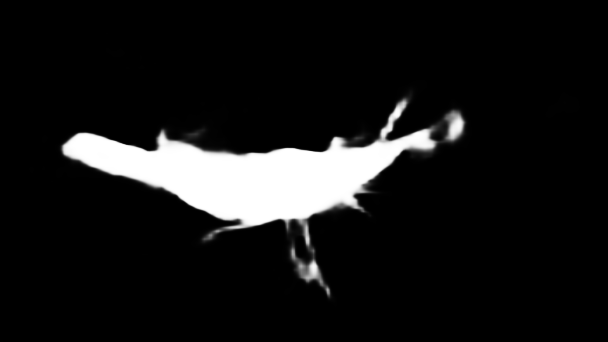} & 
\includegraphics[width=0.118\textwidth]{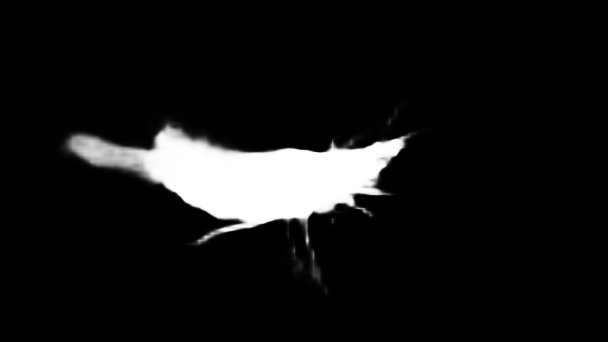} & 
\includegraphics[width=0.118\textwidth]{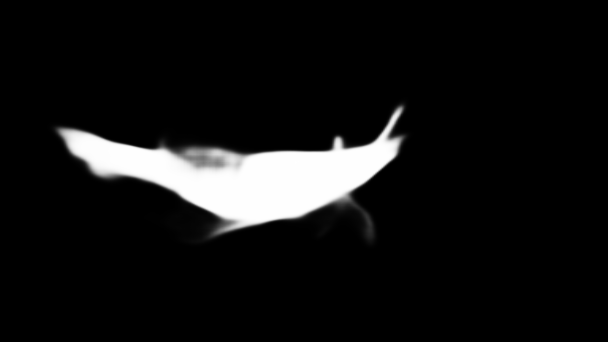} \\
Input & GT & Ours & \cite{pang2024zoomnext} & \cite{xing2023go} & \cite{liu2023mscaf} & \cite{yin2024camoformer} & \cite{huang2023feature}\\ 
\end{tabular}
    \caption{Some instances where our model does not perform detection perfectly compared to the GT. While the model occasionally (a) detects minor false regions (rows 1 and 3), (b) misses small object parts (row 4), or (c) overlooks fine details (row 2), it outperforms SOTA methods on these challenging samples.} 
    \label{fig:FailureCases}
\end{figure}

\noindent\textbf{Challenging Images:} \Cref{fig:FailureCases} shows some instances where our model does not perform detection perfectly. These instances include detecting minor false areas (rows 1 and 3), failing to identify small parts of objects (row 4), and overlooking fine details of objects (row 2). However, these failures are minor, and our model still outperforms SOTA models on identical samples.

\section{Ablation Studies}
In this section, we conduct an ablation study on various components of our model to analyze their impact on overall performance. These components include the encoder, decoder, input shape, and input scales. \Cref{tab:Ablation Studies} shows the results of all experiments conducted in this ablation study.

The base model B0 is fully equipped with all components. This model features PVTv2-B2 as the encoder, an input shape of 352$\times$352, input scales of 0.5$\times$, 1.0$\times$, and 1.5$\times$, and employs a progressive decoding strategy. This approach relies on progressive feature refinement, in which features evolve sequentially through the architecture, enabling minimal cross-resolution interaction and no feedback connections between non-adjacent resolutions. In the following experiments, we modify one component at a time.

We commenced by studying different decoding strategies. For model M1, we implemented a combination of progressive and recursive feedback decoding strategies (as used in MSRNet), in which, for each resolution, decoded features from all preceding resolutions are combined with the current-resolution features. This approach recursively preserves global information from lower-resolution feature maps, enabling strong contextual learning within the network and enabling the detection of multiple camouflaged objects in a single scene. This approach increases the overall performance by {0.21\%} compared to the base model. For model M2, we adopted a dense progressive decoding strategy. In this context, before refining the features at each resolution, all features from preceding resolutions are combined with those at the current resolution. This methodology resulted in a performance decline of {-0.73\%}. This outcome may be attributed to the potential disruption of their meticulously learned scale-specific features when raw features from non-adjacent resolutions are combined prior to refinement. For instance, coarse-level features at lower resolutions may overpower fine features at higher resolutions. We implemented a combination of both decoding strategies in model M3, employing dense-progressive and recursive-feedback decoding. This strategy led to a performance degradation of {-0.35\%}, indicating a smaller decline than M2. These experiments demonstrate the effectiveness of the recursive feedback decoding strategy in improving overall model performance. They also illustrate the negative impact of aggregating unrefined multi-resolution features, particularly those from non-adjacent resolutions, where features operate at different abstraction levels and might conflict. Consequently, in the subsequent experiments, we shall build upon the recursive-feedback decoding strategy implemented in M1 to evaluate other model components. 

Furthermore, we investigated the impact of altering the input shape in M4. This experiment employed a larger input shape (384$\times$384), yielding a performance improvement of {1.42\%}.  Larger input shapes correspond to higher resolutions, helping preserve fine details that might otherwise be lost. This is a crucial characteristic for detecting camouflaged objects. Regarding the encoder, we analyze the effects of different backbone architectures on performance. In model M5, we employ the PVTv2-B3 backbone, which improves performance by {3.01\%} over the base model. The PVTv2-B4 in M6 improves performance by {4.63\%}, while PVTv2-B5 in M7 increases performance by {3.41\%}. These experiments indicate that PVTv2-B4 delivers the best results.

We conducted experiments using various input scale sets. The M8 model employed input scales of 1.0$\times$, 1.5$\times$, and 1.7$\times$, leading to a performance increase of {4.54\%}. Model M9 utilized input scales of 1.0$\times$, 1.5$\times$, and 2.0$\times$, achieving the highest performance increase of {5.12\%}. The results from M9 indicate that using higher input scales enhances overall performance. This is because higher scales correspond to higher resolutions, enabling the model to identify finer features, which is essential for detecting small and tiny camouflaged objects. Furthermore, model M8 illustrates the importance of using well-distributed scales. Although the scales in this experiment are higher than those of M6, they generated lower performance because the proximity of the scales (e.g., 1.5 and 1.7) may lead to similar feature representations, consequently failing to provide novel information to the network. 

\begin{table}[t]
  \centering
  \caption{Ablation study on various model components, including decoder, input shape, encoder, and input scales. \enquote{RFD} stands for Recursive-Feedback Decoding, \enquote{DPD} means Dense Progressive Decoding, \enquote{DRFD} denotes Dense Recursive-Feedback Decoding, and \enquote{IS} stands for input size and input scales}
  \label{tab:Ablation Studies}
  \resizebox{1.0\textwidth}{!}
  {
  \begin{tabular}{l|l|c|ccccc|ccccc|ccccc|ccccc|c}
    \hline
      &   &  &\multicolumn{5}{c|}{\textbf{CAMO}}& \multicolumn{5}{c|}{\textbf{CHAMELEON}}& \multicolumn{5}{c|}{\textbf{COD10K}}& \multicolumn{5}{c|}{\textbf{NC4K}} & \\\cline{4-23}
      
      \textbf{No}& \multicolumn{1}{c|}{\textbf{Models}}& \textbf{Params} & $S_{m}$$\uparrow$& $F^{\omega}_{\beta}$$\uparrow$& MAE$\downarrow$& $F_{\beta}$$\uparrow$&  $E_{m}$$\uparrow$ & $S_{m}$$\uparrow$& $F^{\omega}_{\beta}$$\uparrow$& MAE$\downarrow$& $F_{\beta}$$\uparrow$&  $E_{m}$$\uparrow$  & $S_{m}$$\uparrow$& $F^{\omega}_{\beta}$$\uparrow$& MAE$\downarrow$& $F_{\beta}$$\uparrow$&  $E_{m}$$\uparrow$ & $S_{m}$$\uparrow$& $F^{\omega}_{\beta}$$\uparrow$& MAE$\downarrow$& $F_{\beta}$$\uparrow$&  $E_{m}$$\uparrow$ & \textbf{$\Delta$} \\
    \hline

\textbf{B0} & Baseline & 28.18M &
    0.868& 0.829& 0.049& 0.855& 0.926&
    0.916& 0.876& 0.018& 0.889& 0.971&
    0.881& 0.809& 0.020& 0.834& 0.945&
    0.890& 0.848& 0.031& 0.872& 0.941&
   \textbf{0.0\%}\\
    
 \textbf{M1} & B0 + RFD & 28.18M&
    0.866& 0.829& 0.049& 0.854& 0.925&
    0.918& 0.881& 0.018& 0.892& 0.969&
    0.883& 0.810& 0.020& 0.834& 0.943&
    0.891& 0.850& 0.030& 0.872& 0.941&
   \textbf{$\uparrow$ 0.21\%}\\
   
    \textbf{M2} & B0 + DPD & 28.18M  &
    0.867& 0.826& 0.049& 0.850& 0.923&
    0.914& 0.874& 0.020& 0.888& 0.962& 
    0.883& 0.810& 0.020& 0.834& 0.942& 
    0.890& 0.846& 0.031&0.869& 0.940&
   \textbf{$\downarrow$ 0.73\%} \\
   
   \textbf{M3} & B0 + DRFD & 28.18M & 
    0.867& 0.828& 0.049& 0.853& 0.925&
    0.915& 0.877& 0.019& 0.890& 0.970& 
    0.883& 0.809& 0.020& 0.832& 0.943& 
    0.890& 0.847& 0.031& 0.869& 0.940&
    \textbf{$\downarrow$ 0.35\%}\\

   \textbf{M4} & M1 + IS (384$\times$384) &
    28.18M&
    0.873& 0.834& 0.046& 0.856& 0.930&
    0.923& 0.888& 0.017& 0.900& 0.977&
    0.886& 0.816& 0.019& 0.839& 0.945&
    0.891& 0.851& 0.030& 0.874& 0.942&
   \textbf{$\uparrow$ 1.42\%}\\

   \textbf{M5} & M4 + PVTv2-B3 & 48.06M &
    0.880& 0.849& 0.043& 0.869& 0.938&
    0.926& 0.897& 0.017& 0.905& 0.975 &
    0.893& 0.828& 0.018& 0.848& 0.951 &
    0.898& 0.861& 0.028& 0.881& 0.948 &
   \textbf{$\uparrow$ 3.01\%}\\

   \textbf{M6} & M4  + PVTv2-B4& 65.37M &
    0.888& 0.862& 0.039& 0.881& 0.947 &
    0.929& 0.902& 0.016& 0.911& 0.980&
    0.900& 0.842& 0.017& 0.861& 0.958&
    0.901& 0.867& 0.028& 0.886& 0.951&
   \textbf{$\uparrow$ 4.63\%}\\
   
   \textbf{M7} & M4 + PVTv2-B5& 84.77M &
    0.883& 0.854& 0.042& 0.871& 0.942 &
    0.922& 0.888& 0.017& 0.896& 0.977 &
    0.894& 0.831& 0.018& 0.850& 0.954 &
    0.902& 0.869& 0.027& 0.886& 0.952 &
   \textbf{$\uparrow$ 3.41\%}\\

   \textbf{M8} & M6 + IS (1, 1.5, 1.7) & 65.37M &
    0.889& 0.862& 0.040& 0.880& 0.945& 
    0.929& 0.905& 0.017& 0.915& 0.974&  
    0.904& 0.848& 0.017& 0.865& 0.962& 
    0.903& 0.869& 0.027& 0.887& 0.951&
   \textbf{$\uparrow$ 4.54\%}\\

  \textbf{M9} & M6 + IS (1, 1.5, 2) & 65.37M &
    0.887& 0.859& 0.041& 0.877& 0.944&  
    0.934& 0.911& 0.016& 0.918& 0.979& 
    0.908& 0.853& 0.016& 0.868& 0.962& 
    0.904& 0.871& 0.027& 0.889& 0.951&
   \textbf{$\uparrow$ 5.12\%}\\
\hline
      \end{tabular}}
\end{table}

\section{Conclusion}
\label{sec:conclusion}
In this paper, we propose a transformer-based multi-scale recursive network (MSRNet) to address the challenges of detecting small, multiple camouflaged objects. Our approach extracts multi-scale features via a pyramid vision transformer and selectively merges them with specialized Attention-Based Scale Integration Units. To further enhance feature representations, we introduce Multi-Granularity Fusion Units. A novel recursive feedback decoding strategy that preserves global information is developed to enable detection of multiple objects. We employ large input scales to improve the learning of fine features, allowing the detection of small objects. Extensive experiments across four benchmark COD datasets with 20 SOTA models show that our model achieves SOTA performance on two datasets and ranks second on the other two. Moreover, visual results highlight our model's superior ability to detect small, camouflaged, and multiple objects. Despite our model's robust performance, extracting multi-scale features requires additional computational resources. Furthermore, our model is presently designed to apply COD to static images. Future work could focus on optimizing the computational efficiency of the feature-extraction methodology and on investigating the possibility of extending the model to encompass video-based COD.




\bibliographystyle{splncs04}
\bibliography{main}
\end{document}

%% file: main-table/main-table.tex
\begin{table*}[t]
  \centering
  \caption{Results of different models based on different backbones on static image COD datasets. The highest three results are colored in \textcolor{red!80!black}{red} (1st), \textcolor{green!60!black}{green} (2nd), and \textcolor{blue!90!black}{blue} (3rd). Best viewed on screen when zoomed in.}
  \label{tab:results}
  \resizebox{1.0\textwidth}{!}
  {
  \begin{tabular}{l l|c c|ccccc|ccccc|ccccc|ccccc}
    \hline
      &   &  &  & \multicolumn{5}{c}{\textbf{CAMO}}& \multicolumn{5}{c}{\textbf{CHAMELEON}}& \multicolumn{5}{c}{\textbf{COD10K}}& \multicolumn{5}{c}{\textbf{NC4K}} \\
      
     \textbf{Model} & \textbf{Backbone}  & \textbf{Input Size} & \textbf{\#Params} & $S_{m}$$\uparrow$& $F^{\omega}_{\beta}$$\uparrow$& MAE$\downarrow$& $F_{\beta}$$\uparrow$&  $E_{m}$$\uparrow$ & $S_{m}$$\uparrow$& $F^{\omega}_{\beta}$$\uparrow$& MAE$\downarrow$& $F_{\beta}$$\uparrow$&  $E_{m}$$\uparrow$  & $S_{m}$$\uparrow$& $F^{\omega}_{\beta}$$\uparrow$& MAE$\downarrow$& $F_{\beta}$$\uparrow$&  $E_{m}$$\uparrow$ & $S_{m}$$\uparrow$& $F^{\omega}_{\beta}$$\uparrow$& MAE$\downarrow$& $F_{\beta}$$\uparrow$&  $E_{m}$$\uparrow$ \\
    
    \hline
    \multicolumn{24}{c}{\textbf{{Convolutional Neural Network based Methods}}} \\
    \hline
    SINet~\cite{fan2020camouflaged} & ResNet-50~\cite{he2016deep} & 352$\times$352 & 48.947M & 0.745 &0.644& 0.091& 0.702& 0.829&
    0.872& 0.806& 0.034& 0.827& 0.946&
    0.776& 0.631& 0.043& 0.679& 0.874 &
    0.808& 0.723& 0.058& 0.769& 0.883\\
    C2FNet~\cite{sun2021context} & ResNet-50~\cite{he2016deep} & 352$\times$352 & 28.411M & 0.796 &0.719 &0.080 &0.762 &0.864& 0.888 &0.828 &0.032 &0.844 &0.946& 0.813 &0.686 &0.036 &0.723 &0.900 &0.838 &0.762 &0.049 &0.794 &0.904\\
    SINetV2~\cite{fan2021concealed} & Res2Net-50~\cite{gao2019res2net} & 352$\times$352 & 26.976M &0.820 &0.743 &0.070 &0.782 &0.895&0.888 &0.816 &0.030 &0.835 &0.961&0.815 &0.680 &0.037 &0.718 &0.906 &0.847 &0.770 &0.048 &0.805 &0.914\\
    SegMaR~\cite{jia2022segment} & ResNet-50~\cite{he2016deep} & 352$\times$352 & 56.215M &0.815 &0.753 &0.071 &0.795 &0.884&0.906 &0.860 &0.025 &0.872 &0.959&0.833 &0.724 &0.034 &0.757 &0.906 &0.841 &0.781 &0.046 &0.821 &0.907\\
    CamoFormer-R~\cite{yin2024camoformer} & ResNet-50~\cite{he2016deep} & 352$\times$352 & 71.403M &0.817 &0.752 &0.067 &0.792 &0.885&0.898 &0.847 &0.025 &0.867 &0.956&0.838 &0.724 &0.029 &0.753 &0.930 &0.855 &0.788 &0.042 &0.821 &0.914\\
    ZoomNeXt~\cite{pang2024zoomnext} & ResNet-50~\cite{he2016deep} & 352$\times$352 & 28.458M &0.822 &0.760 &0.069 &0.797 &0.885&0.912 &0.863 &\textcolor{green!60!black}{\textbf{0.020}} &0.878 &\textcolor{blue!90!black}{\textbf{0.969}}&0.855 &0.758 &0.026 &0.791 &0.926 &0.869 &0.808 &0.038 &0.836 &0.925\\
    \rowcolor{gray!15} 
    
    Ours & ResNet-50~\cite{he2016deep} & 352$\times$352 & 28.458M &
    0.820 &0.755& 0.070& 0.792& 0.879& 
    0.913& 0.867& \textcolor{blue!90!black}{\textbf{0.021}}& \textcolor{blue!90!black}{\textbf{0.881}}& 0.967& 
    0.865& 0.777& 0.024& 0.808& 0.930& 
    0.872& 0.816& 0.038& 0.844& 0.926\\ 
    \hline
    
    DGNet~\cite{ji2023deep}& EfficientNet-B4~\cite{tan2019efficientnet}& 352$\times$352& 18.113M&0.838 &0.768 &0.057 &0.805 &0.914&0.890 &0.816 &0.029 &0.834 &0.956&0.822 &0.692 &0.033 &0.727 &0.911&0.857 &0.783 &0.042 &0.813 &0.922\\
    
    ZoomNeXt~\cite{pang2024zoomnext}& EfficientNet-B4~\cite{tan2019efficientnet}& 352$\times$352& 21.381M&
    0.859 &0.815 &0.049 &0.845 &0.920 &0.911 &0.864 &\textcolor{green!60!black}{\textbf{0.020}} &0.877 &0.965 &0.868 &0.789 &\textcolor{blue!90!black}{\textbf{0.023}} &0.818 &0.937 &\textcolor{blue!90!black}{\textbf{0.878}} &0.832 &\textcolor{blue!90!black}{\textbf{0.035}} &\textcolor{blue!90!black}{\textbf{0.857}} &0.932\\
    
    ZoomNeXt~\cite{pang2024zoomnext}& EfficientNet-B4~\cite{tan2019efficientnet}& 384$\times$384& 21.381M&
    \textcolor{blue!90!black}{\textbf{0.867}} &\textcolor{blue!90!black}{\textbf{0.824}} &\textcolor{green!60!black}{\textbf{0.046}} &\textcolor{blue!90!black}{\textbf{0.852}} &\textcolor{blue!90!black}{\textbf{0.925}} 
    
    &0.911 &0.865 &\textcolor{green!60!black}{\textbf{0.020}} &0.879 &0.964 
    
    &\textcolor{blue!90!black}{\textbf{0.875}} 
    &\textcolor{blue!90!black}{\textbf{0.797}} &\textcolor{green!60!black}{\textbf{0.021}} &\textcolor{blue!90!black}{\textbf{0.824}} &\textcolor{blue!90!black}{\textbf{0.941}}
    &\textcolor{green!60!black}{\textbf{0.884}}&\textcolor{blue!90!black}{\textbf{0.837}} &\textcolor{green!60!black}{\textbf{0.032}} &\textcolor{green!60!black}{\textbf{0.862}} &\textcolor{blue!90!black}{\textbf{0.939}}\\
    
    \rowcolor{gray!15} 
    Ours& EfficientNet-B4~\cite{tan2019efficientnet}& 352$\times$352& 21.381M&
    \textcolor{green!60!black}{\textbf{0.868}}& \textcolor{green!60!black}{\textbf{0.827}}& \textcolor{blue!90!black}{\textbf{0.047}}& \textcolor{green!60!black}{\textbf{0.855}}& \textcolor{green!60!black}{\textbf{0.927}}& 

    \textcolor{green!60!black}{\textbf{0.920}}& \textcolor{green!60!black}{\textbf{0.878}}& \textcolor{green!60!black}{\textbf{0.020}}& \textcolor{green!60!black}{\textbf{0.888}}& 0.968& 
    
    \textcolor{green!60!black}{\textbf{0.883}}& \textcolor{green!60!black}{\textbf{0.808}}& \textcolor{red!80!black}{\textbf{0.020}}& \textcolor{green!60!black}{\textbf{0.832}}& \textcolor{green!60!black}{\textbf{0.945}}& 
    
    \textcolor{red!80!black}{\textbf{0.889}}& \textcolor{green!60!black}{\textbf{0.843}}& \textcolor{green!60!black}{\textbf{0.032}}& \textcolor{red!80!black}{\textbf{0.866}}& \textcolor{green!60!black}{\textbf{0.942}}
    \\
    
    \rowcolor{gray!15} 
    Ours& EfficientNet-B4~\cite{tan2019efficientnet}& 384$\times$384&21.381M&
    \textcolor{red!80!black}{\textbf{0.875}}& \textcolor{red!80!black}{\textbf{0.838}}& \textcolor{red!80!black}{\textbf{0.045}}& \textcolor{red!80!black}{\textbf{0.863}}& \textcolor{red!80!black}{\textbf{0.936}}& 
    
    \textcolor{red!80!black}{\textbf{0.923}}& \textcolor{red!80!black}{\textbf{0.881}}& \textcolor{red!80!black}{\textbf{0.019}}& \textcolor{red!80!black}{\textbf{0.891}}& \textcolor{green!60!black}{\textbf{0.970}}& 
    
    \textcolor{red!80!black}{\textbf{0.887}}& \textcolor{red!80!black}{\textbf{0.814}}& \textcolor{red!80!black}{\textbf{0.020}}& \textcolor{red!80!black}{\textbf{0.838}}& \textcolor{red!80!black}{\textbf{0.947}}& 
    \textcolor{red!80!black}{\textbf{0.889}}& \textcolor{red!80!black}{\textbf{0.844}}& \textcolor{red!80!black}{\textbf{0.031}}& \textcolor{red!80!black}{\textbf{0.866}}& \textcolor{red!80!black}{\textbf{0.943}}\\
    \hline
    
    SLSR~\cite{lv2021simultaneously} & ResNet-50~\cite{he2016deep} & 480$\times$480& 50.935M &0.787 &0.696 &0.080 &0.744 &0.854&0.890 &0.822 &0.030 &0.841 &0.948&0.804 &0.673 &0.037 &0.715 &0.892 &0.840 &0.765 &0.048 &0.804 &0.907\\
    MGL-R~\cite{zhai2021mutual} & ResNet-50~\cite{he2016deep} &473$\times$473& 63.595M &0.775 &0.673 &0.088 &0.726 &0.842&0.893 &0.812 &0.030 &0.834 &0.941&0.814 &0.666 &0.035 &0.710 &0.890 &0.833 &0.739 &0.053 &0.782 &0.893\\
    PFNet~\cite{mei2021camouflaged} & ResNet-50~\cite{he2016deep} &416$\times$416& 46.498M&0.782 &0.695 &0.085 &0.746 &0.855&0.882 &0.810 &0.033 &0.828 &0.945&0.800 &0.660 &0.040 &0.701 &0.890 &0.829 &0.745 &0.053 &0.784 &0.898\\
    UJSC~\cite{li2021uncertainty} & ResNet-50~\cite{he2016deep} & 480$\times$480 & 217.982M &0.800 &0.728 &0.073 &0.772 &0.873&0.891 &0.833 &0.030 &0.848 &0.955&0.809 &0.684 &0.035 &0.721 &0.891 &0.842 &0.771 &0.046 &0.806 &0.907\\
    UGTR~\cite{yang2021uncertainty} & ResNet-50~\cite{he2016deep} & 473$\times$473 & 48.868M &0.784 &0.684 &0.086 &0.735 &0.851&0.887 &0.794 &0.031 &0.819 &0.940&0.817 &0.666 &0.036 &0.711 &0.890 &0.839 &0.746 &0.052 &0.787 &0.899\\
    ZoomNet~\cite{pang2022zoom} & ResNet-50~\cite{he2016deep} & 384$\times$384 & 32.382M &0.820 &0.752 &0.066 &0.794 &0.892&0.902 &0.845 &0.023 &0.864 &0.958&0.838 &0.729 &0.029 &0.766 &0.911 &0.853 &0.784 &0.043 &0.818 &0.912\\
    BSA-Net~\cite{zhu2022can} & Res2Net-50~\cite{gao2019res2net} & 384$\times$384 & 32.585M &0.794 &0.717 &0.079 &0.763 &0.867&0.895 &0.841 &0.027 &0.858 &0.957&0.818 &0.699 &0.034 &0.738 &0.901 &0.842 &0.771 &0.048 &0.808 &0.907\\
    BGNet~\cite{sun2022boundary} & Res2Net-50~\cite{gao2019res2net} & 416$\times$416 & 79.853M &0.812 &0.749 &0.073 &0.789 &0.882&0.901 &0.851 &0.027 &0.860 &0.954&0.831 &0.722 &0.033 &0.753 &0.911 &0.851 &0.788 &0.044 &0.820 &0.916\\
    FEDER~\cite{he2023camouflaged} & ResNet-50~\cite{he2016deep} & 384$\times$384 & 44.129M &0.802 &0.738 &0.071 &0.781 &0.873&0.887 &0.835 &0.030 &0.851 &0.954&0.822 &0.716 &0.032 &0.751 &0.905 &0.847 &0.789 &0.044 &0.824 &0.915\\
    ZoomNeXt~\cite{pang2024zoomnext} & ResNet-50~\cite{he2016deep} & 384$\times$384 & 28.458M &0.833 &0.774 &0.065 &0.813 &0.891&0.908 &0.858 &\textcolor{blue!90!black}{\textbf{0.021}} &0.874 &0.963&0.861 &0.768 &0.026 &0.801 &0.925 &0.874 &0.816 &0.037 &0.846 &0.928\\
    \rowcolor{gray!15} 
    
    Ours & ResNet-50~\cite{he2016deep} & 384$\times$384 & 28.458M &
    0.816& 0.754& 0.071& 0.794& 0.872& 
    \textcolor{blue!90!black}{\textbf{0.918}}& \textcolor{blue!90!black}{\textbf{0.876}}& \textcolor{green!60!black}{\textbf{0.020}}& \textcolor{green!60!black}{\textbf{0.888}}& \textcolor{red!80!black}{\textbf{0.975}}& 
    0.868& 0.786& 0.024& 0.816& 0.934& 
    0.869& 0.814& 0.039& 0.844& 0.925\\
   
    \hline
   \multicolumn{24}{c}{\textbf{{Vision Transformer based Methods}}} \\
       \hline
       CamoFormer-P~\cite{yin2024camoformer}  & PVTv2-B4~\cite{wang2022pvt} & 352$\times$352 & 71.403M &0.872 &0.831 &0.046 &0.854 &0.938&0.910 &0.865 &0.022 &0.882 &0.966&0.869 &0.786 &0.023 &0.811 &0.939 &0.892 &0.847 &0.030 &0.868 &0.946\\
       
       ZoomNeXt~\cite{pang2024zoomnext} & PVTv2-B4~\cite{wang2022pvt} & 352$\times$352 & 65.374M 
       &\textcolor{red!80!black}{\textbf{0.893}} &\textcolor{red!80!black}{\textbf{0.862}} &\textcolor{red!80!black}{\textbf{0.040}} &\textcolor{red!80!black}{\textbf{0.881}} &\textcolor{red!80!black}{\textbf{0.949}}
       &0.929 &0.894 &\textcolor{blue!90!black}{\textbf{0.018}} &0.906 &\textcolor{blue!90!black}{\textbf{0.977}}&0.895 &0.825 &\textcolor{blue!90!black}{\textbf{0.018}} &0.845 &0.954 &0.899 &0.859 &0.029 &0.879 &0.949\\
       
       \rowcolor{gray!15} 
       Ours& PVTv2-B4~\cite{wang2022pvt} & 352$\times$352 & 65.373M
       &\textcolor{green!60!black}{\textbf{0.889}}
       &\textcolor{green!60!black}{\textbf{0.861}} &\textcolor{green!60!black}{\textbf{0.041}} &\textcolor{green!60!black}{\textbf{0.878}}
       &\textcolor{blue!90!black}{\textbf{0.944}}
       &0.930 
       &\textcolor{blue!90!black}{\textbf{0.907}} &\textcolor{green!60!black}{\textbf{0.017}}
       &\textcolor{green!60!black}{\textbf{0.915}} &\textcolor{red!80!black}{\textbf{0.979}}
       
       &\textcolor{green!60!black}{\textbf{0.905}}
       &\textcolor{green!60!black}{\textbf{0.848}} 
       &\textcolor{red!80!black}{\textbf{0.016}} &\textcolor{green!60!black}{\textbf{0.865}} &\textcolor{red!80!black}{\textbf{0.962}} 
       
       &\textcolor{green!60!black}{\textbf{0.904}} &\textcolor{blue!90!black}{\textbf{0.870}} 
       &\textcolor{green!60!black}{\textbf{0.027}}
       &\textcolor{blue!90!black}{\textbf{0.887}}
       &\textcolor{green!60!black}{\textbf{0.952}}\\ 
       
       \hline
       
        MSCAF-Net~\cite{liu2023mscaf} & PVTv2-B2~\cite{wang2022pvt} & 352$\times$352 & 30.364M &0.873 &0.828 &0.046 &0.852 &0.937&0.912 &0.865 &0.022 &0.876 &0.970&0.865 &0.775 &0.024 &0.798 &0.936 &0.887 &0.838 &0.032 &0.860 &0.942\\
        
        ZoomNeXt~\cite{pang2024zoomnext} & PVTv2-B2~\cite{wang2022pvt} & 352$\times$352 & 28.181M &0.868 &0.829 &0.049 &0.855 &0.926&0.916 &0.876 &\textcolor{blue!90!black}{\textbf{0.018}} &0.889 &0.971&0.881 &0.809 &0.020 &0.834 &0.945 &0.890 &0.848 &0.031 &0.872 &0.941\\ 
        
        \rowcolor{gray!15} 
        Ours& PVTv2-B2~\cite{wang2022pvt} & 352$\times$352 & 28.180M 
        &0.869& 0.832& 0.050& 0.854& 0.924
        &\textcolor{blue!90!black}{\textbf{0.931}}
        & 0.901
        &\textcolor{green!60!black}{\textbf{0.017}}
        &0.911
        &\textcolor{red!80!black}{\textbf{0.979}}
        &0.893& 0.827& 0.019& 0.848& 0.950
        &0.893& 0.852& 0.030& 0.874& 0.941 \\ 

       \hline
       HitNet~\cite{hu2023high} & PVTv2-B2~\cite{wang2022pvt} & 704$\times$704 & 25.727M &0.849 &0.809 &0.055 &0.831 &0.910&0.921 &0.897 &0.019 &0.900 &0.972&0.871 &0.806 &0.023 &0.823 &0.938 &0.875 &0.834 &0.037 &0.854 &0.929\\
       
       FSPNet~\cite{huang2023feature} & ViT-B/16~\cite{dosovitskiy2020image} & 384$\times$384 & 274.240M &0.856 &0.799 &0.050 &0.831 &0.928&0.908 &0.851 &0.023 &0.867 &0.965&0.851 &0.735 &0.026 &0.769 &0.930 &0.878 &0.816 &0.035 &0.843 &0.937\\
       
       SARNet~\cite{xing2023go} & PVTv2-B3~\cite{wang2022pvt} & 672$\times$672 & 47.477M &0.874 &0.844 &0.046 &0.866 &0.935&\textcolor{red!80!black}{\textbf{0.933}} &\textcolor{red!80!black}{\textbf{0.909}} &\textcolor{green!60!black}{\textbf{0.017}} &\textcolor{green!60!black}{\textbf{0.915}} &\textcolor{green!60!black}{\textbf{0.978}}& 0.885 &0.820 &0.021 &0.839 &0.947 &0.889 &0.851 &0.032 &0.872 &0.940\\
       
       ZoomNeXt~\cite{pang2024zoomnext} & PVTv2-B2~\cite{wang2022pvt} & 384$\times$384 & 28.181M &0.874 &0.839 &0.047 &0.863 &0.931&0.922 &0.884 &\textcolor{green!60!black}{\textbf{0.017}} &0.896 &0.970&0.887 &0.818 &0.019 &0.841 &0.948 &0.892 &0.852 &0.030 &0.874 &0.943\\
       
       ZoomNeXt~\cite{pang2024zoomnext} & PVTv2-B3~\cite{wang2022pvt} & 384$\times$384 & 48.056M & 0.885 &0.854 &\textcolor{blue!90!black}{\textbf{0.042}} &0.872& 0.942&0.927 &0.898 &\textcolor{green!60!black}{\textbf{0.017}}& 0.905 &\textcolor{blue!90!black}{\textbf{0.977}}&0.895& 0.829 &\textcolor{blue!90!black}{\textbf{0.018}} &0.848 &0.952 & 0.900 &0.861 &\textcolor{blue!90!black}{\textbf{0.028}} &0.880 &0.949\\

       ZoomNeXt~\cite{pang2024zoomnext} & PVTv2-B4~\cite{wang2022pvt} & 384$\times$384 & 65.374M &\textcolor{blue!90!black}{\textbf{0.888}} &0.859 &\textcolor{red!80!black}{\textbf{0.040}}  &\textcolor{green!60!black}{\textbf{0.878}}& 0.943&
       0.925 &0.897 &\textcolor{red!80!black}{\textbf{0.016}}&0.906&0.973&0.898 &0.838 &\textcolor{green!60!black}{\textbf{0.017}} &0.857 &0.955 & 0.900 & 0.865 &\textcolor{blue!90!black}{\textbf{0.028}} & 0.884 & 0.949\\

       ZoomNeXt~\cite{pang2024zoomnext} & PVTv2-B5~\cite{wang2022pvt} & 384$\times$384 & 84.774M &\textcolor{green!60!black}{\textbf{0.889}} 
       &0.857 
       &\textcolor{green!60!black}{\textbf{0.041}}
       &0.875
       &\textcolor{green!60!black}{\textbf{0.945}}& 
       0.924&0.885 &\textcolor{blue!90!black}{\textbf{0.018}}& 0.896 & 0.975 & 0.898&0.827 &\textcolor{blue!90!black}{\textbf{0.018}}&0.848 & 0.956 &\textcolor{blue!90!black}{\textbf{0.903}}&0.863 &\textcolor{blue!90!black}{\textbf{0.028}} &0.884 &\textcolor{blue!90!black}{\textbf{0.951}}\\
       
\rowcolor{gray!15} 
Ours & PVTv2-B2~\cite{wang2022pvt} & 384$\times$384 & 28.180M & 0.873& 0.838& 0.047& 0.860& 0.928
&\textcolor{blue!90!black}{\textbf{0.931}}& 0.904& \textcolor{red!80!black}{\textbf{0.016}}& \textcolor{blue!90!black}{\textbf{0.912}}& 0.976&
0.894& 0.829& \textcolor{blue!90!black}{\textbf{0.018}}& 0.849& 0.952& 
0.894& 0.853& 0.030& 0.874& 0.943\\

\rowcolor{gray!15} 
Ours & PVTv2-B3~\cite{wang2022pvt}&384$\times$384 & 48.056M &
0.885& 0.855& 0.043& 0.874& 0.941&
\textcolor{red!80!black}{\textbf{0.933}}& \textcolor{blue!90!black}{\textbf{0.907}}& \textcolor{red!80!black}{\textbf{0.016}}& \textcolor{green!60!black}{\textbf{0.915}}& 0.973&
\textcolor{blue!90!black}{\textbf{0.904}}& \textcolor{blue!90!black}{\textbf{0.847}}& \textcolor{green!60!black}{\textbf{0.017}}&  \textcolor{green!60!black}{\textbf{0.865}}& \textcolor{green!60!black}{\textbf{0.959}}&
\textcolor{blue!90!black}{\textbf{0.903}}& 0.867& \textcolor{green!60!black}{\textbf{0.027}}& 0.886& \textcolor{green!60!black}{\textbf{0.952}}\\
    
\rowcolor{gray!15} 
Ours & PVTv2-B4~\cite{wang2022pvt} & 384$\times$384 & 65.373M &
\textcolor{blue!90!black}{\textbf{0.888}}& \textcolor{green!60!black}{\textbf{0.861}}& \textcolor{red!80!black}{\textbf{0.040}}& \textcolor{green!60!black}{\textbf{0.878}}& 0.942& 
\textcolor{green!60!black}{\textbf{0.932}}& \textcolor{green!60!black}{\textbf{0.908}}& \textcolor{green!60!black}{\textbf{0.017}}& \textcolor{red!80!black}{\textbf{0.916}}& \textcolor{green!60!black}{\textbf{0.978}}& 
\textcolor{red!80!black}{\textbf{0.907}}& \textcolor{red!80!black}{\textbf{0.852}}& \textcolor{red!80!black}{\textbf{0.016}}& \textcolor{red!80!black}{\textbf{0.868}}& \textcolor{red!80!black}{\textbf{0.962}}& 
\textcolor{red!80!black}{\textbf{0.905}}& \textcolor{red!80!black}{\textbf{0.873}}& \textcolor{red!80!black}{\textbf{0.026}}& \textcolor{red!80!black}{\textbf{0.890}}& \textcolor{red!80!black}{\textbf{0.953}}\\

\rowcolor{gray!15} 
Ours & PVTv2-B5~\cite{wang2022pvt} & 384$\times$384 & 84.773M &
\textcolor{blue!90!black}{\textbf{0.888}}& \textcolor{blue!90!black}{\textbf{0.860}}& \textcolor{green!60!black}{\textbf{0.041}}& \textcolor{blue!90!black}{\textbf{0.876}}& 0.943& 

0.925& 0.893& \textcolor{green!60!black}{\textbf{0.017}}& 0.903& 0.971&

0.902& 0.844& \textcolor{green!60!black}{\textbf{0.017}}& \textcolor{blue!90!black}{\textbf{0.862}}& \textcolor{blue!90!black}{\textbf{0.957}}&

\textcolor{blue!90!black}{\textbf{0.903}}& \textcolor{green!60!black}{\textbf{0.871}}& \textcolor{green!60!black}{\textbf{0.027}}& \textcolor{green!60!black}{\textbf{0.889}}& \textcolor{green!60!black}{\textbf{0.952}}\\

    \hline
  \end{tabular}}
\end{table*}